%% file: main.tex
\documentclass{article} 
\usepackage{iclr2027_conference,times}

\input{math_commands.tex}

\usepackage{hyperref}
\usepackage{url}

\usepackage{wrapfig}
\usepackage{graphicx}
\usepackage{array}
\usepackage{multirow}
\usepackage[table,xcdraw]{xcolor}

\definecolor{navylink}{RGB}{16,58,112}
\definecolor{refred}{HTML}{ED1C24}
\hypersetup{
  colorlinks=true,
  citecolor=navylink,
  linkcolor=refred,
  urlcolor=navylink,
  filecolor=navylink
}
\usepackage{booktabs}
\usepackage{multirow}
\usepackage{algorithm}
\usepackage{algpseudocode}
\usepackage{amsthm}
\usepackage{amssymb}
\usepackage{pifont}
\usepackage{caption}
\usepackage{enumitem}
\usepackage{xspace}
\usepackage{epigraph}
\usepackage{tabularx}
\usepackage{alltt}
\usepackage[most]{tcolorbox}
\usepackage{fontawesome5}
\usepackage[table,xcdraw]{xcolor}

\newcolumntype{Y}{>{\raggedright\arraybackslash}X}

\definecolor{promptbg}{HTML}{F8F6F1}
\definecolor{promptborder}{HTML}{E6DFD3}
\definecolor{promptaccent}{HTML}{6E6152}
\definecolor{keybg}{HTML}{F0F7FF}
\definecolor{keyborder}{HTML}{D0D9E5}
\definecolor{keyaccent}{HTML}{4A86E8}
\definecolor{tocaccent}{HTML}{2C5AA0}
\definecolor{promptslot}{HTML}{2E6E7E}
\definecolor{jsonstring}{HTML}{A24632}
\definecolor{jsonnumber}{HTML}{276B62}
\definecolor{jsonliteral}{HTML}{6B4C9A}
\definecolor{jsonpunct}{HTML}{586069}

\lstdefinelanguage{GarmentJSON}{
  basicstyle=\ttfamily\scriptsize,
  string=[b]",
  stringstyle=\color{jsonstring},
  showstringspaces=false,
  breaklines=true,
  breakatwhitespace=true,
  columns=fullflexible,
  keepspaces=true,
  aboveskip=0pt,
  belowskip=0pt,
  literate=
   *{0}{{{\color{jsonnumber}0}}}{1}
    {1}{{{\color{jsonnumber}1}}}{1}
    {2}{{{\color{jsonnumber}2}}}{1}
    {3}{{{\color{jsonnumber}3}}}{1}
    {4}{{{\color{jsonnumber}4}}}{1}
    {5}{{{\color{jsonnumber}5}}}{1}
    {6}{{{\color{jsonnumber}6}}}{1}
    {7}{{{\color{jsonnumber}7}}}{1}
    {8}{{{\color{jsonnumber}8}}}{1}
    {9}{{{\color{jsonnumber}9}}}{1}
    {true}{{{\color{jsonliteral}true}}}{4}
    {false}{{{\color{jsonliteral}false}}}{5}
    {null}{{{\color{jsonliteral}null}}}{4}
    {:}{{{\color{jsonpunct}{:}}}}{1}
    {,}{{{\color{jsonpunct}{,}}}}{1}
    {\{}{{{\color{jsonpunct}{\{}}}}{1}
    {\}}{{{\color{jsonpunct}{\}}}}}{1}
    {[}{{{\color{jsonpunct}{[}}}}{1}
    {]}{{{\color{jsonpunct}{]}}}}{1}
}
\newcommand{\slot}[1]{\textcolor{promptslot}{\{#1\}}}

\newtcolorbox{promptbox}[2][]{
  enhanced, breakable,
  boxsep=4pt, left=8pt, right=7pt, top=5pt, bottom=4pt,
  colback=promptbg, colframe=promptborder, boxrule=0.4pt, arc=3pt,
  frame hidden, borderline west={2.5pt}{0pt}{promptaccent},
  title={#2}, colbacktitle=promptaccent, coltitle=white,
  fonttitle={\fontsize{8}{10}\selectfont\bfseries},
  attach boxed title to top left={yshift=-2.2mm, xshift=3mm},
  boxed title style={enhanced, left=3pt, right=3pt, top=1pt, bottom=1pt,
    boxsep=2pt, arc=2pt, boxrule=0pt, colback=promptaccent},
  #1
}

\newtcolorbox{keybox}[2][]{
  enhanced, breakable,
  boxsep=4pt, left=8pt, right=7pt, top=5pt, bottom=4pt,
  colback=keybg, colframe=keyborder, boxrule=0.4pt, arc=3pt,
  frame hidden, borderline west={2.5pt}{0pt}{keyaccent},
  title={#2}, colbacktitle=keyaccent, coltitle=white,
  fonttitle={\fontsize{8}{10}\selectfont\bfseries},
  attach boxed title to top left={yshift=-2.2mm, xshift=3mm},
  boxed title style={enhanced, left=3pt, right=3pt, top=1pt, bottom=1pt,
    boxsep=2pt, arc=2pt, boxrule=0pt, colback=keyaccent},
  #1
}

\newcommand{\tocsec}[2]{%
  \begingroup\hypersetup{linkcolor=tocaccent}%
  \vspace{5pt}%
  \noindent\textcolor{tocaccent}{\textbf{\hyperref[#1]{\ref*{#1}\hspace{0.9em}#2}}}%
  \nobreak\leaders\hbox{\hspace{2pt}\textcolor{tocaccent!45}{.}\hspace{2pt}}\hfill
  \textcolor{tocaccent}{\hyperref[#1]{\pageref*{#1}}}\par\vspace{2.5pt}%
  \endgroup}
\newcommand{\tocsub}[2]{%
  \begingroup\hypersetup{linkcolor=tocaccent}%
  \noindent\hspace{1.8em}\textcolor{tocaccent}{\hyperref[#1]{\ref*{#1}\hspace{0.9em}#2}}%
  \nobreak\leaders\hbox{\hspace{2pt}\textcolor{tocaccent!45}{.}\hspace{2pt}}\hfill
  \textcolor{tocaccent}{\hyperref[#1]{\pageref*{#1}}}\par\vspace{2.5pt}%
  \endgroup}

\newcommand{\ours}{DressAgent\xspace}

\title{Inverse Garment Design with Visual Agents}

\author{%
  Nanxiang Jiang$^{1}$,
  Yunuo Chen$^{1,5}$,
  Zeng Tao$^{1}$,
  Zizhang Li$^{2}$,
  Qi Guo$^{3}$,
  Jiajun Wu$^{2}$,
  \\[2pt]
  \bfseries
  Yin Yang$^{4,5}$,
  Chenfanfu Jiang$^{1,5}$
  \\[7pt]
  $^{1}$University of California, Los Angeles \quad
  $^{2}$Stanford University
  \\[2pt]
  $^{3}$Independent Researcher \quad
  $^{4}$University of Utah \quad
  $^{5}$Envora
}

\iclrfinalcopy
\begin{document}

\maketitle
\lhead{Preprint}

\begin{center}
  \vspace{-19pt}
  {\ttfamily\small
    \href{https://nxjiang-jnx.github.io/dress-agent}{%
      \raisebox{-0.65em}{\includegraphics[height=2.2em]{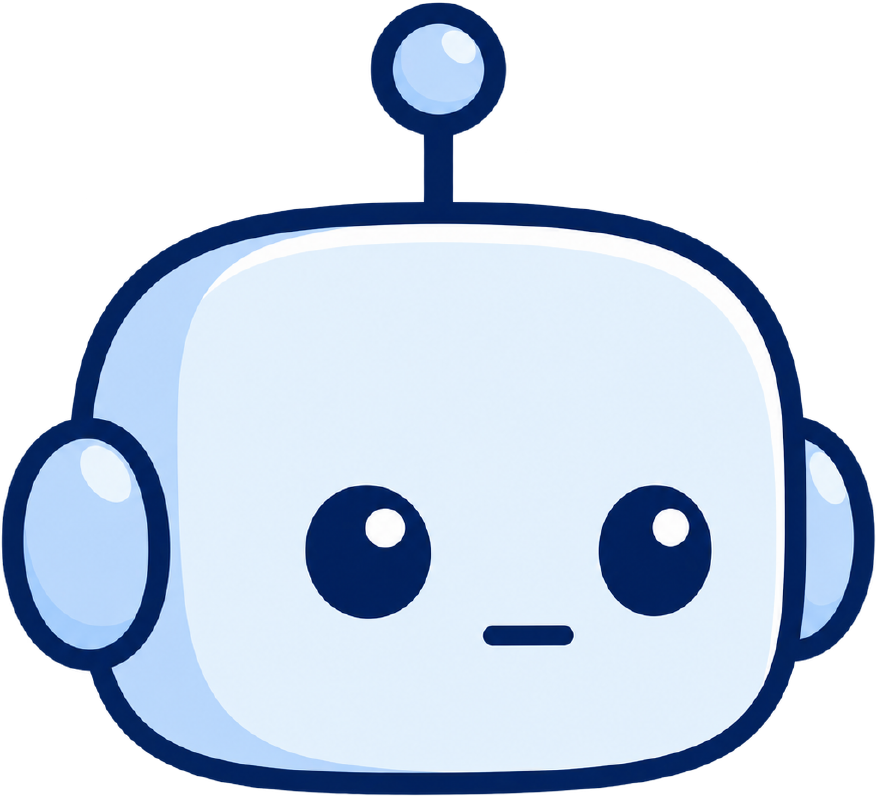}}\,
      nxjiang-jnx.github.io/dress-agent}}
  \vspace{4pt}
\end{center}

\begin{abstract}
We study the problem of inverse garment design, reconstructing simulation-ready 3D outfits with arbitrary layers from a single image, with visual agents.
Unlike differentiable optimization that is time-consuming, or feed-forward methods limited by training data, we introduce \textbf{\ours}, a hierarchical agentic system that recovers garment sewing patterns through planning and closed-loop refinement grounded in physical simulation.
To this end, we first present an enhanced representation based on GarmentCode for diverse styles and arbitrary dressing layers, and build the first multi-layer sewing pattern dataset with over 15K outfits for supervision and benchmarking.
Inverse design then proceeds coarse to fine:
In \emph{layer planning and coarse estimation}, agents infer layer order and relations, peel each layer into its own view, and estimate coarse patterns.
In \emph{fine-grained refinement}, an agentic trio of Tailor, Critic, and Orchestrator refines patterns in a closed loop of editing, simulating, and judging against the reference with visual and measured evidence.
Carefully designed tools and an evolving memory make this refinement converge on the target geometry.
We also design an agentic generation method for high-fidelity textures.
\ours outperforms feed-forward methods by a large margin, surpasses or matches differentiable optimization in a fraction of the time, and generalizes to complex layering, diverse poses, and body shapes.
\end{abstract}

\input{section/introduction}
\input{section/related}
\input{section/method}
\input{section/experiment}
\input{section/conclusion}

\newpage
\bibliography{ref}
\bibliographystyle{iclr2027_conference}

\newpage
\appendix
\input{section/appendix}

\end{document}

%% file: math_commands.tex
\usepackage{amsmath,amsfonts,bm}

\def\eqref#1{equation~\ref{#1}}

\def\1{\bm{1}}

\DeclareMathAlphabet{\mathsfit}{\encodingdefault}{\sfdefault}{m}{sl}
\SetMathAlphabet{\mathsfit}{bold}{\encodingdefault}{\sfdefault}{bx}{n}



%% file: section/introduction.tex
\vspace{-10pt}

\epigraph{A couturier must be an architect for design, a sculptor for shape, a painter for color, a musician for harmony, and a philosopher for temperance.}{--- \textit{Crist\'obal Balenciaga}}

\vspace{-10pt}

\section{Introduction}
\label{sec:intro}

\begin{figure}[t]
    \centering
    \includegraphics[width=1.0\textwidth]{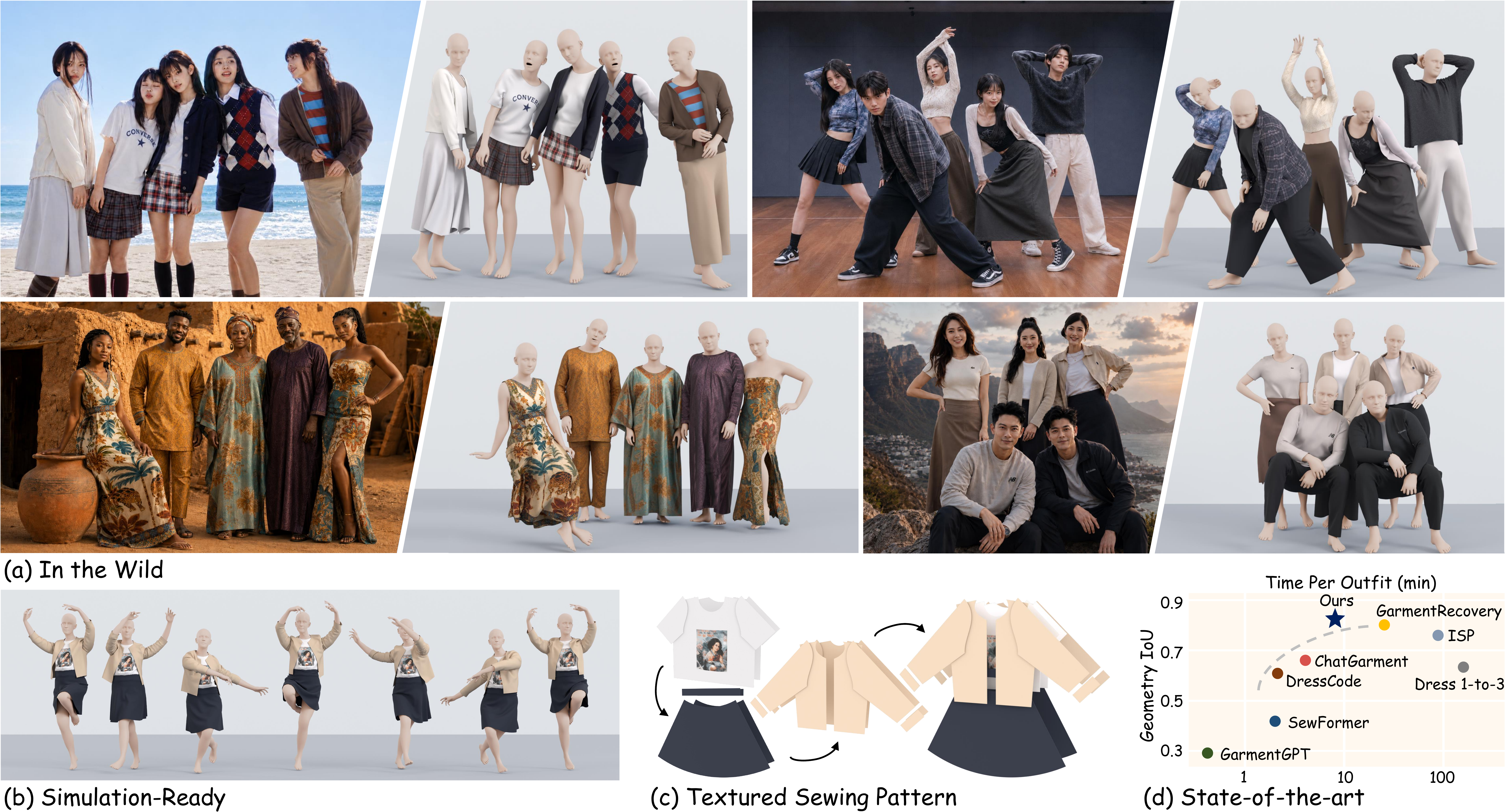}\
    \vspace{-12pt}
    \caption{\textbf{We introduce \ours for inverse garment design with agentic planning and refinement.} (a) Results on in-the-wild photos with arbitrary layering, diverse poses, and body shapes. (b) The outfits drape and animate under physics. (c) Every garment is a sewing pattern with high-fidelity texture. (d) \ours pushes the Pareto frontier of quality and speed.}
    \label{fig:teaser}
    \vspace{-10pt}
\end{figure}

Imagine a photo of a beautiful outfit you would love to wear yourself.
A skilled couturier can read that photo: how many layers, where the hem falls, how the fabric hangs on the body.
From this reading, the couturier drafts the flat patterns, fits a trial on the body, and redraws them until the copy matches the original.
We call this problem \emph{inverse garment design}: from a single in-the-wild image, recover the outfit as sewing patterns ready to simulate, animate, and even manufacture (Fig.~\ref{fig:teaser}).

Recent work on inverse garment design with sewing patterns mainly follows two paradigms:
\emph{Differentiable optimization methods}~\citep{isp, dress123, garmentrecovery} simulate the garment, compare it with the image, and update the pattern by gradient descent. They recover faithful geometry, but are time-consuming and vulnerable to local minima in optimization.
\emph{Feed-forward methods}~\citep{sewformer, chatgarment, dresswild} train a network on large-scale datasets, and predict a pattern in one pass.
They are efficient but limited by the data distribution and struggle to generalize to occlusion or complex layering.
Vision-language models (VLMs) are recently used to write the pattern program in a single pass~\citep{ngl,design2garmentcode}, leveraging broad semantic priors to generalize beyond the training data and bypass costly optimization.
However, this one-shot process is inherently open-loop, and without fine-grained visual grounding, even powerful VLMs struggle to recover subtle details.
Moreover, people wear layered outfits in daily life (\textit{e.g.}, a cardigan over a dress), yet few methods support zero-shot recovery of arbitrary layers and their relations.

We propose \textbf{\ours}, a hierarchical agentic system for inverse garment design.
Since garments can be written as code~\citep{garmentcode}, our key insight is to solve inverse garment design the way agents already excel at coding: writing code and refining it against execution feedback.
We therefore design a closed-loop system with agentic planning and refinement, grounded in physical simulation.
To unleash this coding ability on real-world outfits, the agent needs a language expressive enough to write them, so we first introduce GarmentCode-XL (eXtended \& Layered), an enhanced representation based on GarmentCode, which adds diverse styles and arbitrary dressing layers with their relations (\textit{e.g.}, tucked into, worn over), and extend the cloth simulator to realize these relations.
As no data exists for this new language, we build GarmentCodeData-XL, the first multi-layer sewing pattern dataset with over 15K outfits for supervision and benchmarking.

Building on this foundation, \ours recovers the outfit in two stages, from coarse to fine.
In \emph{layer planning and coarse estimation}, a Layer Planner decomposes the outfit into ordered layers with carefully designed skills, and peels each layer into its own view.
A Visual Reader then generates a style template per layer, decoded into a coarse program.
In \emph{fine-grained refinement}, three agents interact in a closed loop: a Tailor that edits and simulates the pattern, a Critic that judges the drape against the reference with visual and measured evidence, and an Orchestrator that manages the iterative refinement.
With an evolving memory of past edits and their measured effects, the loop converges progressively.
Since sewing patterns capture only geometry, we further design a novel agentic generation method with tool calls to recover high-fidelity textures from the input image.

Experiments show that \ours outperforms feed-forward methods by a large margin, surpasses or matches differentiable optimization in a fraction of the time, and achieves state-of-the-art preference (Fig.~\ref{fig:teaser}).
In summary, our contributions are:
\begin{itemize}[nosep,leftmargin=*]
    \item We introduce \ours, a hierarchical agentic system for inverse garment design. It recovers outfits with arbitrary layers and their relations from a single in-the-wild image, through agentic planning and closed-loop refinement grounded in physical simulation.
    \item To unleash the most potential of coding agents on real-world garments, we propose GarmentCode-XL, an enhanced garment representation and GarmentCodeData-XL, the first multi-layer sewing pattern dataset with 15K outfits, together with an agentic method for high-fidelity textures.
    \item We show that \ours achieves the best overall accuracy and human preference, generalizing far beyond feed-forward methods at a fraction of the cost of differentiable optimization.
\end{itemize}

%% file: section/related.tex
\section{Related Work}

\textbf{Garment Representations.}
3D garment representations have been widely explored.
\emph{Template-based methods}~\citep{bcnet,tailornet,garverselod} deform a predefined mesh and stay topologically valid, but cannot generalize beyond that template.
\emph{Free-form fields}~\citep{smplicit,drapenet,clothwild} represent arbitrary shapes but lose garment structure, and are hard to rig and simulate.
\emph{Sewing patterns}~\citep{neuraltailor,isp} retain both: a garment is described by 2D panels and their stitches, just as it is sewn, so it stays editable and simulation-ready.
GarmentCode~\citep{garmentcode} further writes the pattern as code, which is a natural fit for coding agents and the basis of our work.

\textbf{Inverse Garment Design with Sewing Patterns.}
Single-image garment recovery with sewing patterns mainly follows two paradigms:
\emph{Differentiable optimization methods}~\citep{garmentrecovery,dress123,diffavatar} fit each image by gradient descent through a differentiable cloth simulator, recovering faithful geometry at a high cost in time and expertise.
\emph{Feed-forward methods}~\citep{sewformer,dresscode,garmentgpt,dresswild} train on paired datasets and predict in one pass, but stay limited by that distribution and lose accuracy on in-the-wild photographs (\textit{e.g.}, occlusion, unusual poses, complex layering).
VLMs are now used in both lines~\citep{chatgarment,aipparel,design2garmentcode} to estimate the pattern from an image, which generalizes better but remains open-loop, so even powerful VLMs cannot verify or refine subtle details.

\textbf{Agentic Systems for Inverse Graphics.}
Agents can now solve hard tasks by writing code and refining it against execution feedback~\citep{reflexion,textgrad,robopara}, and this loop is reaching inverse graphics: VLM agents refine 6D object poses from rendered feedback~\citep{6dobjectpose}, and reconstruct Blender scenes through execution and verification~\citep{VIGA}.
Garments are harder than such rigid scenes, since cloth is deformable with complex structures and relations.
To the best of our knowledge, we introduce the first agentic system for inverse garment design, with planning and closed-loop refinement grounded in physical simulation.

%% file: section/method.tex
\section{\ours}

\begin{figure}[t]
  \centering
  \includegraphics[width=0.9\textwidth]{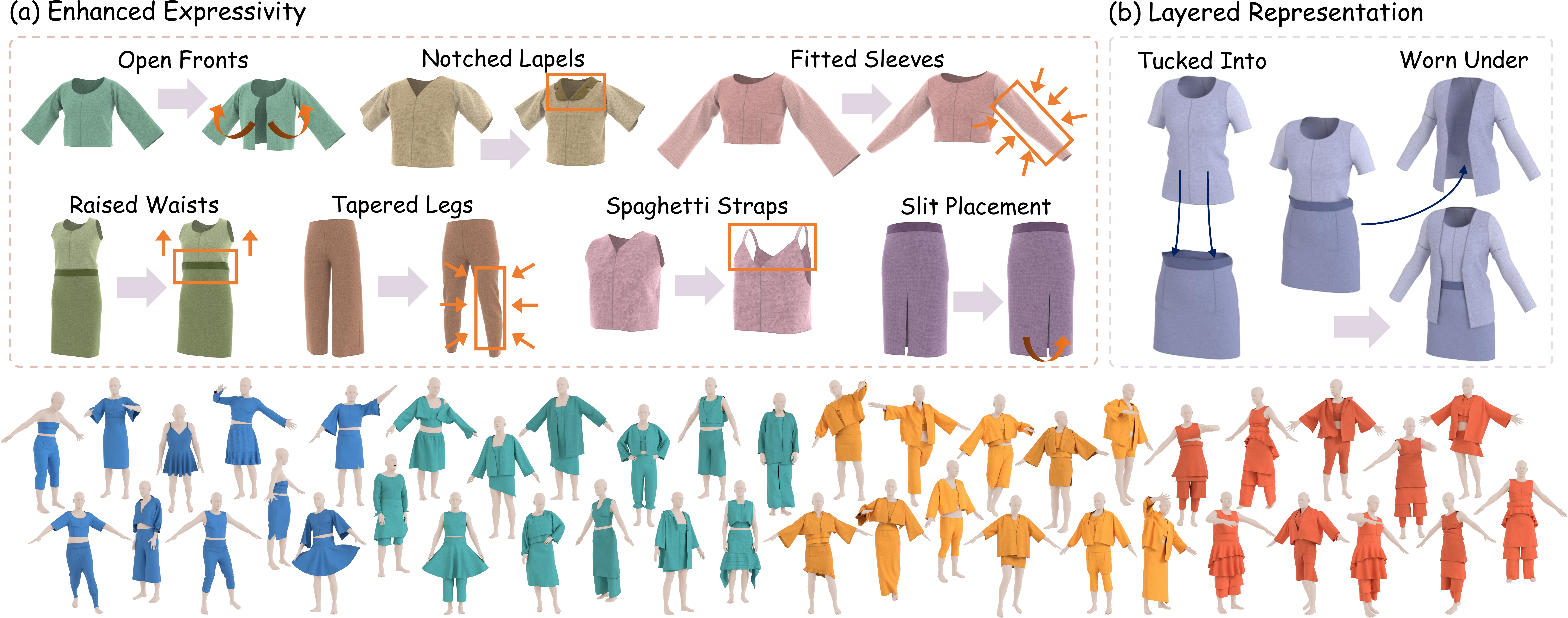}
  \vspace{-5pt}
  \caption{\textbf{Enhanced Garment Representation and Dataset.} \emph{Top}: Examples of enhanced expressivity and layered representation in GarmentCode-XL. \emph{Bottom}: Examples from GarmentCodeData-XL. Its $15{,}205$ outfits contain $4{,}640$ one-layer, $9{,}282$ two-layer, $721$ three-layer, and $562$ four-layer examples. From left to right, colors indicate one to four layers. For more details, see Appendix~\ref{appendix:garmentcodexl}.}
  \label{fig:garmentcodexl}
  \vspace{-10pt}
\end{figure}

\textbf{Overview.}
We present \ours, a hierarchical agentic system for inverse garment design from a single in-the-wild image.
We first introduce an enhanced sewing pattern representation and the first multi-layer garment dataset (Sec.~\ref{sec:prelim}).
Inverse design then proceeds coarse to fine: in \emph{layer planning and coarse estimation}, agents infer ordered layers and their relations and estimate sewing pattern programs (Sec.~\ref{sec:stage1}).
In \emph{fine-grained refinement}, agents refine these programs through closed-loop tool calls to match the reference (Sec.~\ref{sec:stage2}).
Finally, we design an agentic generative method for high-fidelity textures, and animate the outfit with physics-based simulation (Sec.~\ref{sec:postprocess}).

\subsection{Garment Representation and Dataset}
\label{sec:prelim}

In the garment design industry, a garment is made of flat panels sewn together along their edges.
GarmentCode~\citep{garmentcode} turns this practice into code programs.
A garment program $P$ generates the sewing pattern $S$ from body measurements $B$:
\begin{equation}
  S = P(B) = \bigcup_i f_i(d_i,B), \qquad P=\langle\mathcal{F},\mathcal{D}\rangle,
\end{equation}
where $\mathcal{F}=\{f_i\}$ are panel constructors and $\mathcal{D}=\{d_i\}$ are their design parameters.
Each $f_i$ drafts a component (\textit{e.g.}, a sleeve or bodice) using 2D curves controlled by $d_i$ and body measurements $B$.
For simulation, a Warp-based cloth solver~\citep{warp} stitches and drapes $S$ into a mesh $M=\mathrm{Drape}(P,B)$.
Our goal is to recover $P$ whose simulated garment matches the input image.

However, the original GarmentCode covers limited styles and no multi-layer outfits, so agents lack a language expressive enough for real-world outfits.
We therefore introduce GarmentCode-XL (eXtended \& Layered) with two main extensions (Fig.~\ref{fig:garmentcodexl}).
First, we add new panel constructors to $\mathcal{F}$ and widen parameter ranges in $\mathcal{D}$ for diverse real-world styles (\textit{e.g.}, open-front shells, spaghetti straps).
Second, we introduce a layered data structure $\mathcal{P}=\langle(P_1,\dots,P_K),\mathcal{R}\rangle$ with $K$ layers ordered from inner to outer, and the relations $\mathcal{R}$ between pieces (\textit{e.g.}, tucked into, worn over).
Each layer $P_k$ contains optional upper and lower pieces (\textit{e.g.}, a tee with jeans, then a jacket over them).
We simulate the outfit in this inner-to-outer order, using each settled layer as a collider for the next.

With this enhanced representation, we build GarmentCodeData-XL, the first multi-layer sewing pattern dataset (Fig.~\ref{fig:garmentcodexl}).
It contains $15{,}205$ synthetic outfits with their programs, sewing patterns, draped meshes, and rendered views.
We use it for subsequent training and benchmarking.

\subsection{Stage 1: Layer Planning and Coarse Estimation}
\label{sec:stage1}

\begin{figure}[t]
  \centering
  \includegraphics[width=1.0\textwidth]{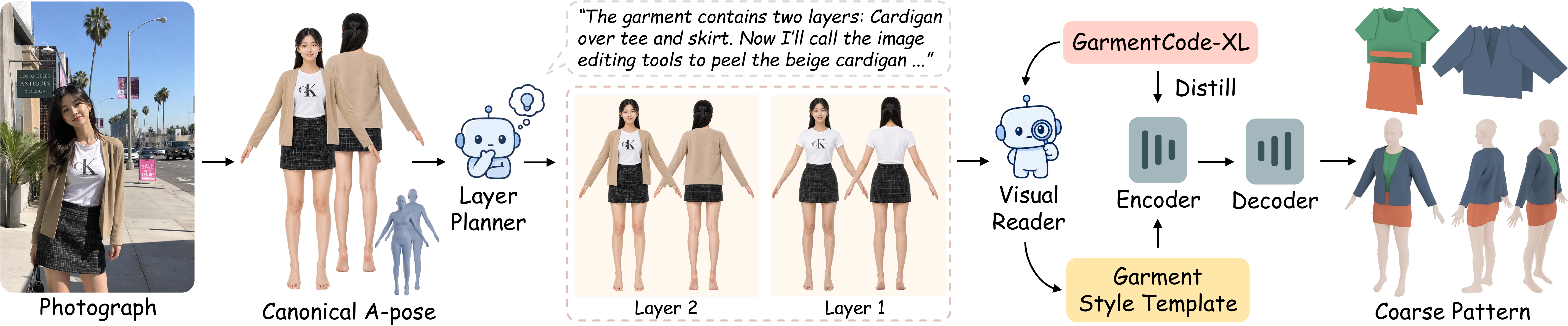}
  \caption{\textbf{Layer Planning and Coarse Pattern Estimation.} The Layer Planner reasons ordered layers and peels each into view. The Visual Reader generates a Garment Style Template per layer, then an encoder-decoder with distilled GarmentCode-XL semantics maps it to a coarse program.}
  \label{fig:stage1}
  \vspace{-10pt}
\end{figure}

Correct layer structure is the foundation of inverse garment design.
This stage thus explicitly models the layer structure and gives initial estimates per layer.
The workflow of Stage 1 is shown in Fig.~\ref{fig:stage1}.

\textbf{Canonical A-pose Views.}
Garments are hard to read in an in-the-wild photograph due to occlusion, unusual poses, or poor image quality.
We therefore canonicalize the photograph $I$ into an A-pose view, where garment structure is easier to read~\citep{garmentcodedata}.
We use an image editing model to generate the front view $I^{\mathrm{f}}=g_\text{canonical}(I)$, standardizing pose and background while preserving clothes.
The model also generates the back view $I^{\mathrm{b}}=g_\text{back}(I^{\mathrm{f}})$, keeping the same body pose and position.
From these views, we fit an SMPL-X~\citep{smplx} body with NLF~\citep{nlf} and replace its hands using Hand4Whole++~\citep{hand4whole} for finer finger geometry.
This fit provides body measurements $B$ for pattern drafting, a body mesh for simulation, and cameras $\pi^{\mathrm{f}},\pi^{\mathrm{b}}$ to align rendered garments with the canonical front and back views.

\textbf{Layer Planning.}
To support zero-shot recovery of arbitrary-layer outfits, we need the fully visible views per layer as supervision.
The \emph{Layer Planner} agent uses a carefully designed reasoning skill to infer the layer structure and relations $\mathcal{R}$.
It writes a description $r_k$ for each layer, then calls image editing tools to generate the views layer by layer:
\begin{equation}
  I^{\mathrm{f}}_K=I^{\mathrm{f}},\qquad I^{\mathrm{f}}_{k-1}=g_\text{peel}(I^{\mathrm{f}}_k,r_k)\ \ (k=K,\dots,2),\qquad I^{\mathrm{b}}_k=g_\text{back}(I^{\mathrm{f}}_k),
\end{equation}
where $I^{\mathrm{f}}_k$ and $I^{\mathrm{b}}_k$ are the front and back views in which layer $k$ is the outermost garment.
The tool $g_\text{peel}$ is an image editor that removes the layer described by $r_k$.

\textbf{Template Generation.}
With every layer now visible, we first try prompting a pre-trained VLM to write GarmentCode-XL programs directly.
However, the results are poor due to semantic misalignment~\citep{comparativestudydslcode}: the model struggles to interpret parameter names (\textit{e.g.}, \texttt{SkirtCircle}, \texttt{Skirt2}) and their bounds.
After extensive exploration, we find a surprisingly direct solution: let an LLM read the source code to formulate visual questions for a \emph{Visual Reader} agent to answer.
The code reveals the geometric meaning of each parameter, allowing the model to interpret it correctly.
A pre-trained LLM $g_\text{ques}$ generates this questionnaire, with one visual question $q_i$ about the garment property controlled by each parameter $d_i$:
\begin{equation}
  \mathcal{Q}=g_\text{ques}(\mathcal{F},\mathcal{D})=(q_1,\dots,q_{|\mathcal{D}|}).
\end{equation}
The Visual Reader answers $\mathcal{Q}$ for each layer using its views $I^{\mathrm{f}}_k,I^{\mathrm{b}}_k$ and description $r_k$.
We find it answers multiple choice questions more reliably than numerical questions.
We therefore give each numerical question discrete options such as \textit{``quarter length''}, \textit{``mid length''}, or \textit{``full length''}.
Its answers form a \emph{Garment Style Template} for each layer.

\textbf{Program Decoding.}
We append an \emph{encoder-decoder} to convert each template into an executable sewing pattern program.
We distill GarmentCode-XL semantics into an encoder $E$ through LoRA~\citep{lora}, using source code paired with visual descriptions and questions.
This allows $E$ to interpret visual answers in terms of the underlying program parameters (Appendix~\ref{appendix:s1_lora}).
Then, we encode each question with its selected answer and take the final hidden state, projected through an MLP, as the conditioning vector for a lightweight autoregressive decoder $\Psi$~\citep{gpt2}.
Trained on paired templates and programs, $\Psi$ uses the learned parameter distribution to decode each garment style template into a complete design configuration.
This configuration is a fixed sequence of $|\mathcal{D}|=131$ parameter tokens, quantized by type (Appendix~\ref{appendix:s1_arch}).
We decode one program $P^{0}_k$ per layer, then assemble these programs in dressing order into the coarse outfit.

\subsection{Stage 2: Fine-Grained Refinement}
\label{sec:stage2}

\begin{figure}[t]
  \centering
  \includegraphics[width=1.0\textwidth]{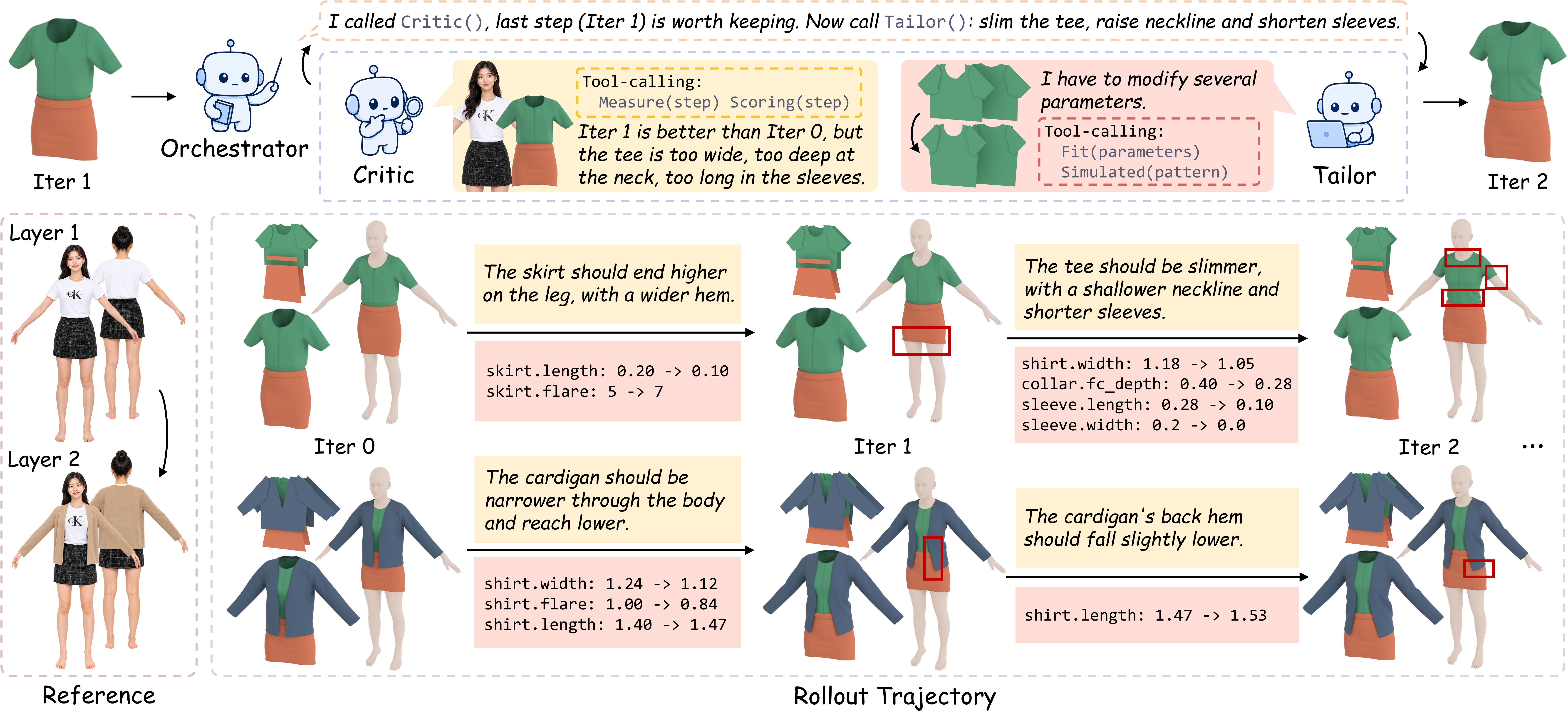}
  \caption{\textbf{Fine-Grained Refinement with an Agentic Trio.} \emph{Top}: illustration of one refinement round (from Iter 1 to Iter 2 of the inner layer). The Critic scores each garment and identifies discrepancies, the Orchestrator saves a checkpoint and instructs the Tailor, and the Tailor applies the requested edits and simulates the pattern. \emph{Bottom}: examples of rollout trajectories. For agent implementation, detailed prompts, and the complete record of this example, see Appendix~\ref{appendix:stage2}.}
  \label{fig:stage2}
  \vspace{-10pt}
\end{figure}

Building on the coarse garment programs from Stage 1, our goal here is to align their draped geometry with the reference views through closed-loop refinement.
We therefore design an agentic trio with complementary roles: a \emph{Tailor}, a \emph{Critic}, and an \emph{Orchestrator}.
This decomposition separates garment editing, evaluation, and control flow, enabling structured refinement while keeping each agent's responsibility simple.
It also reduces self-evaluation bias: an independent Critic is better positioned to identify errors that an edit-focused Tailor may overlook.

The \textbf{Tailor} proposes pattern edits using structured tools.
The \textbf{Critic} evaluates the simulated garment against the reference for construction, proportion, fit, layering, and physical feasibility, providing per-garment scores and natural-language feedback.
The \textbf{Orchestrator} coordinates this interaction, determining when to accept edits, request further refinement, or finish the current layer.
To prevent degradation during refinement, it saves pattern checkpoints and reverts to one when the Critic's scores worsen.
Refinement proceeds from inner to outer layers, as illustrated in Fig.~\ref{fig:stage2}.

\textbf{Agent Tools.}
Agents interact with the garment exclusively through structured tools:
\emph{Observation tools} (\textit{e.g.}, \texttt{Render}, \texttt{Measure}) render the simulated garment beside the reference on a shared body ruler and measure every garment edge as a percentage of body height, providing complementary visual and metric evidence for judgement.
\emph{Modification tools} (\textit{e.g.}, \texttt{Fit}, \texttt{Simulate}) edit the sewing pattern and call the cloth solver to simulate it.
\emph{Control tools} (\textit{e.g.}, \texttt{Rollback}, \texttt{Finish}) manage global control operations.
Tool access is role-dependent: the Critic is restricted to observation tools, the Tailor uses observation and modification tools, and the Orchestrator invokes the Tailor and Critic as tools while managing global control.
The complete tool interface is provided in Appendix~\ref{appendix:s2_tools}.

\textbf{Closed-Loop Refinement.}
Let $P^{t}$ denote the sewing pattern program of the current layer at iteration $t$, $M^{t}=\mathrm{Drape}(P^{t},B)$ the simulated result, and $O^{t}$ its observation through the two cameras together with the measurements. Given the reference views $I^{\mathrm{f}}_k$, $I^{\mathrm{b}}_k$ and $I$, the Critic returns
\begin{equation}
  (s^{t},F^{t},r^{t}_{\text{critic}})=\Phi_{\text{critic}}(I^{\mathrm{f}}_k,I^{\mathrm{b}}_k,I,O^{t},O^{t-1}),
  \label{eq:critic}
\end{equation}
where $s^{t}$ is the per-garment scores, $F^{t}$ is the list of discrepancies with their measured evidence, and $r^{t}_{\text{critic}}$ is the rationale for the scores, which also states whether this step improves on the previous one. The Orchestrator reads the history $H^{t}$ of all checkpoints and decides
\begin{equation}
  (a^{t},c^{t},r^{t}_{\text{orch}})=\Phi_{\text{orch}}(H^{t},s^{t},F^{t}),
  \label{eq:orchestrator}
\end{equation}
where $a^{t}$ is the action (chosen from \texttt{Continue}, \texttt{Rollback} and \texttt{Finish}), $c^{t}$ is the instruction to the Tailor, which names the discrepancies to address in this round, and $r^{t}_{\text{orch}}$ is the rationale for the action. The Tailor then produces the edit
\begin{equation}
  (u^{t},r^{t}_{\text{tailor}})=\Phi_{\text{tailor}}(c^{t},F^{t},O^{t},H^{t}),\qquad P^{t+1}=P^{t}\oplus u^{t},
  \label{eq:tailor}
\end{equation}
where $u^{t}$ is the edit and $r^{t}_{\text{tailor}}$ is the rationale for the edit. The solver then simulates $P^{t+1}$, and the next iteration begins. We note that the edit takes two forms: a construction edit switches a categorical choice such as the neckline (\textit{e.g.}, crew neck or V-neck) or the garment structure, which resolves the ambiguity the coarse estimate might leave. A size edit tunes the numeric parameters so that the garment matches the reference precisely. The complete procedure is shown in Algorithm~\ref{alg:stage2}.

\begin{algorithm}[t]
\caption{Fine-grained refinement of one dressing layer.}
\label{alg:stage2}
\begin{algorithmic}[1]
\algtext*{EndIf}
\Require coarse pattern $P^{0}$ of the current layer, body $B$, reference views $I^{\mathrm{f}}_k,I^{\mathrm{b}}_k,I$, budget $T$
\State $M^{0}\gets\mathrm{Drape}(P^{0},B)$;\quad $(s^{0},F^{0},\cdot)\gets\Phi_{\text{critic}}(I^{\mathrm{f}}_k,I^{\mathrm{b}}_k,I,O^{0})$;\quad $H^{0}\gets\{0\}$
\For{$t=0,1,\dots,T-1$}
  \State $(a^{t},c^{t},\cdot)\gets\Phi_{\text{orch}}(H^{t},s^{t},F^{t})$ \Comment{Chooses \texttt{Continue}, \texttt{Rollback}, or \texttt{Finish}}
  \If{$a^{t}=\texttt{Finish}$} \textbf{break} \EndIf
  \If{$a^{t}=\texttt{Rollback}$ to $j\in H^{t}$} $(P^{t},O^{t},s^{t},F^{t})\gets(P^{j},O^{j},s^{j},F^{j})$ \EndIf
  \State $(u^{t},\cdot)\gets\Phi_{\text{tailor}}(c^{t},F^{t},O^{t},H^{t})$ \Comment{Proposes a construction or size edit}
  \State $P^{t+1}\gets P^{t}\oplus u^{t}$;\quad $M^{t+1}\gets\mathrm{Drape}(P^{t+1},B)$
  \State $(s^{t+1},F^{t+1},\cdot)\gets\Phi_{\text{critic}}(I^{\mathrm{f}}_k,I^{\mathrm{b}}_k,I,O^{t+1},O^{t})$ \Comment{Judges with visual and metric evidence}
  \State $H^{t+1}\gets H^{t}\cup\{t+1\}$ \Comment{Context memory update}
\EndFor
\State \Return the best-scoring checkpoint in $H^{T}$
\end{algorithmic}
\end{algorithm}

\textbf{Evolving Memory for Convergent Fitting.}
To support convergence, we design a persistent, structured memory $H^{t}$ that retains context across iterations.
Each iteration archives sewing patterns, grounded observations, critiques, and all rationales.
To bound context growth, views older than two iterations are compacted into summaries.
Note that \texttt{Rollback} restores only the pattern and leaves the memory intact, so a failed attempt serves as a reference for later edits.
As shown in Fig.~\ref{fig:stage2}, referencing past pattern edits alongside their visual and measured errors empowers the agents to refine progressively and converge on the target geometry.
For instance, to land a cardigan hem on the reference, the Tailor actively contrasts prior length parameters (1.40 vs. 1.47) to deduce the precise value (1.53).
Without this memory, each edit degrades into an isolated, single-step guess, and the loop oscillates instead of converging on precise parameters.

\subsection{Post-Processing}
\label{sec:postprocess}

\begin{figure}[t]
  \centering
  \includegraphics[width=1.0\textwidth]{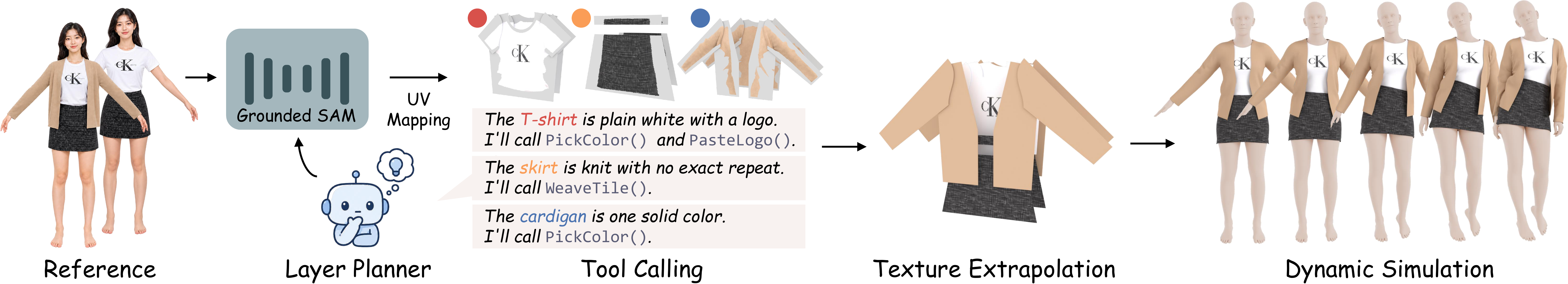}
  \vspace{-18pt}
  \caption{\textbf{Texture Generation and Simulation.} We design an agentic generative approach to recover textures for the sewing patterns, and use an XPBD-based cloth solver for dynamic simulation.}
  \label{fig:post_processing}
  \vspace{-14pt}
\end{figure}

\textbf{Texture Generation.}
To meet the needs of the garment fabrication industry, the final step of our method is to recover high-fidelity texture for every sewing pattern.
Unlike direct UV mapping which can introduce boundary artifacts, we design an agentic generative approach (Fig.~\ref{fig:post_processing}).
First, we use Grounded SAM~\citep{groundedsam} to segment each garment with the Layer Planner's descriptions, and map the pixels to its sewing pattern.
We then divide common garment textures into five classes and design a dedicated texture extrapolation tool for each:
\texttt{PickColor} reads a robust color for solid fabrics.
\texttt{WeaveTile} turns an observed swatch into seamless knitted or woven texture with FabricDiffusion~\citep{fabricdiffusion}.
\texttt{RepeatPattern} extends the measured repeat of regular motifs (\textit{e.g.}, plaid, stripes) across the panels.
\texttt{PasteLogo} pastes the logo back at its exact position, composable with any other tool.
\texttt{ContinuePrint} handles all remaining prints that only admit reconstruction as a whole, keeping observed regions and generatively continuing them into unobserved cloth.
Finally, we reuse the Layer Planner to classify each garment's texture and call the corresponding tools.
See Appendix~\ref{appendix:texture_generation} for detailed tool implementations.

\textbf{Dynamic Simulation.}
We use an XPBD solver~\citep{xpbd} on Warp~\citep{warp} to stitch the seams and drape each layer under gravity and frictional contact with the kinematic SMPL-X body.
To repose, we interpolate the body to the target pose as a moving boundary condition, advect the cloth by its nearest body points under gravity and contact, and reuse each settled inner layer as a moving collider for the next (Fig.~\ref{fig:post_processing}).

%% file: section/experiment.tex
\section{Experiments}

\subsection{Implementation Details}

\textbf{Setup.}
We use GPT-5.6-luna~\citep{gpt56} with medium reasoning effort for all agents and $g_\text{ques}$.
The image tools use GPT-Image-2~\citep{gptimage2}.
The encoder $E$ is a fine-tuned Qwen3.6-27B~\citep{qwen36}, and the decoder $\Psi$ is trained the official training split of GarmentCodeData and the training split of our GarmentCodeData-XL.
Full implementation details are in the Appendix.

\textbf{Baselines.}
We compare with state-of-the-art sewing pattern reconstruction methods: Feed-forward methods include SewFormer~\citep{sewformer}, DressCode~\citep{dresscode}, ChatGarment~\citep{chatgarment}, and GarmentGPT~\citep{garmentgpt}.
Differentiable optimization methods include ISP~\citep{isp}, GarmentRecovery~\citep{garmentrecovery}, and Dress-1-to-3~\citep{dress123}.

\subsection{Results}
\label{sec:results}

\begin{figure}[t]
  \centering
  \includegraphics[width=1.0\textwidth]{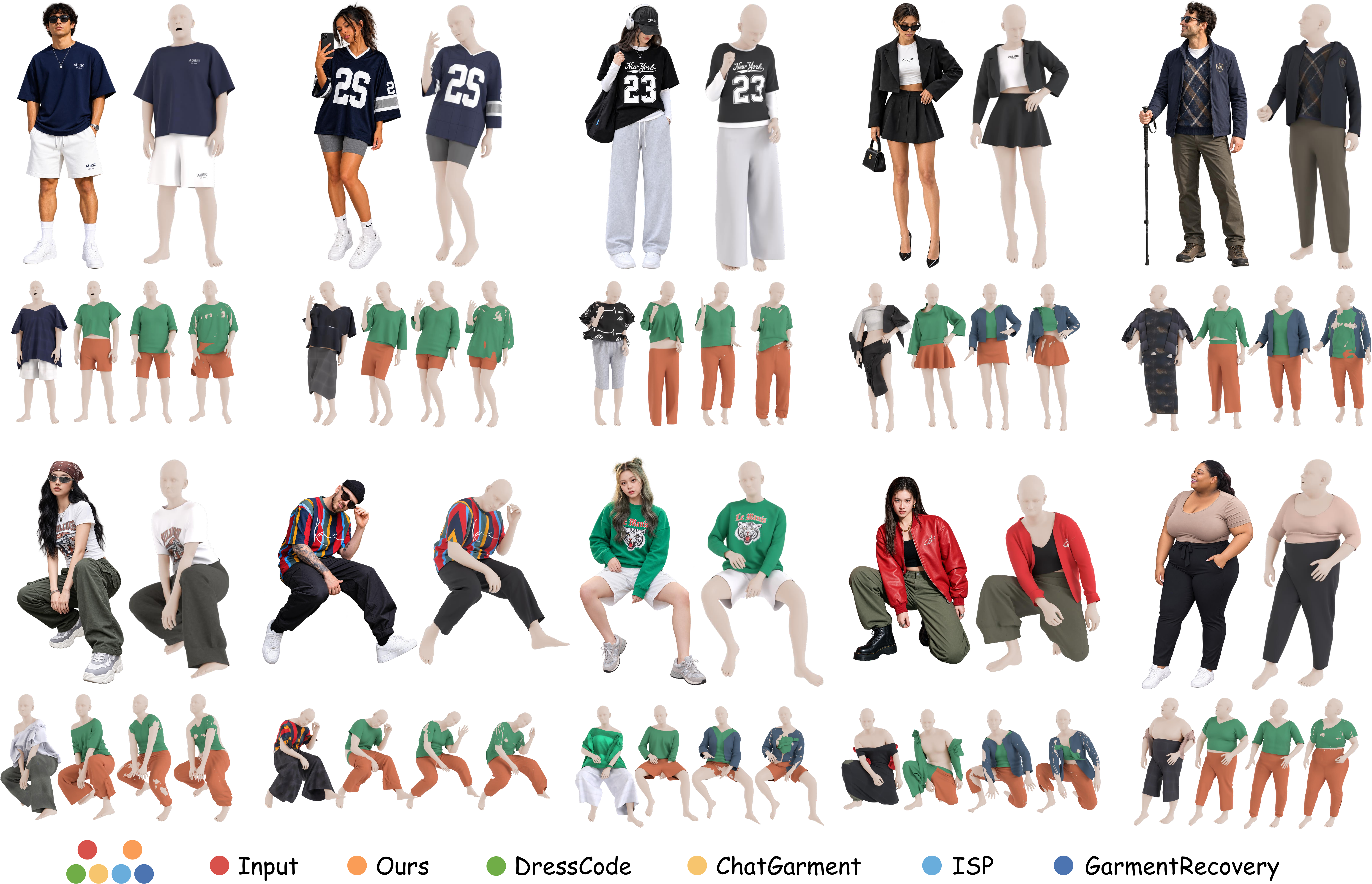}
  \vspace{-10pt}
  \caption{\textbf{Qualitative comparison on in-the-wild images.} \emph{Top five}: reconstructing outfits with one to three layers (left to right). \emph{Bottom five}: generalization to diverse poses and body shapes. ChatGarment, ISP, and GarmentRecovery lack texture generation, so we show only their geometry.}
  \label{fig:comparison}
\end{figure}

\textbf{Comparison to baselines.}
We report the \emph{structure accuracy} of the sewing pattern, measured by accuracy for the number of panels, edges, and stitches on the GarmentCodeData~\citep{garmentcodedata} test split, and the \emph{geometry accuracy} of the draped garment, measured by Chamfer Distance (CD) and Intersection over Union (IoU) on CloSe~\citep{close} and 4D-Dress~\citep{4ddress}.
The results are shown in Table~\ref{tab:comparison}.
Our method outperforms feed-forward methods by a large margin, and surpasses or matches differentiable optimization in far less time.
The gain comes from our enhanced GarmentCode-XL representation, per-layer planning and supervision, and closed-loop refinement by our agentic trio.
Fig.~\ref{fig:comparison} shows visual comparisons.
DressCode produces coarse textures, GarmentRecovery often interpenetrates the body in simulation, ChatGarment and ISP recover plausible shapes but misjudge the number of layers, since neither models layer planning explicitly.
In contrast, our method stays robust across complex layering, diverse poses, and body shapes.

\begin{table}[t]
  \caption{\textbf{Quantitative comparison.} We report the structure accuracy of sewing patterns and the geometry accuracy of draped garments. ISP and GarmentRecovery are excluded from structure accuracy, as they do not produce explicit sewing patterns with panel geometry. Our method achieves the best quality on most metrics, and is significantly faster than differentiable optimization.}
  \label{tab:comparison}
  \vspace{-5pt}
  \resizebox{\textwidth}{!}{
  \begin{tabular}{@{}lcccccccc@{}}
  \toprule
   & \multicolumn{3}{c}{\textsc{Structure Acc.}~$\uparrow$} & \multicolumn{2}{c}{\textsc{Geometry (CloSe)}} & \multicolumn{2}{c}{\textsc{Geometry (4D-Dress)}} & \cellcolor{gray!8} \\ \cmidrule(lr){2-4} \cmidrule(lr){5-6} \cmidrule(lr){7-8}
  \multirow{-2}{*}{\textsc{Method}} & \textsc{Panel} & \textsc{Edge} & \textsc{Stitch} & \textsc{CD}~$\downarrow$ & \textsc{IoU}~$\uparrow$ & \textsc{CD}~$\downarrow$ & \textsc{IoU}~$\uparrow$ & \multirow{-2}{*}{\cellcolor{gray!8}\begin{tabular}[c]{@{}c@{}}\textsc{Avg. Time}~$\downarrow$\\\textsc{(min)}\end{tabular}} \\ \midrule
  SewFormer & 18.67 & 28.80 & 7.24 & 7.40 & 0.348 & 7.58 & 0.496 & \cellcolor{gray!8}2.0 \\
  ChatGarment & 51.84 & 57.15 & 70.18 & 2.39 & 0.580 & 1.94 & 0.763 & \cellcolor{gray!8}4.2 \\
  GarmentGPT & 43.62 & 38.54 & 65.43 & 5.50 & 0.381 & 10.46 & 0.204 & \cellcolor{gray!8}0.4 \\
  DressCode & 19.67 & 26.47 & 30.97 & 2.77 & 0.542 & 2.56 & 0.685 & \cellcolor{gray!8}2.2 \\ \midrule
  ISP & - & - & - & 1.79 & 0.687 & \textbf{1.16} & 0.859 & \cellcolor{gray!8}90.0 \\
  GarmentRecovery & - & - & - & 1.77 & 0.740 & 1.51 & 0.865 & \cellcolor{gray!8}25.0 \\
  Dress 1-to-3 & 17.67 & 31.07 & 26.14 & 1.63 & 0.652 & 2.05 & 0.617 & \cellcolor{gray!8}176.0 \\ \midrule
  \rowcolor{blue!10}
  Ours & \textbf{87.36} & \textbf{76.56} & \textbf{88.90} & \textbf{1.20} & \textbf{0.764} & 1.21 & \textbf{0.880} & 8.6 \\ \bottomrule
  \end{tabular}
  }
  \vspace{-5pt}
\end{table}

\begin{figure}[t]
  \captionsetup{skip=5pt}
  \begin{minipage}[t]{0.405\textwidth}
    \vspace{0pt}
    \includegraphics[width=\linewidth]{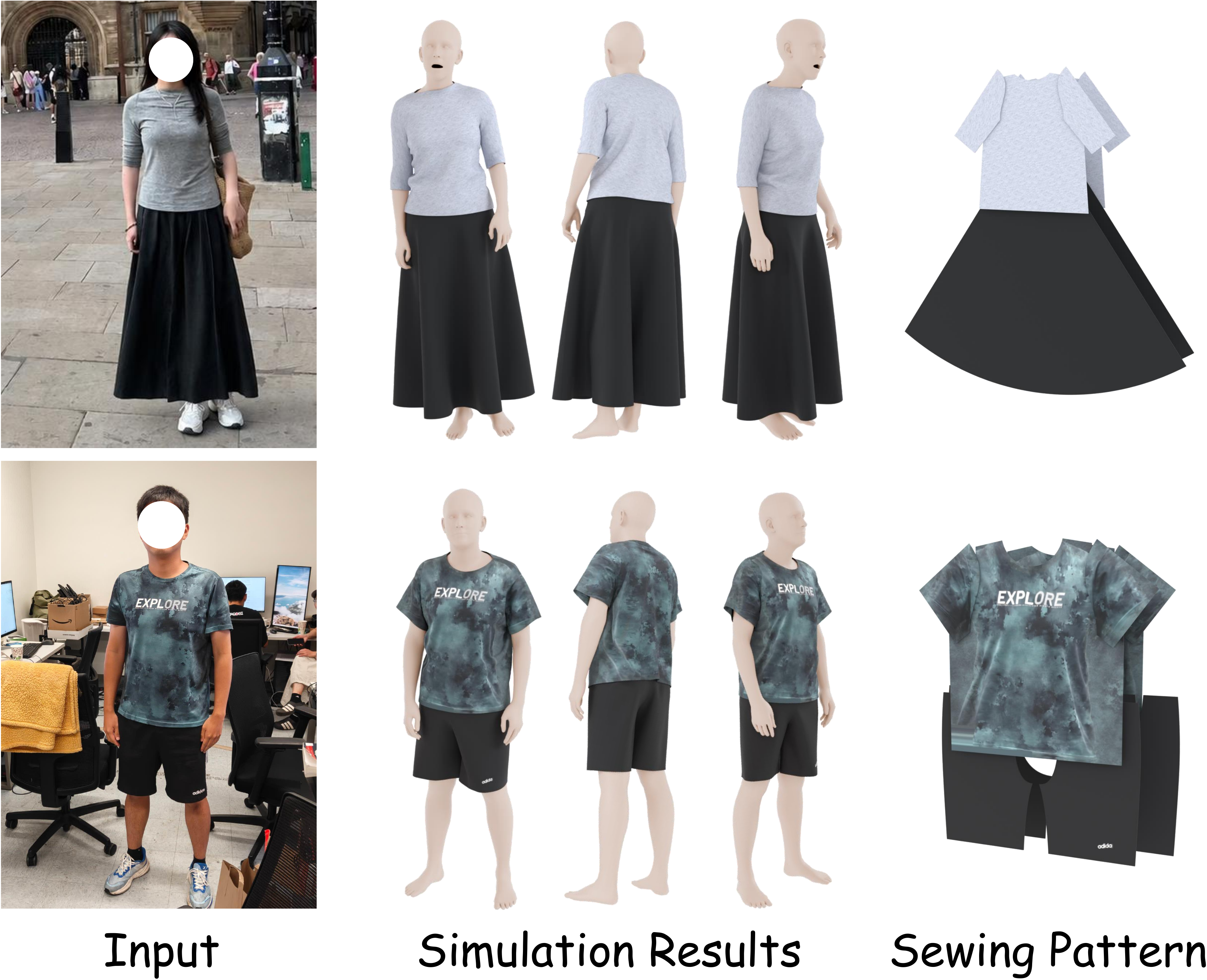}
    \captionof{figure}{\textbf{User-uploaded photo results.}}
    \label{fig:user_visual}
  \end{minipage}
  \hfill
  \begin{minipage}[t]{0.565\textwidth}
    \vspace{0pt}
    \captionof{table}{\textbf{User study.} We report the percentage of votes as the best match to the photograph for geometry, and the average rating from 1 to 5 for plausibility and texture.}
    \label{tab:userstudy}
    \centering
    \setlength{\tabcolsep}{1.5pt}
    \resizebox{\linewidth}{!}{
    \begin{tabular}{@{}lccc@{}}
    \toprule
    \textsc{Method} & \textsc{Geometry}~$\uparrow$ & \textsc{Plausibility}~$\uparrow$ & \textsc{Texture}~$\uparrow$ \\ \midrule
    SewFormer & 0.00\% & 1.68 & 2.86 \\
    ChatGarment & 14.50\% & 3.61 & - \\
    GarmentGPT & 0.00\% & 1.24 & - \\
    DressCode & 9.50\% & 2.69 & 2.57 \\ \midrule
    ISP & 9.00\% & 4.43 & - \\
    GarmentRecovery & 8.50\% & 3.02 & - \\
    Dress 1-to-3 & 4.50\% & 3.17 & 3.42 \\ \midrule
    \rowcolor{blue!10}
    Ours & \textbf{54.00\%} & \textbf{4.51} & \textbf{4.33} \\ \bottomrule
    \end{tabular}
    }
  \end{minipage}
\end{figure}

\textbf{User study.}
To capture perceptual quality that metrics cannot measure, we conduct a user study on 200 cases, mixing cases from our experiments, in-the-wild photos randomly sampled from Pinterest\footnote{\url{https://www.pinterest.com}}, and photos uploaded in advance by users (Fig.~\ref{fig:user_visual}).
For each case, participants pick the result that best matches the photo in geometry, and rate the simulation plausibility of each garment on the body and the fidelity of its texture from 1 to 5.
We hire 65 participants, and Table~\ref{tab:userstudy} shows the results.
Our method leads by a large margin on all three criteria, indicating strong perceptual quality in geometric reconstruction, simulation plausibility, and texture fidelity.

\subsection{Ablation Study}

\begin{figure}[t]
  \captionsetup{skip=5pt}
  \begin{minipage}[t]{0.35\textwidth}
    \vspace{0pt}
    \includegraphics[width=\linewidth]{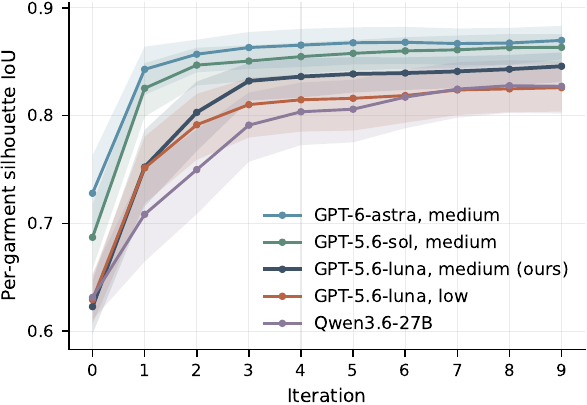}
    \captionof{figure}{\textbf{Ablation of foundation models and iteration budget.} Iteration 0 is the coarse estimate, and the bands show the 95\% bootstrap confidence interval.}
    \label{fig:ablation_models}
  \end{minipage}
  \hfill
  \begin{minipage}[t]{0.62\textwidth}
    \vspace{0pt}
    \includegraphics[width=\linewidth]{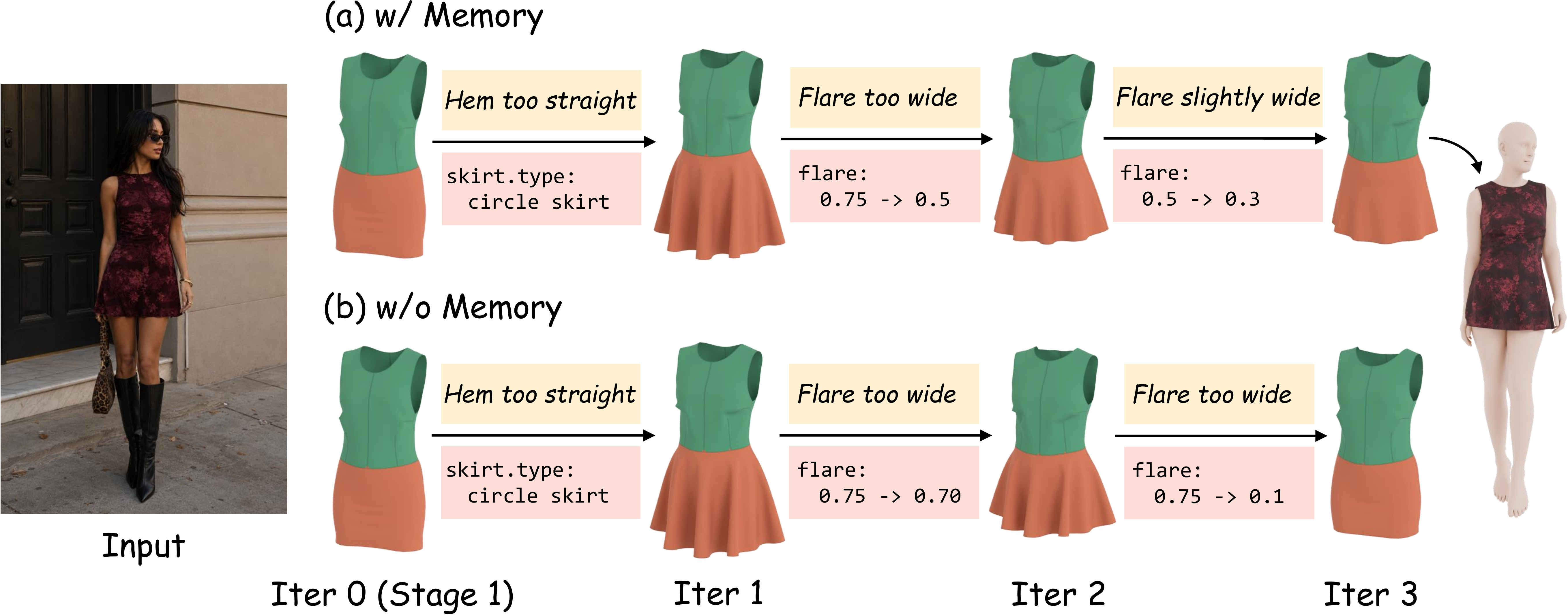}
    \captionof{figure}{\textbf{Ablation on evolving memory.} We show how the flare of the skirt is adjusted over iterations. With memory, the agents keep the measurements of earlier steps as reference and converge progressively. Without it, the agents suffer from severe context loss and oscillate.}
    \label{fig:ablation_memory}
  \end{minipage}
  \vspace{-10pt}
\end{figure}

\textbf{Foundation Models and Iteration Budget.}
We benchmark different model backbones on 200 cases from our GarmentCodeData-XL test set, as shown in Fig.~\ref{fig:ablation_models}.
The coarse estimate varies widely across models, but the refinement loop closes most of the gap and converges after about 6 iterations.
To balance quality and cost, we adopt GPT-5.6-luna (medium) with a budget of $T=6$ as default.

\textbf{Evolving Context Memory.}
Fig.~\ref{fig:ablation_memory} demonstrates the necessity of context memory.
With the measured effect of every earlier edit in context, the agents align the skirt flare onto the target progressively.
Without it, each edit degrades into an isolated guess and the skirt flare oscillates.

\setlength{\intextsep}{0pt}
\begin{wraptable}{r}{0.5\textwidth}
  \centering
  \begingroup
  \captionsetup{skip=4pt}
  \caption{\textbf{Ablation on representation and the agentic trio.} DeepFashion2 IoU is measured against the annotated clothing masks.}
  \label{tab:ablation}
  \endgroup
  \setlength{\tabcolsep}{1pt}
  \resizebox{\linewidth}{!}{
  \begin{tabular}{@{}lccc@{}}
  \toprule
   & \multicolumn{2}{c}{\textsc{GarmentCodeData-XL}} & \textsc{DeepFashion2} \\ \cmidrule(l){2-3} \cmidrule(l){4-4}
  \multirow{-2}{*}{\textsc{Comparison}} & \textsc{Struct. Acc.}~$\uparrow$ & \textsc{IoU}~$\uparrow$ & \textsc{IoU}~$\uparrow$ \\ \midrule
  w/o Refinement & 63.25 & 0.623 & 0.562 \\ \midrule
  w/o GarmentCode-XL & 57.99 & 0.574 & 0.650 \\
  w/o Critic & 69.56 & 0.822 & 0.765 \\
  w/o Measure & 67.21 & 0.804 & 0.772 \\ \midrule
  \rowcolor{blue!10}
  Full & \textbf{73.59} & \textbf{0.839} & \textbf{0.783} \\ \bottomrule
  \end{tabular}
  }
\end{wraptable}

\textbf{Representation and Agentic Trio.}
We ablate the refinement stage, GarmentCode-XL, the Critic, and the Measure tool on 200 synthetic data from our GarmentCodeData-XL test set and 200 in-the-wild photos from DeepFashion2~\citep{DeepFashion2} (Table~\ref{tab:ablation}).
Without refinement, the coarse estimate is a one-pass predictor trained on our data, and the closed loop lifts it by a large margin on both synthetic and real photos.
Beyond refinement, reverting to the original GarmentCode costs the most, as it cannot recover a style outside the design space.
Without the Critic, the Tailor falls into self-bias~\citep{selfbias,selfpreference,selfcorrect}, overrating flawed proposals.
Without Measure, the agents lack the measured error to reason out a precise edit, so accuracy drops slightly.

\textbf{Others.}
For more experiments (\textit{e.g.}, quantitative texture comparison, additional visual results) and ablations (\textit{e.g.}, rollback, texture generation), see Appendix~\ref{appendix:additional_experiments} and Appendix~\ref{appendix:ablation_study}, respectively.

\setlength{\intextsep}{12pt plus 2pt minus 2pt}%

%% file: section/conclusion.tex
\section{Conclusion}

We present \ours, a hierarchical agentic framework for simulation-ready inverse garment design from a single in-the-wild image.
We first introduce GarmentCode-XL, an enhanced representation for diverse styles and arbitrary layers, and GarmentCodeData-XL, the first multi-layer sewing pattern dataset with over 15K outfits.
For coarse estimation, a Layer Planner decomposes the outfit into ordered layers and a Visual Reader estimates each layer.
For fine refinement, an agentic trio of Tailor, Critic, and Orchestrator edits each pattern in a closed loop with evolving memory.
We further generate high-fidelity textures with agentic tools.
Experiments demonstrate the best overall accuracy and preference, generalization far beyond feed-forward methods, and far less time than differentiable optimization.
We hope this work advances the garment industry and inverse graphics.

%% file: section/appendix.tex
\begingroup
\section*{Contents}
\setlength{\parskip}{0pt}
\tocsec{appendix:notation}{Notations}
\tocsec{appendix:garmentcodexl}{Garment Representation and Dataset}
\tocsec{appendix:stage1}{Stage 1: Layer Planning and Coarse Estimation}
\tocsub{appendix:s1_workflow}{Agent implementation and interaction}
\tocsub{appendix:s1_canonical}{Canonical A-pose views}
\tocsub{appendix:s1_structure}{Layer planning}
\tocsub{appendix:s1_ques}{Questionnaire synthesis}
\tocsub{appendix:s1_read}{Template generation with the Visual Reader}
\tocsub{appendix:s1_lora}{Encoder-decoder from template to program}
\tocsec{appendix:stage2}{Stage 2: Fine-Grained Refinement}
\tocsub{appendix:s2_workflow}{Agent implementation and interaction}
\tocsub{appendix:s2_tools}{Tool interface}
\tocsub{appendix:s2_tailor}{Program editing with the Tailor}
\tocsub{appendix:s2_fit}{Size fitting with the Fit tool}
\tocsub{appendix:s2_critic}{Drape evaluation with the Critic}
\tocsub{appendix:s2_orch}{Session control with the Orchestrator}
\tocsub{appendix:s2_round}{An example refinement round}
\tocsec{appendix:texture_generation}{Texture Generation}
\tocsub{appendix:tex_route}{Texture routing}
\tocsub{appendix:tex_tools}{Tool implementation}
\tocsec{appendix:additional_experiments}{Additional Experiments}
\tocsub{appendix:baseline_protocol}{Baseline implementation}
\tocsub{appendix:texture_comparison}{Quantitative comparison on texture generation}
\tocsub{appendix:layer_comparison}{Quantitative comparison on layer structure}
\tocsub{appendix:view_faithfulness}{Faithfulness of synthesized views}
\tocsub{appendix:stress_tshirt}{Stress test: refinement from a plain T-shirt}
\tocsub{appendix:more_results}{Additional visual results}
\tocsec{appendix:ablation_study}{Ablation Study}
\tocsub{appendix:ablation_views}{Ablation on synthesized views}
\tocsub{appendix:ablation_trio}{Ablation on agentic trio design}
\tocsub{appendix:ablation_single}{Ablation on single-agent refinement}
\tocsub{appendix:ablation_texture}{Ablation on texture generation}
\tocsec{appendix:user_study}{User Study}
\tocsec{appendix:limitations}{Limitations and Future Work}
\endgroup

\section{Notations}
\label{appendix:notation}

Table~\ref{tab:notation} summarizes the main symbols used throughout this paper.

\begin{table}[p]
\caption{\textbf{Notations.} Summary of the main symbols and notations used in the paper.}
\label{tab:notation}
\begin{center}
\footnotesize
\setlength{\tabcolsep}{5pt}
\renewcommand{\arraystretch}{1.0}
\begin{tabularx}{\textwidth}{@{}l Y@{}}
\toprule
\textbf{Symbol} & \textbf{Meaning} \\
\midrule
\multicolumn{2}{@{}l}{\textit{Garment Representation}} \\
$S=P(B)$ & Sewing pattern of one garment: its flat panels and the stitches that join them, drafted by $P$ on body $B$. \\
$P=\langle\mathcal{F},\mathcal{D}\rangle$ & Garment program: panel constructors with their design parameters. \\
$\mathcal{F}=\{f_i\}$ & Panel constructors, one per component such as a sleeve or a bodice. \\
$f_i$ & The $i$-th constructor, drafting a component as parametric 2D curves controlled by $d_i$ and $B$. \\
$\mathcal{D}=\{d_i\}$ & Design parameters, $131$ in GarmentCode-XL. \\
$d_i$ & The $i$-th design parameter. \\
$B$ & Body measurements from the SMPL-X fit, used to draft every panel. \\
$M=\mathrm{Drape}(P,B)$ & Garment mesh draped by the cloth solver from $P$ and $B$. \\
$\mathcal{P}=\langle(P_1,\dots,P_K),\mathcal{R}\rangle$ & Outfit: an ordered stack of layer programs and their relations. \\
$K$ & Number of layers, indexed from $1$ (innermost) to $K$ (outermost). \\
$\mathcal{R}$ & Relations among pieces, such as tucked into or worn over. \\
$P_k$ & Program of layer $k$, an optional upper piece with an optional lower piece. \\
$P^{0}_k$, $\mathcal{P}^{0}$ & Coarse program of layer $k$ from Stage 1, and the coarse outfit they assemble. \\
\midrule
\multicolumn{2}{@{}l}{\textit{Canonical Views and Layer Planning}} \\
$I$ & In-the-wild input photograph, with arbitrary pose, background, and quality. \\
$I^{\mathrm{f}},I^{\mathrm{b}}$ & Canonical A-pose front and back views of the same person. \\
$g_\text{canonical}$, $g_\text{back}$ & Image editing models that canonicalize $I$ into the front view and generate its back view. \\
$g_\text{peel}$ & Image editing tool that removes the garment described by $r_k$ and completes what it covered. \\
$r_k$ & Natural language description of layer $k$, written by the Layer Planner. \\
$I^{\mathrm{f}}_k,I^{\mathrm{b}}_k$ & Front and back views in which layer $k$ is the outermost garment. \\
$\pi^{\mathrm{f}},\pi^{\mathrm{b}}$ & Cameras that align a render of the drape with the two views. \\
\midrule
\multicolumn{2}{@{}l}{\textit{Template Generation}} \\
$g_\text{ques}$ & Pre-trained LLM that reads the GarmentCode-XL source once and writes $\mathcal{Q}$. \\
$\mathcal{Q}=(q_1,\dots,q_{|\mathcal{D}|})$ & Questionnaire, one multiple choice question per design parameter. \\
$q_i$ & Visual question about the garment property controlled by $d_i$, with a closed list of options. \\
$\tau_i$ & The option the Visual Reader selects for $q_i$. \\
$\text{GST}_k$ & Garment Style Template of layer $k$, the answered questionnaire of that layer. \\
$\mathcal{V}(\tau_i)$ & Normalized interval of parameter values covered by option $\tau_i$. \\
\midrule
\multicolumn{2}{@{}l}{\textit{Program Decoding}} \\
$E$ & Encoder, a VLM with distilled GarmentCode-XL semantics, embedding each question with its answer. \\
$\mathcal{X}$ & Program and description pairs that distill GarmentCode-XL semantics into $E$. \\
$\Psi$ & Autoregressive decoder that turns $\text{GST}_k$ into the design parameters of $P^{0}_k$. \\
$\mathbf{Q}$ & Type-based quantizer of Eq.~\ref{eq:quantize}. \\
$t_i=\mathbf{Q}(d_i)$, $\mathcal{T}$ & Token of the $i$-th parameter, and the token sequence of one program. \\
$\mathrm{Norm}(d_i)$, $\mathrm{Enum}(d_i)$ & $d_i$ mapped to $[0,1]$ over its legal range, and the index of a categorical option. \\
$\lambda=100$ & Scale of the continuous quantizer, one step per percent of the legal range. \\
\midrule
\multicolumn{2}{@{}l}{\textit{Fine-Grained Refinement}} \\
$\Phi_{\mathrm{tailor}},\Phi_{\mathrm{critic}},\Phi_{\mathrm{orch}}$ & The agentic trio: the Tailor edits, the Critic judges, the Orchestrator controls. \\
$t$, $T$ & Iteration index within one layer, and the simulation budget of that layer. \\
$P^{t}$ & Program of the current layer at iteration $t$, starting from the coarse program $P^{0}$. \\
$M^{t}$ & Simulated mesh $\mathrm{Drape}(P^{t},B)$ at iteration $t$. \\
$O^{t}$ & Observation of $M^{t}$: renders through $\pi^{\mathrm{f}},\pi^{\mathrm{b}}$ together with the measurements. \\
$s^{t}$ & Per-garment scores on construction, proportion, fit, layering, and physical feasibility. \\
$F^{t}$ & Discrepancies against the reference views, each with its measured evidence. \\
$a^{t}$ & Action of the Orchestrator: \texttt{Continue}, \texttt{Rollback}, or \texttt{Finish}. \\
$c^{t}$ & Instruction to the Tailor, naming the discrepancies to address in this round. \\
$u^{t}$ & Edit of the Tailor, applied as $P^{t+1}=P^{t}\oplus u^{t}$. \\
$r^{t}_{\mathrm{critic}},r^{t}_{\mathrm{orch}},r^{t}_{\mathrm{tailor}}$ & Rationales the three agents attach to their outputs. \\
$H^{t}$ & Evolving memory: every checkpoint with its pattern, observation, and critique. \\
\bottomrule
\end{tabularx}
\end{center}
\end{table}

\section{Garment Representation and Dataset}
\label{appendix:garmentcodexl}

\begin{figure}[t]
  \centering
  \includegraphics[width=\textwidth]{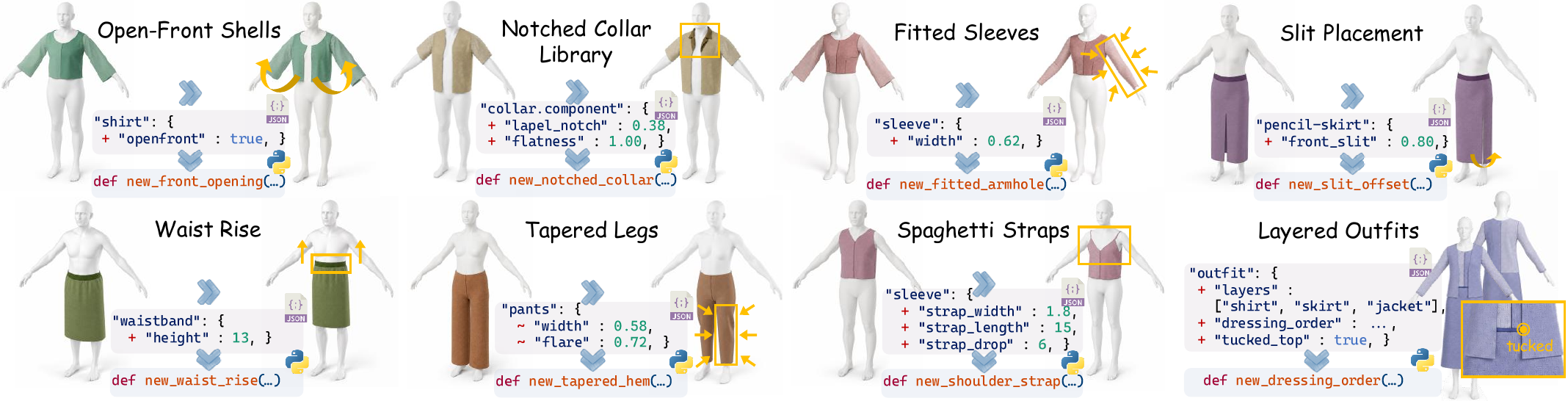}
  \caption{\textbf{Examples of Construction details of GarmentCode-XL.} New panel constructors and outfit fields are marked $+$, and expanded ranges of existing parameters are marked $\sim$. The first seven widen the design space to real-world styles, and the last adds the layered outfit structure with dressing order and relations.}
  \label{fig:appendix_garmentcodexl_construction}
\end{figure}

Since garments can be written as code~\citep{garmentcode}, our key insight is to solve inverse garment design the way agents already excel at coding: writing code and refining it against execution feedback, therefore turning the inverse garment design problem into a coding task.
The ability of such an agent, however, is bounded by the expressiveness of its programming language.
GarmentCode is the representative language, where a compact program compiles to a sewing pattern and its draped mesh, but it covers a limited set of garment styles and cannot represent multi-layer outfits with arbitrary structures and explicit relations.
To unleash the potential of coding agents on real-world outfits, we therefore first introduce a more expressive language.
GarmentCode-XL extends GarmentCode along two axes: a broader garment design space and a structured representation of layered outfits.
Figure~\ref{fig:appendix_garmentcodexl_construction} shows examples of how we construct both extensions through new panel constructors, expanded parameter ranges, and outfit-level fields.

For layered outfits, GarmentCode-XL represents an outfit as $\mathcal{P}=\langle(P_1,\ldots,P_K),\mathcal{R}\rangle$.
Each $P_k$ defines one dressing layer and may contain an upper piece, a lower piece, or both.
The ordered sequence $(P_1,\ldots,P_K)$ runs from the innermost layer to the outermost.
The relation set $\mathcal{R}$ records typed relations within and across layers (\textit{e.g.}, tucked into, worn over).
This structure also determines the simulation order: we drape the layers from inner to outer, and each settled layer becomes a collider for the next.

\begin{figure}[!ht]
  \centering
  \includegraphics[width=0.6\textwidth]{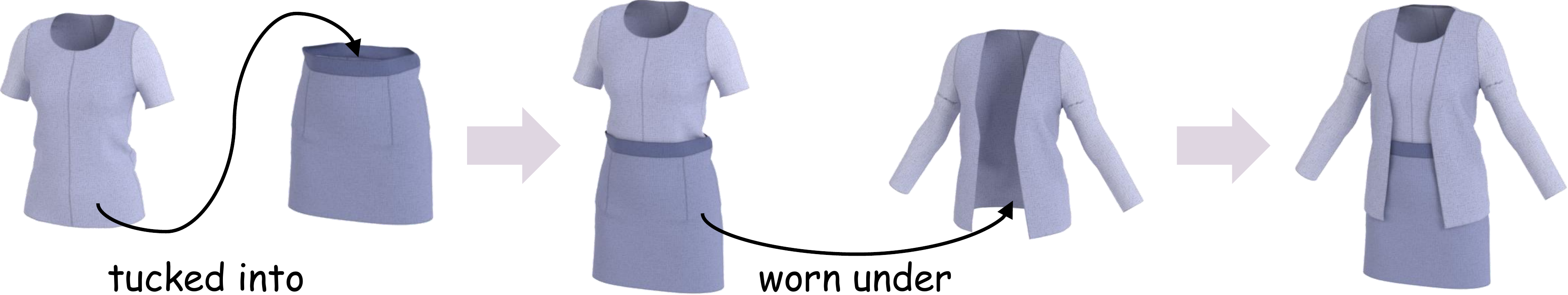}
  \caption{\textbf{Example of a two-layer GarmentCode-XL outfit.} The tee is \emph{tucked into} the skirt within the inner layer $L_1$. The cardigan forms the outer layer $L_2$, with $L_1$ \emph{worn under} $L_2$.}
  \label{fig:appendix_layered}
\end{figure}

We use the outfit in Figure~\ref{fig:appendix_layered} as a concrete example.
The listing below gives its GarmentCode-XL program.
For clarity, we omit inactive branches, secondary parameters, default values, and schema annotations from the listing.

\begin{promptbox}{Abridged GarmentCode-XL outfit program for Figure~\ref{fig:appendix_layered}}
\begin{lstlisting}[language=GarmentJSON]
{
  "outfit": {
    "layers": {
      "L1": {"program": "P1",
             "garments": ["white_tee", "black_skirt"]},
      "L2": {"program": "P2", "garments": ["beige_cardigan"]}
    },
    "relations": [
      {"within_layer": "white_tee tucked_into black_skirt"},
      {"between_layers": "L1 worn_under L2"}
    ]
  },
  "programs": {
    "P1": {
      "meta": {"upper": "Shirt", "wb": "StraightWB",
               "bottom": "PencilSkirt"},
      "shirt": {"length": 1.610, "openfront": false},
      "collar": {"f_collar": "CircleNeckHalf", "fc_depth": 0.402},
      "sleeve": {"length": 0.101, "width": 0.920},
      "waistband": {"width": 0.199},
      "pencil-skirt": {"length": 0.201, "rise": 1.000,
                        "flare": 0.996}
    },
    "P2": {
      "meta": {"upper": "Shirt", "wb": null, "bottom": null},
      "shirt": {"length": 1.610, "openfront": true},
      "collar": {"f_collar": "VNeckHalf", "fc_depth": 1.235},
      "sleeve": {"length": 1.150, "width": 0.600,
                 "connecting_width": 0.887}
    }
  }
}
\end{lstlisting}
\end{promptbox}

Since no data exists for this new language, we build GarmentCodeData-XL, the first multi-layer sewing pattern dataset, for subsequent supervision for Stage 1 and benchmarking for different methods and models.
We sample programs from the expanded design space, compose compatible garments into outfits of multiple layers, and discard invalid drapes.
Each example contains the outfit program, sewing patterns, body measurements, rendered views, and simulated meshes.
The dataset contains $15{,}205$ outfits in total: $4{,}640$ with one layer, $9{,}282$ with two layers, $721$ with three layers, and $562$ with four layers.
It provides both diverse garment designs and valid multi-layer compositions.
We hope it serves as a common benchmark for multi-layer garment recovery and supports future research on layered clothing.

\textbf{Simulation Extensions for Layered Outfits.}
The layered representation also requires an enhanced simulator.
The original GarmentCode solver~\citep{garmentcode} drapes a single garment on a static body and does not model contact across layers.
We keep its XPBD core~\citep{xpbd} on Warp~\citep{warp} and build an outfit-level pipeline on top of it that turns $\mathcal{P}$ into a dressing procedure.
The pipeline first partitions the outfit into stitch-connected garments and schedules them by the layer sequence and the relations in $\mathcal{R}$.
A worn-over garment is draped after every garment beneath it.
A tucked top is draped before its bottom, and its cloth below the waistline is projected onto the body surface so that the bottom can close over it.
Once a garment settles, it is frozen and registered as a thin-shell triangle-mesh collider for the layers above it, with a small collision standoff and the same soft frictional contact model as the body.

A layered drape is accepted only when every layer has settled into static equilibrium, clear of the body and free of self-intersection.
The same dressing order extends from static draping to dynamic simulation.
The body acts as a kinematic boundary condition, each cloth vertex is advected with its nearest body vertex on the same body part, and every settled inner layer is replayed as a moving collider for the layer above it, so the layered structure persists under motion.

\section{Stage 1: Layer Planning and Coarse Estimation}
\label{appendix:stage1}

This section provides the implementation details of Stage 1 (Sec.~\ref{sec:stage1}), including the model of each agent, the prompt and skill design, the agent interaction, the structure of the Garment Style Template, and the encoder $E$ and decoder $\Psi$.

\subsection{Agent Implementation and Interaction}
\label{appendix:s1_workflow}

\textbf{Agent Implementation.}
Stage 1 is a multi-agent workflow built on state-of-the-art foundation models, each equipped with carefully designed skills and tools:

\begin{itemize}[leftmargin=1.5em, itemsep=3pt, topsep=3pt, parsep=0pt]
\item The Layer Planner and the Visual Reader are both GPT-5.6-luna~\citep{gpt56}, a state-of-the-art multimodal reasoning model, run at medium reasoning effort.
The Layer Planner reads the canonical front view $I^{\mathrm{f}}$ with a carefully designed reasoning skill, infers the layer structure $\langle K,\mathcal{R}\rangle$ with a description $r_k$ per layer, and calls $g_\text{peel}$ to expose each layer in its own view.
The Visual Reader reads the two views $I^{\mathrm{f}}_k$ and $I^{\mathrm{b}}_k$ of one layer, answers the questionnaire of that layer, and returns its Garment Style Template $\text{GST}_k$.

\item The image tools $g_\text{canonical}$, $g_\text{back}$, and $g_\text{peel}$ are GPT-Image-2~\citep{gptimage2}, a state-of-the-art image generation and editing model, wrapped in a strict-edit interface that pins identity, pose, and framing and returns one $1024\times1024$ view.
$g_\text{canonical}$ turns the input photograph $I$ into the canonical A-pose front view, $g_\text{back}$ turns any front view into the matching back view, and $g_\text{peel}$ removes the layer described by $r_k$ from $I^{\mathrm{f}}_k$ to give $I^{\mathrm{f}}_{k-1}$.

\item The questionnaire writer $g_\text{ques}$ is GPT-5.6-luna at medium reasoning effort, run once offline (Appendix~\ref{appendix:s1_ques}).
It reads the source code behind every design parameter and writes the questionnaire $\mathcal{Q}$, which is frozen thereafter.

\item The encoder $E$ is Qwen3.6-27B~\citep{qwen36}, an open-weight VLM of hidden width $5{,}120$, into which we distill GarmentCode-XL semantics with LoRA (Appendix~\ref{appendix:s1_lora}).
It reads a question with its selected option and returns the hidden state that conditions the decoder.

\item The decoder $\Psi$ is a $51.5$M-parameter transformer that we design and train (Appendix~\ref{appendix:s1_arch}).
It reads the $131$ encoded answers of one layer and decodes the coarse program $P^{0}_k$.
\end{itemize}

\textbf{Agent Interaction.}
Here we summarize the data flow that ties the two agents and the encoder-decoder together.
The Layer Planner opens the workflow by writing the outfit structure $\langle K,\mathcal{R}\rangle$ and one description $r_k$ per garment.
The relations $\mathcal{R}$ order the peel and $r_k$ names the garment that $g_\text{peel}$ removes at each step, which leaves every layer in a view where it is the outermost garment.
We note that peeling stops at the innermost layer, which is already fully visible, so no view ever exposes the bare body.
The same $r_k$ then tells the Visual Reader which garment to read in that view.

The questionnaire $\mathcal{Q}$ is written once by $g_\text{ques}$, which turns every design parameter $d_i$ into a question $q_i$ with a closed option list, and the garment type and the asymmetry flag from the Layer Planner select the subset of $\mathcal{Q}$ that reaches the Visual Reader (Appendix~\ref{appendix:s1_read}).
The Visual Reader answers one option $\tau_i$ per question and fills the template $\text{GST}_k$, and $E$ encodes the pair $(q_i,\tau_i)$ into the condition vector of the decoder $\Psi$, so a question and its answer share one embedding space.
$\Psi$ turns each $\text{GST}_k$ into a program $P^{0}_k$, and $\mathcal{R}$ from the Layer Planner assembles these programs into the coarse outfit $\mathcal{P}^{0}$ handed to Stage 2.

\subsection{Canonical A-pose Views}
\label{appendix:s1_canonical}

$g_\text{canonical}$ canonicalizes the input photograph $I$ into the A-pose front view $I^{\mathrm{f}}$, and $g_\text{back}$ turns that view into the matching back view $I^{\mathrm{b}}$.
Both edits change only the pose, the background, and keep the person and every garment as they are.
Both views are locked to a $1024\times1024$ canvas with the head at the top edge and the feet at the bottom, so an analytic $180^\circ$ orbit of $\pi^{\mathrm{f}}$ lands on the back view and no alignment network is needed.
We show the prompt of $g_\text{canonical}$ below:

\begin{promptbox}{Prompt of $g_\text{canonical}$}
\scriptsize
\begin{verbatim}
You are given ONE image: an in-the-wild photo of a REAL person wearing an outfit.

Task: output a single clean, photorealistic FULL-BODY photograph of the SAME real
person, re-posed into the canonical standing pose described below, standing still.

Keep exactly (the real subject -- do NOT turn them into a mannequin):
- THE PERSON: their true proportions, build and body shape, with natural skin.
- WHAT EACH GARMENT IS: type, color, print, fabric, neckline shape, sleeve length,
  waistband and hem position, overall length, silhouette, collars, cuffs, slits,
  pleats and seams.
- HOW IT IS WORN: the layer order and how each garment is fastened -- a zip stays
  open to the same height, an open front stays open by the same amount, buttons
  done up stay done up. Never close a garment the photo leaves open.
- WHAT SHOWS THROUGH: every inner garment stays visible wherever the photo shows
  it, in the same color and with the same neckline. Hiding an inner layer deletes
  a garment from the outfit.

Canonical standing pose (a strict catalog pose, NOT a relaxed fashion pose). The
ARMS matter most and are most often rendered wrong, so read them first:
- Arms: both straight arms raised out to the sides and then lowered about a third
  of the way toward hanging, like a T-pose whose arms have dropped by a third.
  Each wrist ends level with the bottom of the ribcage, a hand's width below the
  chest and a hand's width above the natural waist, far out to the side, with a
  wide open triangle of white background between each arm and the torso. Both
  shoulders level, both elbows straight, matching left and right joint heights,
  neither hand touching the body. A narrow pose with the hands at the hips is
  wrong; a horizontal T-pose is also wrong, but if unsure, raise rather than lower.
- Upright torso facing the camera straight-on; legs straight and vertical, feet
  flat on the ground and slightly apart, not a wide fashion stance.
- Head upright and level: no tilt, no lean, no turn, eyes into the camera.
- Perfectly bilaterally symmetric, centered, and filling the frame head-to-toe.
- Studio: the background is PURE WHITE, exactly RGB 255,255,255 -- no tint, no
  gradient, no floor line, no shadow. Soft even light. No text, props or
  reflections.

Settle the cloth under gravity to a STATIC rest state: no wind, no motion, no
billowing fabric. If the photo shows an action pose, settle every garment to how
it would hang while the person stands still in this pose.

Remove every occluder so the garment is FULLY visible:
- Hair: keep the person's own length and style. Only if it falls over the
  neckline, shoulders, chest or straps, tie it back behind the shoulders.
- No bags, held objects, jewelry, scarves over the garment, or arms crossing
  the torso.
- If a garment is cropped or occluded, complete it conservatively as an opaque
  everyday garment consistent with the visible fabric, color and placement; do
  not invent extra layers or expose more of the body than the original outfit.
- Remove all footwear and redraw the feet as bare skin, so leg length is
  unambiguous.

Before you finish, check the arms once more: wrists level with the bottom of the
ribcage and far out from the body, a wide white triangle under each armpit,
elbows straight, matching joint heights. Check the feet too: redraw any remaining
shoe or sock as bare feet.

Output image size: exactly 1024x1024 pixels -- a SQUARE 1:1 image. The person
stands centered and spans the full height of the square, head near the top edge
and feet near the bottom edge.
\end{verbatim}
\end{promptbox}

$g_\text{back}$ then generates the back view from the front view, as if the camera orbits $180^\circ$ around the person while the pose and framing stay fixed.
The prompt asks for the same framing as the front view, so that the two views align and the back camera follows from the analytic orbit of $\pi^{\mathrm{f}}$, and infers the back of each garment from its front.
We show the prompt of $g_\text{back}$ below:

\begin{promptbox}{Prompt of $g_\text{back}$}
\scriptsize
\begin{verbatim}
You are given ONE image: a clean FRONT catalog photo of a real person wearing an
outfit, standing in the canonical pose with both straight arms raised out to the
sides and lowered a third of the way toward hanging.

Task: output a single photorealistic FULL-BODY photograph of the SAME person, in
the SAME pose, seen from DIRECTLY BEHIND -- the person has turned 180 degrees
about their own vertical axis, and nothing else has changed.

FRAMING MUST BE IDENTICAL TO THE INPUT (this is critical):
- Exactly the same scale and position in frame: the top of the head and the soles
  of the feet sit at the SAME image height as in the input; the body is centered
  the same way and fills the frame identically.
- Preserve the pose exactly: shoulder, elbow, wrist and fingertip rows and their
  horizontal distances from the image center match the input. Both arms stay
  straight and symmetric on the same diagonals, wrists level with the bottom of
  the ribcage, the same wide white triangle beside the torso. Legs straight and
  close; standing still and barefoot.
- Studio: PURE WHITE background, exactly RGB 255,255,255, soft even light, no
  floor line or shadow, no text, props or reflections. Cloth settled to a static
  rest state under gravity.

BACK APPEARANCE (infer it from the front):
- The input defines the COMPLETE and EXACT garment set. Keep those garments and
  only those: identical types, layer order, colours, prints, fabrics, sleeve
  length, hem and waistband heights, overall silhouette.
- Render the BACK of each garment plausibly: back neckline, back closures (zips,
  buttons, ties), yokes, pleats, vents, slits, hood, and how the hem and sleeves
  read from behind.
- Where the back is genuinely ambiguous, render the simplest closed, opaque,
  symmetric back consistent with the front.
- HAIR: keep this person's own hair -- same length, colour and style. Only if it
  is long enough to fall over the back of the neck, gather it into a tight bun so
  that the back of the neck, the back neckline of every garment and both
  shoulders are bare of hair.

Output one photorealistic full-body REAR-VIEW image of this person in the same
pose, on a plain studio background, framed identically to the input.

Output image size: exactly 1024x1024 pixels -- a SQUARE 1:1 image. The person
stands centered and spans the full height of the square, head near the top edge
and feet near the bottom edge.
\end{verbatim}
\end{promptbox}

\subsection{Layer Planning}
\label{appendix:s1_structure}

The Layer Planner reads $I^{\mathrm{f}}$ with a carefully designed reasoning skill, infers the layer structure $\langle K,\mathcal{R}\rangle$, and writes a description $r_k$ for each garment.
We show the full skill below:

\begin{promptbox}{Reasoning skill of the Layer Planner}
\begin{lstlisting}[basicstyle=\ttfamily\scriptsize,breaklines=true,breakatwhitespace=true,columns=fullflexible,keepspaces=true,showstringspaces=false,aboveskip=0pt,belowskip=0pt,xleftmargin=0pt,xrightmargin=0pt]
# Garment Layer Planner

You are given one canonical FRONT A-pose image. Identify every visible garment,
its coarse structure, and its inner-to-outer order. Use this image only. Do not
infer fine GarmentCode parameters. A later visual reader answers those questions
from the canonical front and back views.

## Garments

- One record per distinct top, skirt, pants, dress, or jumpsuit. Ignore shoes
  and accessories.
- A dress or jumpsuit is one garment. Call it separates only on visible
  evidence: a top hem, a waistband, a change of fabric at the waist.
- Report a garment when some part of it is visible, and only then. Inner layers
  show at the openings -- a collar above an outer garment's neckline, fabric
  between the edges of an open front, a hem or cuff emerging below or beyond
  the garment above. Read those places before concluding that an outer garment
  is worn against bare skin.
- `name` is short, visible, and unique, such as `brown bomber jacket`.
- `type` is `top | skirt | pants | dress | jumpsuit`. An outer coat, jacket,
  cardigan, or vest is still `top`.

## Structure

For every garment report:

- `closure`: `open` when its front edges visibly separate and expose the
  garment below, otherwise `closed`.
- `neck`: `covered` only for a turtleneck, mock neck, cowl, or raised hood.
  Use `bare` for ordinary necklines.
- `sleeves`: `none`, `short` above the elbow, or `long` at or below the elbow.
  Use `none` for lower-body garments.
- `tucked_into`: exact name of the lower garment whose waistband visibly
  covers this top's hem, so the top is tucked in. Null when the hem hangs
  over the waistband, and for every lower-body garment.
- `asymmetric`: true only when the two sides differ in the garment's overall cut:
  one sleeve present and the other absent, a one-shoulder neckline, a hem that
  sits high on one side and low on the other. Detail on one side is not a
  different cut: a single chest pocket, an off-centre print, a rip on one
  trouser leg.
- `occluded`: a short factual note about parts hidden by another garment.

## Focus

Every garment also carries a `focus`, alongside `name`, `type` and `structure`
rather than inside `structure`. It is one or two sentences addressed to the
reader who will inspect that garment in the view where its layer is
outermost: every outer layer has been removed, while the garments of its own
layer and of the layers below stay on. Name the garment, say which of the
remaining garments are still in that view and must be ignored, and name the
parts of it the reader should judge. Write it from the image, not from a
template: a garment with nothing left beside it needs no warning about
distractors.

## Layer order

Write `order.layers` from inner to outer. Each entry is one layer. A
layer is about covering the same part of the body, never about which edge
lies over which at the waist.

- Layer 1 holds every garment worn directly on the body: nothing under it
  covers the same part of the body. A top and a bottom worn together are both
  layer 1, whether the top's hem hangs over the waistband or is tucked into
  it. A dress or jumpsuit worn alone is layer 1.
- A garment belongs to an outer layer only when a garment beneath it covers
  the same part of the body and shows at its openings: a jacket, cardigan,
  vest or second top over a top or dress (torso over torso); a skirt over
  pants, or a dress over a top or pants (the inner garment continues under the
  outer one's fabric). A garment with nothing under it over that region goes
  to layer 1, however loose, long or coat-like it is.
- Where a top meets a bottom the two always overlap by a few centimetres,
  either hem over waistband or hem tucked under it. That overlap is recorded
  in `tucked_into`, and the two garments stay in the same layer. Putting the
  top in an outer layer because its hem covers the waistband is an error: the
  reader of the pants would then be shown a view without the top, which is a
  view of a bare torso that was never in the photograph.
- So every layer holds at most one upper-body garment and at most one
  lower-body garment, or one dress or jumpsuit, and every layer above 1 sits
  on a garment from a lower layer that covers the same region.

Examples: a sweatshirt hanging over trousers is `[["sweatshirt", "trousers"]]`.
A tee tucked into a skirt under an open cardigan is
`[["tee", "skirt"], ["cardigan"]]`. A turtleneck and leggings under a pinafore
dress is `[["turtleneck", "leggings"], ["pinafore dress"]]`. A shirt and jeans
with a skirt over the jeans and a coat over the shirt is
`[["shirt", "jeans"], ["skirt", "coat"]]`. Every garment must appear exactly
once.

The reader of each garment works on the view where that garment's layer is
outermost, so `focus` must describe exactly what is still on the body there:
the garments in its own layer and every garment below it. Do not request
a separate image-editing step.

## Output

Reply with one `layers.json` object following the format below. This is only a
format example. Do not copy its names, colours, structures, or layer order.
Fill every value from the input image. Return raw JSON without Markdown fences.

```json
{
  "garments": [
    {
      "name": "white tee",
      "type": "top",
      "structure": {
        "closure": "closed",
        "neck": "bare",
        "sleeves": "short",
        "tucked_into": "black skirt",
        "asymmetric": false,
        "occluded": "sides covered by cardigan"
      },
      "focus": "Read the white tee. The black skirt is the only other garment left in this view; ignore it and judge the tee's neckline, sleeve length and how far its hem reaches."
    },
    {
      "name": "black skirt",
      "type": "skirt",
      "structure": {
        "closure": "closed",
        "neck": "bare",
        "sleeves": "none",
        "tucked_into": null,
        "asymmetric": false,
        "occluded": "waist partly covered by cardigan"
      },
      "focus": "Read the black skirt. The white tee above it is the only other garment left; ignore it and judge the skirt's waist height, hem length and flare."
    },
    {
      "name": "beige cardigan",
      "type": "top",
      "structure": {
        "closure": "open",
        "neck": "bare",
        "sleeves": "long",
        "tucked_into": null,
        "asymmetric": false,
        "occluded": ""
      },
      "focus": "Read the beige cardigan. The white tee and black skirt stay visible underneath through its open front; ignore both and judge the cardigan's body length, sleeves, front opening and collar."
    }
  ],
  "order": {
    "direction": "inner_to_outer",
    "layers": [["white tee", "black skirt"], ["beige cardigan"]]
  },
  "summary": "A white tee tucked into a black skirt under an open beige cardigan."
}
```
\end{lstlisting}
\end{promptbox}

\textbf{Schema validation.}
The reply of the Layer Planner is checked against a schema before downstream reads it: the names are unique, the garment type and structural states come from their legal sets, every garment carries a focus note, and the layers hold every garment exactly once.
A reply that fails any check goes back to the Layer Planner with the offending fields named, and the agent answers again.
The Visual Reader is checked in the same way on its own answer sheet.

\textbf{Peel tool.}
$g_\text{peel}$ is an image editing tool that the Layer Planner calls once per layer from the outside in.
One call takes the current front view and the garment to remove, and returns the same photograph with that garment taken off and the garments beneath it completed, so the layer below becomes the outermost.
The peel stops at the innermost layer, so no view exposes the bare body.
The call is a strict edit that preserves identity, pose, and framing, so $I^{\mathrm{f}}_{k-1}$ stays aligned with $I^{\mathrm{f}}_k$.
We show the tool call below, with the fields that the Layer Planner fills at call time marked in \textcolor{promptslot}{this colour}.

\begin{promptbox}{Tool call of $g_\text{peel}$}
\scriptsize
\begin{alltt}
You are given ONE image: a photorealistic full-body photo of a REAL person in the
canonical pose seen from the direct FRONT view on a plain studio background,
wearing this outfit: \slot{TARGET GARMENT}, \slot{REMAINING GARMENTS}.

Task: produce a STRICT EDIT of this exact image showing the SAME fully clothed
person wearing only the inner part of the same outfit: \slot{REMAINING GARMENTS}.
In the edited photograph the \slot{TARGET GARMENT} is simply not being worn, the way
a catalogue shows the same look with the outer piece off. Everything else is
pixel-identical. Edit the given pixels -- do NOT redraw the scene.

Keep EXACTLY -- do NOT alter:
- The person: same identity, face, skin tone, hair, body shape and proportions,
  pose, arm angle, position, camera framing and scale.
- The scene: PURE WHITE background, exactly RGB 255,255,255, soft even light, no
  floor line or shadow, no text, props or reflections.
- Every garment that stays on (\slot{REMAINING GARMENTS}): keep their colour,
  length, neckline, sleeves, hem and silhouette exactly as they appear. If a top
  is tucked into a bottom, keep it tucked.

What changes (\slot{TARGET GARMENT}):
- It is absent from the edited photograph, together with its sleeves, collars,
  lapels and waistbands.
- The garments it used to cover are now fully visible and continue normally
  across the region it occupied: \slot{WHAT THE TARGET GARMENT COVERS}
- The person remains fully covered by those garments. No skin is uncovered
  anywhere the outfit did not already leave bare in the input.
- Invent nothing: no undershirt, no accessory, no shoes, socks or jewelry. Only
  what the list above names stays on.

Output a single photorealistic full-body image: the SAME real person in the SAME
pose, camera viewpoint, framing, and studio scene, wearing \slot{REMAINING GARMENTS}.
Output image size: exactly 1024x1024 pixels -- a SQUARE 1:1 image. The person
stands centered and spans the full height of the square, head near the top edge
and feet near the bottom edge.
\end{alltt}
\end{promptbox}

\subsection{Questionnaire Synthesis}
\label{appendix:s1_ques}

Here, the biggest problem we met at first is semantic misalignment: a VLM prompted to write GarmentCode-XL programs directly cannot tell the meaning and bounds of a parameter from its name alone, and its programs drift far from the photograph.
After many attempts, we found a surprisingly effective solution: let an LLM $g_\text{ques}$ read the source code in advance and write one visual question per parameter for the agent to answer.
For every design parameter, $g_\text{ques}$ receives the source of the panel constructor that reads the parameter, the measurements of a mean body, and the fixed list of choice identifiers and numeric bins.
Since the code reveals the geometric meaning of each parameter, we ask it to reason from the program but describe only what is visible: trace how the value shapes a panel, then state the result on the wearer in plain words, anchored to body landmarks such as the waist, hip, and knee.
The resulting questionnaire $\mathcal{Q}$ is written once and frozen thereafter.
We show the system prompt below.

\begin{promptbox}{System prompt of $g_\text{ques}$ for questionnaire synthesis}
\scriptsize
\begin{verbatim}
You are the GarmentCode-XL questionnaire writer. GarmentCode-XL is a sewing-pattern DSL:
each design parameter is read by a Python program that drafts panels for one body. For
every parameter you receive, write one visual multiple-choice question that lets a
vision-language model recover the parameter from a front and a back photograph of a
person wearing the garment. The reader never sees code.

Reason from the program, write for the eye. Read how the program consumes the value:
which panel edge or length it scales, what formula turns it into centimetres on the
body, which other parameters gate it. Then describe only what that produces on the
wearer, as visible geometry anchored to body landmarks (neck base, shoulder point,
collarbone, bust line, underbust, natural waist, hip line, crotch, knee, ankle, elbow,
wrist). When a choice covers a numeric bin, convert the bin through the program with
the supplied body measurements, then state the visible result: where a hem lands, how
far a seam stands off the body, how much wider one edge is than another, how deep an
opening reaches. A reader with a photograph and no ruler must be able to pick the choice.

Write like a tailor describing the finished garment on a person. Never mention the
program, drafting, panels, control points, coefficients, scales, internal lengths,
variable names, percentages of internal quantities, or angles in degrees. Centimetres
are allowed only for a distance a viewer can relate to a body part. A value with no
visible effect of its own is described by its visible consequence.
Bad: "The quadratic control point sits at 0.23 to 0.46 of the curve scale."
Good: "The neckline bows below the straight shoulder-to-centre line by about a fifth of
that line's length, a softly rounded V."

Rules:
- The answer space is fixed. Use every choice id exactly as given, in the given order,
  and nothing else. Numeric bins are already calibrated; describe them, never move them.
- applies_if states the branch of the program under which this parameter is read, in
  plain language naming the parent selection, style or flag.
- For booleans say what true and what false make the program build.
- For select parameters every option names a program construct; describe the visible
  geometry it drafts. Copy the label and description of every token in the glossary.
- Labels are short visual names. Never put a raw identifier into a label or description.
- Never answer the questions. Return JSON only: {"questions":[{"id", "text",
  "visible_feature", "applies_if", "choices":[{"id","label","description"}]}]}.
\end{verbatim}
\end{promptbox}

\textbf{Consistent visual semantics.}
GarmentCode-XL exposes $30$ categorical program tokens such as \texttt{SkirtCircle} and \texttt{Skirt2}, and $17$ of them recur under multiple parameters, so questions written independently may describe the same token differently.
We therefore first ask $g_\text{ques}$ to assign every token one visual label and description, and reuse this glossary wherever the token appears.

\textbf{Question construction and validation.}
With this glossary in place, $g_\text{ques}$ writes one question for each of the $131$ design parameters.
It uses the complete parameter hierarchy to determine when each question applies.
Each record contains a question, its visible feature, the condition under which it applies, and a closed set of choices.
A discrete parameter covers every legal value exactly once, while a numeric parameter uses 3 to 6 ordered visual states instead of raw values.
We validate identifiers, ordering, coverage, and visual labels against the schema, and regenerate any failing record.
Together, these records form a fixed questionnaire $\mathcal{Q}$ whose choices stay consistent across parameters and refer only to observable garment geometry.
Table~\ref{tab:answer_space} summarizes this answer space, synthesized once and frozen, so no language model writes questions at inference time.

\begin{table}[!ht]
\caption{\textbf{The frozen answer space.} Here, every design parameter of GarmentCode-XL becomes exactly one multiple choice question.}
\label{tab:answer_space}
\begin{center}
\footnotesize
\setlength{\tabcolsep}{6pt}
\renewcommand{\arraystretch}{1.1}
\begin{tabular}{@{}l r r l@{}}
\toprule
\textbf{Parameter type} & \textbf{Count} & \textbf{Options} & \textbf{Answer format} \\
\midrule
Length (float) & $83$ & $308$ & 3 to 6 ordered visual states \\
Count (int) & $15$ & $55$ & ordered states, or one option per admissible count \\
Boolean & $14$ & $28$ & the two geometric states of the flag \\
Categorical & $11$ & $56$ & one option per legal value \\
Categorical with absence & $8$ & $34$ & one option per legal value, plus absence \\
\midrule
Total & $131$ & $481$ & \\
\bottomrule
\end{tabular}
\end{center}
\end{table}

\textbf{Mapping visual choices to parameter intervals.}
Since the Visual Reader answers multiple choice questions more reliably than numerical ones, every numeric question offers discrete visual choices, and each choice $\tau_i$ maps to an interval $\mathcal{V}(\tau_i)\subseteq[0,1]$ of the normalized parameter range.
The intervals of one question partition $[0,1]$, so every parameter value belongs to exactly one choice, and their boundaries sit at body landmarks rather than uniform splits of the legal range.
The choice read from the image and the interval the decoder is trained on therefore refer to the same garment geometry.
The shirt-length question below shows how six choices cover the full range.

\begin{promptbox}{Example: shirt-length choices and parameter intervals}
\scriptsize
\begin{verbatim}
id:              design.shirt.length      (float, range [0.5, 3.5], default 1.2)
text:            How long is the visible shirt, judged by where its hem ends on the
                 body (above the waist, at the waist, on the hip, on the thigh, at the
                 knee, or below the knee)?
visible feature: shirt length
applies if:      when the shirt is enabled

  ::cropped   [0.00, 0.14)  "Cropped"
      The hem stops above the natural waist and leaves the midriff bare, anywhere from
      a crop just above the waistband to a very short crop up at the ribcage.
  ::waist     [0.14, 0.28)  "Waist length"
      The hem ends at the natural waist or just onto the high hip, about where a
      trouser waistband sits, like a typical waist-length top.
  ::hip       [0.28, 0.40)  "Hip length"
      The hem covers the hip and seat completely, ending at or just past the hip
      crease, like a standard T-shirt.
  ::thigh     [0.40, 0.60)  "Mid-thigh"
      The hem falls onto the thigh, below the seat but clearly above the knee, a tunic
      or short-dress length.
  ::knee      [0.60, 0.78)  "Knee length"
      The hem ends at the knee, just above or just below the kneecap.
  ::calf      [0.78, 1.00]  "Below the knee"
      The hem falls past the knee to mid-calf or as far as the ankle, a long dress
      length.
\end{verbatim}
\end{promptbox}

\subsection{Template Generation with the Visual Reader}
\label{appendix:s1_read}

The Visual Reader answers the questionnaire for each layer from its views $I^{\mathrm{f}}_k$ and $I^{\mathrm{b}}_k$ and description $r_k$, and its answers form the Garment Style Template $\text{GST}_k$.
Rather than asking all $131$ questions for every garment, we use its type and asymmetry state to select only the compatible branches of $\mathcal{Q}$.
Garments in the same layer share one multimodal call but keep separate answer sheets.
The reader resolves construction choices first, and uses their activation conditions to skip questions about geometry the garment does not have.
We show the prompt below:

\begin{promptbox}{Prompt of the Visual Reader}
\scriptsize
\begin{alltt}
You are the Visual Reader. Image 1 is the FRONT view and Image 2 is the BACK view.
The current layer contains these target garments: \slot{TARGET GARMENT NAMES}.
Read every target in this request, but keep a separate answer sheet for each. Use each
target's focus text and questionnaire. Other garments may provide wearing context, but
never use one garment as visual evidence for another.

Question ids are template paths named for the program that constructs the garment, not
for its everyday category. For example, design.shirt.length refers to the body length
of the current upper garment, including a cardigan, jacket, or vest.

Resolve construction questions first, then follow each applies_if condition. For every
active question, choose exactly one listed choice id. Match its geometric description
against the two images and body landmarks. When a region is occluded, infer its full
extent only from the target's construction and visible cues. Use "not applicable" when
the condition is false and "not visible" when an active feature cannot be judged.

Return JSON only:
\{"garments":[\{"index":...,"name":...,"answers":[
  \{"id":...,"answer":...,"confidence":"high|medium|low","evidence":...\}
]\}]\}.
Keep garment identity, garment order, question ids, and question order unchanged.
Emit every requested row, including inactive rows as "not applicable".

\slot{ONE FOCUS NOTE AND PRUNED QUESTIONNAIRE PER TARGET GARMENT}
\end{alltt}
\end{promptbox}

\textbf{Garment Style Template formation.}
The selected options form $\text{GST}_k$.
Questions that are inactive or not visible are masked, so the decoder $\Psi$ receives only supported answers.
Table~\ref{tab:questionnaire_size} lists the number of questions retained for each garment type and symmetry state.

\begin{table}[!ht]
\caption{\textbf{Layer-specific questionnaire size.} Number of questions retained from $\mathcal{Q}$ for each garment type and symmetry state.}
\label{tab:questionnaire_size}
\begin{center}
\footnotesize
\setlength{\tabcolsep}{7pt}
\renewcommand{\arraystretch}{1.1}
\begin{tabular}{@{}l r r r r@{}}
\toprule
& \multicolumn{2}{c}{\textbf{Symmetric}} & \multicolumn{2}{c}{\textbf{Asymmetric}} \\
\cmidrule(lr){2-3}\cmidrule(lr){4-5}
\textbf{Garment type} & Questions & Numeric & Questions & Numeric \\
\midrule
Top & $48$ & $34$ & $79$ & $56$ \\
Skirt & $42$ & $33$ & $42$ & $33$ \\
Pants & $15$ & $12$ & $15$ & $12$ \\
Dress & $90$ & $67$ & $121$ & $89$ \\
Jumpsuit & $63$ & $46$ & $94$ & $68$ \\
\bottomrule
\end{tabular}
\end{center}
\end{table}

\subsection{Encoder-Decoder from Template to Program}
\label{appendix:s1_lora}

The encoder-decoder maps each Garment Style Template back to an executable program.
The encoder $E$, with distilled GarmentCode-XL semantics, embeds every answered question, and the decoder $\Psi$ turns these embeddings into the design parameters of $P^{0}_k$.

\textbf{Encoder $E$.}
The encoder $E$ reads each question with its selected option and returns a hidden state that conditions the decoder.
To interpret an answer in terms of the program parameters, $E$ needs the semantics of every GarmentCode-XL program, which requires fine-tuning.
We therefore build $E$ on the open-weight Qwen3.6-27B~\citep{qwen36} and distill the program semantics into it from $g_\text{ques}$, which has already read the source code.
Here $g_\text{ques}$ takes a second pass over the source with the opposite goal: the questionnaire turns each parameter into visible geometry for the Visual Reader, while the descriptions explain each program unit for the encoder.
Pairing every description with its source program gives the training set $\mathcal{X}$ of $417$ units at two levels of the program hierarchy: $122$ component classes and top-level programs that specify complete panel construction and assembly, and $295$ methods that implement local drafting operations such as constructing curves, transforming edges, placing darts, and joining sewing interfaces.
Together they cover all $131$ design parameters and connect each parameter to the geometry it controls.
We show the prompt below.

\begin{promptbox}{Prompt of $g_\text{ques}$ for program description synthesis}
\scriptsize
\begin{verbatim}
You are preparing Program Learning data for GarmentCode-XL. Read the given
implementation and explain it as detailed natural-language pattern-drafting
instructions at the implementation's own level of abstraction. For garment components,
explain panels, geometry, interfaces, stitches, assembly, and the visual effect and
direction of every design parameter used. For geometric primitives, explain the
exact drafting operation and how later components use its result. Resolve abstract
component names from their actual geometry and assembly logic. Use only facts
supported by the code. Do not reproduce code, invent missing behavior, or merely
paraphrase identifiers. Write at least 60 words so that the description-to-program pair
contains enough semantic context.

User: Comment `{module}.{symbol}`. Return JSON only as {"instruction": "..."}.

```python
{source code of the program unit}
```
\end{verbatim}
\end{promptbox}

\textbf{Distillation targets.}
We fine-tune $E$ on two kinds of pairs.
The first maps each description to its source program, so $E$ learns how language translates into drafting code.
The second maps the program excerpt that reads each design parameter to the question written for it, so $E$ learns the geometry behind each question.
$E$ then encodes a question with its selected answer into a vector reflecting the program parameters.

\textbf{LoRA setting.}
We apply LoRA with rank $16$ and $\alpha=32$ to every linear projection in the attention and feed-forward blocks, and keep the vision encoder frozen.
We use sequence length $8{,}192$, per-device batch size $1$, and gradient accumulation $8$.
The first two epochs use a cosine schedule with a peak learning rate of $5\times10^{-4}$ and $3\%$ warmup, then a tenfold lower peak rate with evaluation every quarter epoch.
The full training takes within 1 hour on two NVIDIA RTX PRO 6000 GPUs.

\begin{figure}[!ht]
  \centering
  \includegraphics[width=0.88\textwidth]{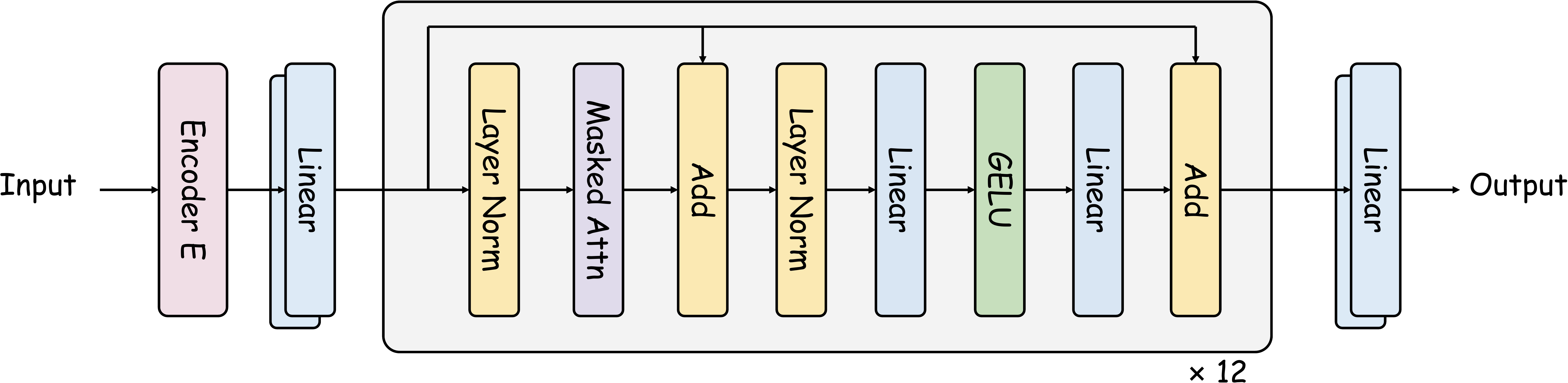}
  \caption{\textbf{Architecture of the decoder $\Psi$.} The frozen encoder $E$ embeds each question with its selected answer. A projection MLP maps the embeddings to the decoder width, and a causal Transformer with $12$ blocks decodes the numeric program parameters.}
  \label{fig:projector_arch}
\end{figure}

\textbf{Decoder $\Psi$.}
\label{appendix:s1_arch}
Each numeric question offers visual ranges rather than raw values, and the template records the selected range.
The decoder $\Psi$ maps it back to a precise value that the program can execute: it keeps the $33$ discrete choices fixed by the template and autoregressively decodes the $98$ numeric parameters.
A projection MLP maps the $5{,}120$-dimensional condition vectors of $E$ to the decoder width of $512$, and a causal Transformer decodes the parameter tokens (Fig.~\ref{fig:projector_arch}).

\textbf{Parameter tokenization.}
We represent the $|\mathcal{D}|=131$ (total number of design parameters in GarmentCode-XL) design parameters as an ordered token sequence $\mathcal{T}=(t_1,\dots,t_{|\mathcal{D}|})$, where the quantizer $\mathbf{Q}$ respects the type of each parameter:
\begin{equation}
    t_i=\mathbf{Q}(d_i) =
    \begin{cases}
        0 \text{ or } 1, & \text{if } d_i \text{ is a boolean, such as a strapless top}, \\
        d_i, & \text{if } d_i \text{ is a count, such as the number of panels}, \\
        \operatorname{round}\!\left(\lambda \cdot \mathrm{Norm}(d_i)\right), & \text{if } d_i \text{ is a length, normalized to } [0,1] \text{ over its range}, \\
        \mathrm{Enum}(d_i), & \text{if } d_i \text{ is a categorical choice, selected among its options}.
    \end{cases}
    \label{eq:quantize}
\end{equation}
We use $\lambda=100$, which keeps centimeter-level precision for every length.

\textbf{Training details.}
We train $\Psi$ on $232{,}500$ garment programs from the official training split of GarmentCodeData-v2 ($205{,}487$) and the training split of GarmentCodeData-XL ($27{,}013$), and validate on the official validation split of GarmentCodeData-v2 ($13{,}195$). The GarmentCodeData-XL is split at the outfit level, so no outfit appears on both sides.
$\Psi$ has $51.5$M trainable parameters and is trained for $12$ epochs with batch size $256$ using AdamW, a peak learning rate of $5\times10^{-4}$, weight decay $0.05$, and cosine decay with $5\%$ warmup.
We use dropout $0.08$, gradient clipping at $1.0$, and cross-entropy over the $98$ numeric tokens with teacher forcing.
The training takes within 2 hours on two NVIDIA RTX PRO 6000 GPUs.

\section{Stage 2: Fine-Grained Refinement}
\label{appendix:stage2}

This section provides the implementation details of Stage 2 (Sec.~\ref{sec:stage2}), including the implementation of the agentic trio, the complete tool interface, the design of observations and measurements, the \texttt{Fit} tool for size fitting, and the full prompt of each agent.

\subsection{Agent Implementation and Interaction}
\label{appendix:s2_workflow}

\textbf{Agent Implementation.}
All three agents are GPT-5.6-luna~\citep{gpt56} at medium reasoning effort.
Each agent has its own system prompt, its own tool set, and a persistent memory that spans the whole session of one layer.
As in Sec.~\ref{sec:stage2}, the agents interact with the garment exclusively through structured tools, which are backed by a shared session that drafts and simulates programs, keeps every simulated step as a checkpoint together with its critique, tracks the simulation budget, and composes the observations of Appendix~\ref{appendix:s2_tools}.
Each layer opens with a budget of $T$ simulations (we set $T{=}6$ by default).

\textbf{Agent Interaction.}
The session opens by draping the coarse program $P^{0}$ of Stage 1 as step $0$.
After every simulation, the Orchestrator invokes the Critic as a tool, and stores its scores $s^{t}$ and discrepancies $F^{t}$ with the program and observation of that step as a checkpoint in the history $H^{t}$.
The Orchestrator then reads the history of all steps and takes one of its three actions: it continues with an instruction to the Tailor, rolls the program back to an earlier checkpoint, or finishes the layer.
Within its round, the Tailor edits freely with the observation and modification tools and ends the round with one \texttt{Simulate} call, or returns with \texttt{Done} when no remaining discrepancy is worth a simulation.

\textbf{Session Memory.}
Each agent keeps its own message history across rounds.
The views of the two most recent rounds stay inline, older views are compacted into text notes, and every round restates the history of steps, so the agents keep the whole trajectory in context at a bounded size.
\texttt{Rollback} restores the program only and leaves the history and each agent's memory intact for reference, which realizes the evolving memory of Sec.~\ref{sec:stage2}.

\subsection{Tool Interface}
\label{appendix:s2_tools}

Table~\ref{tab:stage2_tools} lists the complete tool interface of Stage 2.
Tool access is role-dependent: the Tailor has access to observation and modification tools, the Critic is restricted to observation tools, and the Orchestrator invokes the Tailor and Critic as tools while managing global control operations such as rollback.
We note that only \texttt{Simulate} spends the budget, and every other tool is free, so the Tailor can explore the pattern space extensively between two simulations.

\begin{table}[!ht]
\caption{\textbf{The complete tool interface of Stage 2.} Access: T = Tailor, C = Critic, O = Orchestrator.}
\label{tab:stage2_tools}
\begin{center}
\footnotesize
\setlength{\tabcolsep}{5pt}
\renewcommand{\arraystretch}{1.15}
\begin{tabularx}{\textwidth}{@{}l c l Y@{}}
\toprule
\textbf{Tool} & \textbf{Access} & \textbf{Arguments} & \textbf{Description} \\
\midrule
\multicolumn{4}{@{}l}{\textit{Observation}} \\
\texttt{Render} & T & \texttt{pictures} & Views of any step: the photograph, the target and drape pair of either view, the before/after strip, per-garment mask diffs, and the flat sewing pattern. \\
\texttt{Measure} & T & \texttt{step} & Measurement tables of one simulated step: every garment edge on the drape against the target, and the same sizes read off the drafted pattern, in percent of body height. \\
\midrule
\multicolumn{4}{@{}l}{\textit{Modification}} \\
\texttt{Draft} & T & \texttt{edits} & Drafts the current program with the edits and returns the garment sizes of the flat pattern, with the change against the current step. \\
\texttt{Fit} & T & \texttt{part}, \texttt{channel}, \texttt{lever}, \texttt{target} & Value of one numeric lever that lands one measured size on its target, found through drafting probes (Appendix~\ref{appendix:s2_fit}). Costs no simulation. \\
\texttt{Simulate} & T & \texttt{edits}, \texttt{note} & Drapes the current program with the edits applied, one call per round; ends the round. \\
\midrule
\multicolumn{4}{@{}l}{\textit{Control}} \\
\texttt{Scoring} & C & \begin{tabular}[t]{@{}l@{}}\texttt{scores}, \texttt{discrepancies},\\ \texttt{comparison}, \texttt{done}\end{tabular} & The Critic's structured judgement of one step (Appendix~\ref{appendix:s2_critic}). \\
\texttt{Done} & T & \texttt{note} & Returns the round to the Orchestrator when no remaining discrepancy is worth a simulation. \\
\texttt{Critic} & O & \texttt{step} & Invokes the Critic on a simulated step; called after every simulation. \\
\texttt{Tailor} & O & \texttt{instruction} & Continues from the current step; the Tailor answers with one simulation. \\
\texttt{Rollback} & O & \texttt{step}, \texttt{instruction} & Rolls the program back to an earlier checkpoint and keeps every memory. \\
\texttt{Finish} & O & \texttt{step}, \texttt{reason} & Ends the layer on one step; a step with cloth through the body is refused if a valid step exists. \\
\bottomrule
\end{tabularx}
\end{center}
\end{table}

\textbf{Observation Design.}
Every simulated step is observed in three complementary forms.
The paired views place our render beside the canonical target on one canvas, cropped identically and overlaid with the same body landmark lines, so a hem height is read on the same ruler in both images.
A before/after strip adds the parent step, and per-garment mask diffs mark in green the target cloth our drape lacks and in red our cloth beyond the target.
The tables measure eight channels per garment on both views in percent of body height: four positions (top edge, bottom edge, sleeve end, neck depth) and four extents (width, hem width, sleeve width, neck width), each reported as ours, target and gap, beside the same sizes read off the drafted pattern.
Finally, the solver's report is attributed per garment (e.g., cloth through the body, self-intersecting edges) together with the edits of the step that touched that garment, so each physical issue is traced to the parameters that caused it.

\subsection{Program Editing with the Tailor}
\label{appendix:s2_tailor}

In each round, the Tailor receives comprehensive context for its edit.
This includes the Orchestrator's instruction $c^{t}$ and the critique $F^{t}$ of the current step, the solver's report, the measurement tables, the history $H^{t}$ with the measured effect of every past edit including those rolled back, and the list of active parameters.
This list gives every parameter its legal range and the visual question that $g_\text{ques}$ wrote for it in Stage 1, with the visual meaning of every numeric interval and categorical option, so the same semantics that ground the Visual Reader also define the action space of the Tailor.
As in Sec.~\ref{sec:stage2}, the Tailor makes two kinds of edits: a construction edit switches a categorical choice such as the neckline or the garment family, and a size edit tunes the numeric parameters to match the reference.
We show the prompt of the Tailor below.

\begin{promptbox}{Prompt of the Tailor $\Phi_{\mathrm{tailor}}$}
\scriptsize
\begin{alltt}
You are the Tailor in a fitting session. A GarmentCode sewing program drafts the
panels of the garments in this layer, a cloth simulator drapes them on the
recovered body, a Critic compares the drape with the photograph and says what
differs, and an Orchestrator tells you what to work on and keeps every simulated
step as a checkpoint. You change the program.

## This layer
\slot{LAYER CONTEXT}

## How the session runs
Each round you receive the Orchestrator's instruction, the Critic's critique of the
current step, the measurements of the current step, and pictures. You end the round
with one `Simulate` call, or with `Done` when no remaining discrepancy is worth a
simulation. Simulations are the scarce resource and drafting is free: use `Draft`
and `Fit` as often as you like before you simulate.

There are two kinds of edit. Construction is categorical: the garment family, the
neckline, the sleeves, the collar, the closure. It decides what the garment is, so
settle construction first. Size is numeric: where a hem falls, how wide a sleeve
is, how deep the neck opens. Reach sizes with `Fit`, which reads the drafted
pattern and returns the parameter value that lands the size on its target. The
answer also names the interval its value lands in; a fit that crosses into another
interval changes the shape, so take it only when the photograph shows that shape.

The solver's report names the garment it objects to. Cloth through the body cannot
be delivered, so clearing it comes before any other improvement, through that
garment's own parameters.

Measurements are evidence, not verdicts: they are read off masks of the segmented
photograph and the rendered drape on one body ruler, and a reading can be wrong
where an edge is hidden or the back view is generated. When a number and the
pictures disagree, say which you trust and why.

Judge your last step before acting on the next: the before/after picture and the
Critic's comparison say whether it helped. Do not repeat an edit that did not move
what it was meant to move.

## GarmentCode parameters
Each entry gives the path, its legal values, and the visual question that defines
its meaning. Only these parameters are active in this layer. Switching the garment
family swaps which parameters exist from the next round.

\slot{ACTIVE PARAMETERS WITH THEIR VISUAL QUESTIONS}
\end{alltt}
\end{promptbox}

\subsection{Size Fitting with the \texttt{Fit} Tool}
\label{appendix:s2_fit}

A key design of our refinement is that the agents reason with measurements: the Critic reports each discrepancy as a measured gap in percent of body height, and the Tailor turns that gap into a parameter value.
The second step is not obvious, since a size edit asks for a hem a few percent lower or a sleeve a third wider, while GarmentCode-XL parameters are fractions of construction quantities.
We therefore design the \texttt{Fit} tool, which infers the parameter value from the history and the measurements of the layer.

Given a target size on the drape, the tool first estimates the relation between drafted and draped sizes from the simulated steps in the history.
It then maps the target into draft space, and searches for the parameter value with a few drafting probes until the drafted size lands on the goal. 
In this way, numeric measurements give each edit a precise target, while the visual intervals and the rendered views keep the shape faithful to the photograph, and the agents judge with both.

\subsection{Drape Evaluation with the Critic}
\label{appendix:s2_critic}

The Critic returns its judgement through the \texttt{Scoring} tool in a fixed schema.
The scores $s^{t}$ give each garment of the layer five integers from $0$ to $10$, one per dimension (construction, proportion, fit, layering, and physical feasibility).
The discrepancies $F^{t}$ are listed largest first, at most six, and each names the garment, the difference against the target, the place on the body, the measured channel when the table has one, and a severity.
The measured value of every named channel is then attached, so a critique reaches the Tailor with its metric evidence in place.
The rationale $r^{t}_{\text{critic}}$ closes with a comparison against the previous step (better, worse, or similar) and a done flag that ends the layer when no remaining difference would visibly improve with another simulation.
The construction vocabulary given to the Critic reuses the categorical questions of Stage 1, and the current program is summarized in the same words, so the Critic compares the photograph and our construction in one vocabulary.
We show the prompt of the Critic below.

\begin{promptbox}{Prompt of the Critic $\Phi_{\mathrm{critic}}$}
\scriptsize
\begin{alltt}
You are the Critic in a fitting session. A Tailor edits a GarmentCode sewing
program, a cloth simulator drapes it on the recovered body, and you compare the
drape with the photograph. You never edit. Your judgement decides what the Tailor
works on next and whether a step is kept.

## This layer
\slot{LAYER CONTEXT}

## What you see
The photograph; our render of the current step beside the canonical target view,
with the same landmark lines and body ruler on both; a before/after strip against
the previous step; per-garment mask diffs (green is target cloth ours lacks, red is
our cloth beyond the target); and a table of measurements in percent of body
height. Compare geometry: what each garment is, where its edges fall, how much room
it has, how it hangs. Ignore colour, print, texture, lighting and the seam lines the
simulator draws between panels.

The numbers are evidence with known failure modes: the masks come from a segmenter,
the back view is generated, and an edge hidden under another garment is not an edge
of that garment. Read them beside the pictures and say when they disagree.

## What you report
Score each garment of this layer on five dimensions from 0 to 10, each on its own:
construction (the same kind of garment, neckline, sleeves, collar and closure),
proportion (where its edges fall on the body), fit (how much room it has and the
shape it makes), layering (how it sits with the garments underneath), and physical
feasibility (whether the cloth hangs as the garment would; cloth through the body
is a failure). Then list the discrepancies worth a simulation, largest first, each
naming the garment, what the target has against what ours has, and where you see
it, with the measured channel when the table has one. Say whether this step is
better or worse than the previous one, and whether the fit is done: done means no
remaining difference would visibly improve with another simulation.

A construction difference counts only when the vocabulary below can build the
target's construction. Finishing details (ribbing, topstitching, buttons, pockets,
prints) are not construction.

## Construction vocabulary in this layer
\slot{CONSTRUCTION QUESTIONS AND OPTIONS}
\end{alltt}
\end{promptbox}

\subsection{Session Control with the Orchestrator}
\label{appendix:s2_orch}

The Orchestrator controls the refinement flow of a layer.
After each critique, it reads the history $H^{t}$ and the critique $(s^{t},F^{t})$, and takes one of three actions: \texttt{Continue} with an instruction $c^{t}$ that tells the Tailor which discrepancies to address, \texttt{Rollback} to an earlier checkpoint when the current step is judged worse, or \texttt{Finish} on the best valid checkpoint.
We show the prompt of the Orchestrator.

\begin{promptbox}{Prompt of the Orchestrator $\Phi_{\mathrm{orch}}$}
\scriptsize
\begin{alltt}
You are the Orchestrator of a fitting session. A Tailor edits a sewing program, a
simulator drapes it, and a Critic scores the drape against the photograph. After
each critique you decide whether the session continues from the current step,
returns to an earlier checkpoint, or ends, and you tell the Tailor what to work on.

## This layer
\slot{LAYER CONTEXT}

Every simulated step is a checkpoint with its critique. A step with cloth through
the body cannot be delivered: clearing it comes first in your instruction, and such
a checkpoint is not a place to finish. Returning to a checkpoint costs nothing and
keeps the Tailor's memory of what was tried, so a step the Critic judged worse is
rolled back rather than built on.

Simulations are the budget: a round that would not change what the Critic sees is
not worth one, but while the best checkpoint still carries a major discrepancy, the
session continues from that checkpoint with a more careful instruction rather than
finishing early. End the session when the Critic finds the fit done, when the
remaining differences are beyond the program's vocabulary, or when the budget is
out, and finish on the step whose garments together score best among those without
cloth through the body.

Your instruction to the Tailor is short: which discrepancies to address this round,
in what order, and what to leave alone.
\end{alltt}
\end{promptbox}

\subsection{An Example Refinement Round}
\label{appendix:s2_round}

\begin{figure}[!ht]
  \centering
  \includegraphics[width=0.494\textwidth]{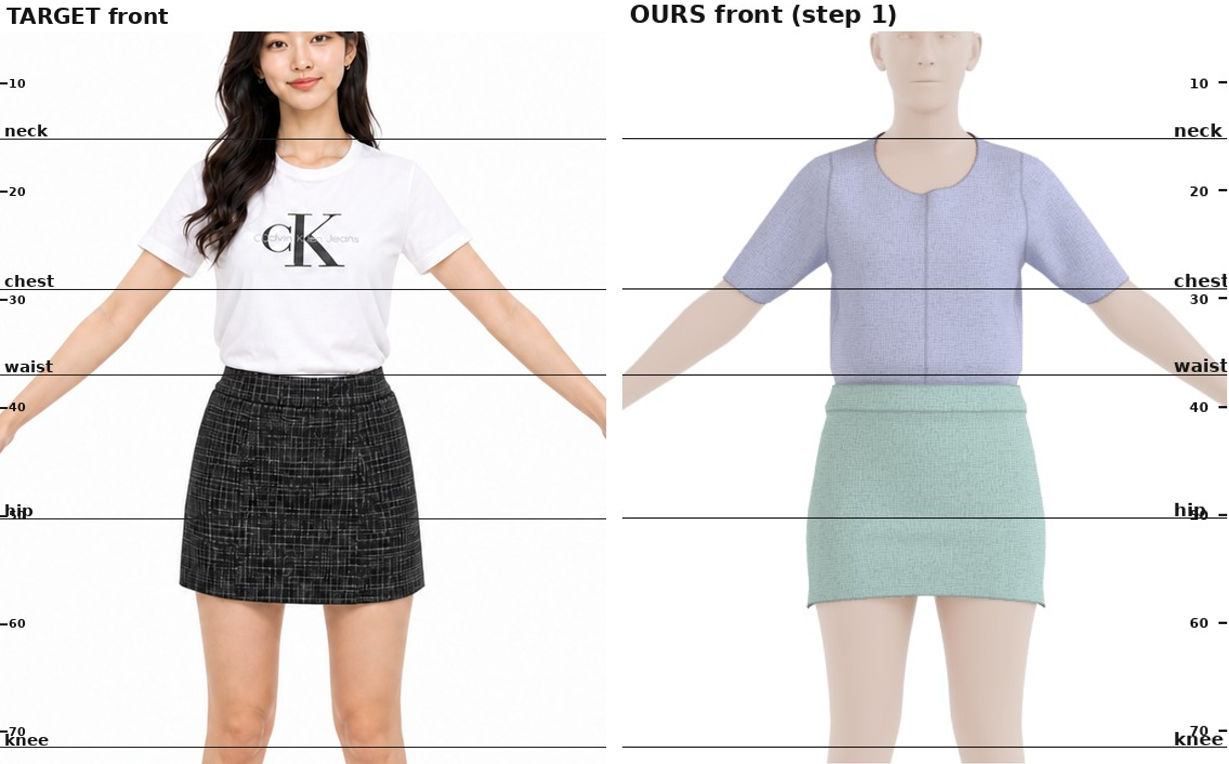}\hfill
  \includegraphics[width=0.494\textwidth]{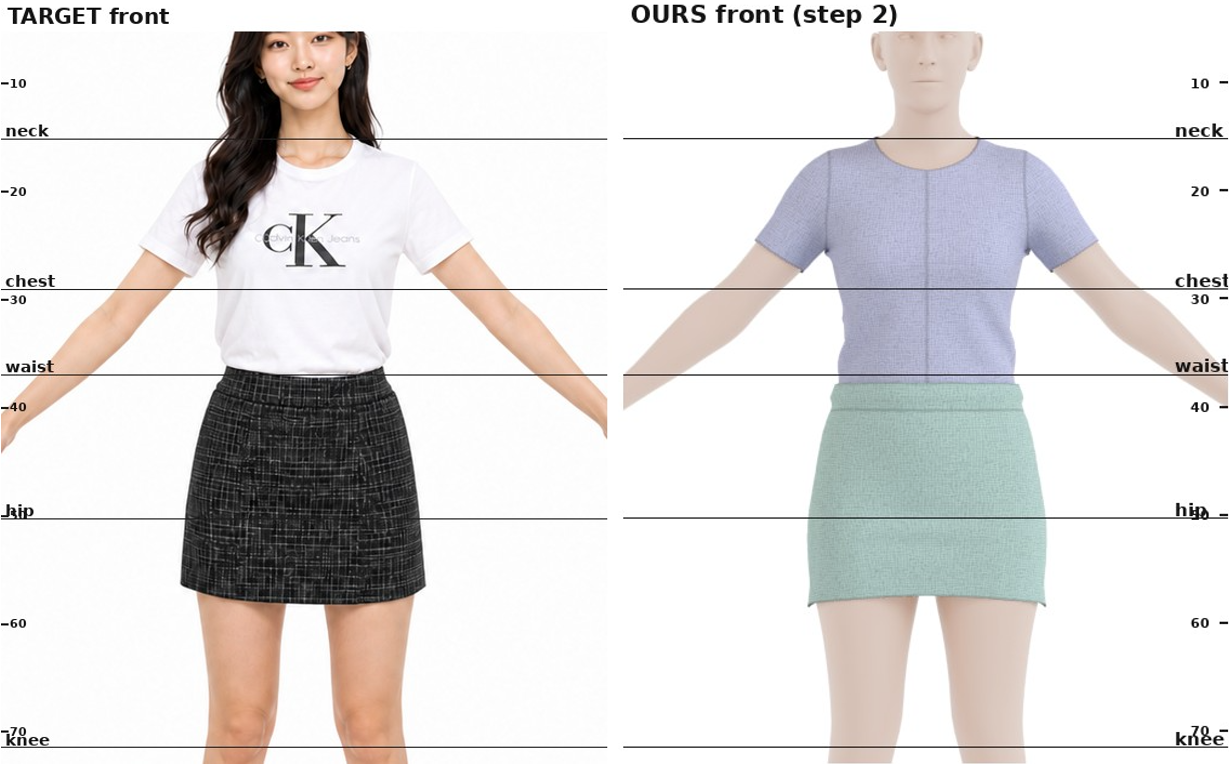}
  \caption{\textbf{The observation $O^{t}$ given to the agents in the recorded round.} \emph{Left}: the paired front view of step 1 (Iter 1 of the inner layer in Fig.~\ref{fig:stage2}), which the Critic and the Tailor read. The skirt has been corrected, and the tee is still wide, deep at the neck, and long in the sleeve. \emph{Right}: the same observation at step 2, after the five fitted edits below. Both panels carry the same landmark lines and body ruler in percent of body height.}
  \label{fig:appendix_round}
\end{figure}

We show the complete record of one refinement round, from Iter 1 to Iter 2 of the inner layer in Fig.~\ref{fig:stage2}.
Fig.~\ref{fig:appendix_round} shows the observation $O^{t}$ given to the agents before and after the round: the rendered drape beside the reference view, with the same landmark lines and body ruler.
At step 1, the tee is still too wide, too deep at the neck, and too long in the sleeve.
The Critic returns the critique below.

\begin{promptbox}{Critique of step 1, returned by the Critic through \texttt{Scoring}}
\scriptsize
\begin{verbatim}
scores  upper: construction 9, proportion 6, fit 6, layering 9, feasibility 8
        lower: construction 8, proportion 9, fit 8, layering 9, feasibility 9
against the previous step: better. "The skirt edit is a clear improvement: its
  front and back hems now align with the target within measurement noise and its
  front hem width is essentially matched. The upper garment is unchanged, so its
  broad torso, deep neckline, and long sleeves remain."
discrepancies, largest first:
  1. [upper, major, width] The target tee is closer and slimmer through the
     torso, while ours remains visibly wider. Front and back bodice; ours is
     about 2.4-3.2 cm wider.
  2. [upper, minor, hem_width] The target tee has a narrower lower body, while
     ours flares wider at the tucked-in lower edge. Front gap +2.2 %bh, about
     3.7 cm, though this edge is partly affected by the tuck.
  3. [upper, minor, neck_depth] The target front neckline is shallower, while
     ours drops lower. Ours is about 2.9 cm deeper; the back comparison is less
     reliable and shows the opposite direction.
  4. [upper, minor, sleeve_end] The target sleeves finish higher, while ours
     extend farther down the upper arms. Both sleeve openings; ours is about
     2.7-3.4 cm lower.
  5. [upper, minor, bottom_edge] The target visible bodice ends slightly
     higher, while ours extends lower into the tucked region.
  6. [lower, minor, hem_width] The target back hem is slightly fuller, while
     ours remains a little narrower: 0.8 %bh, about 1.4 cm. Front hem width is
     within noise.
summary: "Keep this step: the skirt shortening and flare increase successfully
  corrected the dominant lower-garment error, with excellent front alignment
  and only a small back-hem width residual. The remaining visible improvement
  is in the unchanged tee: reduce its torso width, then address the deeper
  front neck and overly long sleeves; the fit is not done."
\end{verbatim}
\end{promptbox}

The Orchestrator reads the history and this critique, keeps the checkpoint, and continues with the instruction below.

\begin{promptbox}{Instruction of the Orchestrator, passed through \texttt{Tailor}}
\scriptsize
\begin{verbatim}
Keep the skirt unchanged, including its successful hem length and flare. First
narrow the tee through the torso and especially the visible tucked lower body;
then, in the same edit if stable, raise/shallow the front neckline and shorten
both sleeves. Leave the skirt's minor back-hem residual alone.
\end{verbatim}
\end{promptbox}

The Tailor then calls the \texttt{Fit} tool once per requested size to solve for the parameter values.
For the sleeve end, both sleeve parameters hit their range limit before reaching the target, so the Tailor sets them to the limit and reports the remaining gap in its note.

\begin{promptbox}{Round of the Tailor: five \texttt{Fit} calls and one \texttt{Simulate}}
\scriptsize
\begin{verbatim}
Fit(upper, width, shirt.width, target 15.9)
  -> value 1.0484; probes 1.176 -> 27.9, 1.296 -> 30.7, 1.0484 -> 24.9 %bh
     drafted; predicted 15.9 %bh on the drape
Fit(upper, hem_width, shirt.flare, target 15.3)
  -> value 0.8617; probes 0.997 -> 27.9, 1.132 -> 31.4, 0.8617 -> 24.4;
     predicted 15.3
Fit(upper, neck_depth, collar.fc_depth, target 17.9)
  -> value 0.2786; probes 0.400 -> 23.2, 0.700 -> 27.4, 0.2786 -> 21.5;
     predicted 17.9
Fit(upper, sleeve_end, sleeve.length, target 29.2)
  -> value 0.1; "sleeve.length reaches its range limit at 0.1 with
     upper.sleeve_end at 30.3 %bh against a target of 29.2; another parameter
     is needed for the rest."
Fit(upper, sleeve_end, sleeve.connecting_width, target 29.2)
  -> value 0.0; "sleeve.connecting_width reaches its range limit at 0 with
     upper.sleeve_end at 29.9 %bh against a target of 29.2; another parameter
     is needed for the rest."
Simulate(shirt.width 1.0484, shirt.flare 0.8617, collar.fc_depth 0.2786,
         sleeve.length 0.1, sleeve.connecting_width 0.0)
  note: "Narrow the tee torso and its tucked visible lower body, shallow the
  front neckline to the target, and use the shortest sleeve plus the minimum
  armhole drop to raise the sleeve ends as far as the sleeve parameters
  allow; leave the corrected skirt unchanged."
\end{verbatim}
\end{promptbox}

After simulation, the drape matches the sizes that \texttt{Fit} predicted (Fig.~\ref{fig:appendix_round}, right): the front silhouette IoU rises from $0.851$ to $0.937$, and the Critic judges step 2 better, raising the tee from $6$ to $8$ on proportion and fit, with the neckline width as the main remaining discrepancy.

\section{Texture Generation}
\label{appendix:texture_generation}

\begin{figure}[t]
  \centering
  \includegraphics[width=1.0\textwidth]{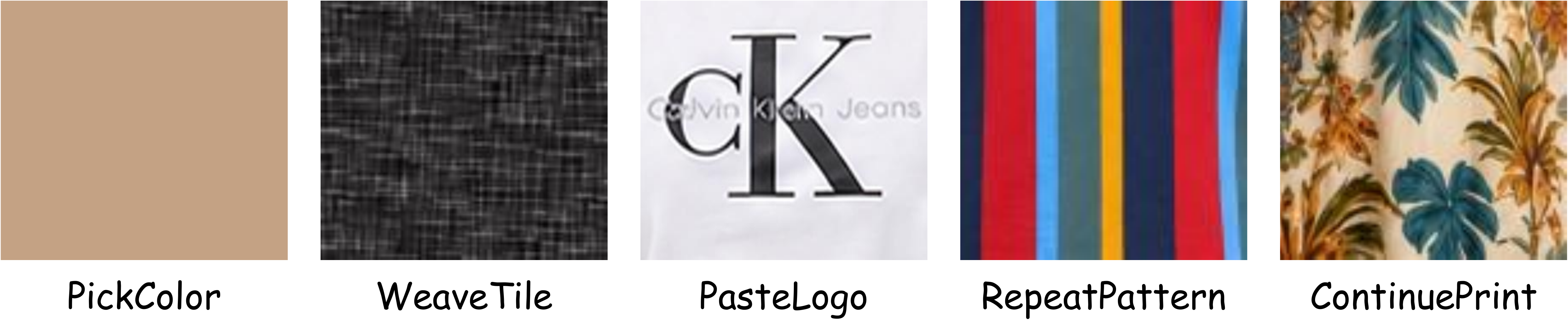}
  \caption{\textbf{Examples of the five texture classes.} From left to right: a solid fabric, a knit without an exact repeat, placed artwork over a base fabric, an exact repeating lattice, and a non-repeating print, each handled by the tool named below it.}
  \label{fig:appendix_texture_example}
\end{figure}

This section provides the implementation details of texture generation (Sec.~\ref{sec:postprocess}).
As in the main text, we first segment each garment with Grounded SAM and map its pixels onto the sewing pattern through the fitted cameras, which gives a partial texture where observed texels hold the colors of the photograph.
However, direct UV mapping introduces boundary artifacts, since the observation is sparse and part of the cloth is occluded by other garments or folds.
Currently, garment texture extrapolation is still an open problem due to the diversity of garment textures.
We therefore design a novel agentic generation method: we divide garment textures into five classes (Fig.~\ref{fig:appendix_texture_example}), design a texture extrapolation tool for each, and let the Layer Planner classify each garment and call the corresponding tool.
Below we describe the routing and the implementation of each tool.

\subsection{Texture Routing}
\label{appendix:tex_route}

We describe the five texture classes to the Layer Planner and let it classify each garment from a crop of the observation, with other garments masked out.
It returns the tool to call, a description of the print, and a box for each piece of placed artwork.
We show the prompt below.

\begin{promptbox}{Texture routing prompt of the Layer Planner}
\scriptsize
\begin{alltt}
You are routing ONE garment's texture to one of five tools.
The image is a crop centered on that garment. Everything that is not the
garment -- other garments, skin, background -- is painted flat gray. Judge
ONLY the unpainted pixels.

Measured evidence:
- per-channel color spread of the observed pixels: \slot{COLOR STATISTICS}
- repeating-pattern autocorrelation: \slot{REPEAT MEASUREMENT}

Tools:
- PickColor: one uniform fabric color. Small shading variation is still solid.
- WeaveTile: textured fabric without a repeating motif (denim, knit, linen).
- RepeatPattern: IDENTICAL small motifs on an exact regular grid (plaid,
  stripes, polka dots). Large organic motifs that vary from one to the next
  are ContinuePrint, even if loosely evenly spaced.
- PasteLogo: any base fabric PLUS distinct placed artwork (a chest logo,
  slogan text). Name the base tool and report each artwork's bounding box.
  A trim, hem, or any band spanning the garment's full width is fabric
  structure, NOT artwork.
- ContinuePrint: one non-repeating print covering most of the garment.

Answer with ONLY a JSON object: the tool, the base tool and artwork boxes
for PasteLogo, and for prints ONE dense sentence a texture generator could
paint from: the motifs and their colors, the background color, the layout
on the garment, and the approximate motif scale.
\end{alltt}
\end{promptbox}

\subsection{Tool Implementation}
\label{appendix:tex_tools}

\textbf{\texttt{PickColor}.}
A solid fabric is described by one color.
\texttt{PickColor} takes the channel-wise median over the observed texels, a robust estimate under shading and stray silhouette pixels, and fills the panels with it.
The other tools reuse the same median as the base color of their fabric.

\textbf{\texttt{WeaveTile}.}
A fabric such as denim, knit, or melange has no exact repeat, so any clean swatch describes it.
We select the swatch among the largest windows inscribed in the observation, preferring the one whose color histogram is closest to the whole garment, which avoids windows in a shadow crease, and rejecting windows observed at grazing angles.
FabricDiffusion~\citep{fabricdiffusion} then turns the swatch into a seamlessly tileable texture.
Since the low-frequency shading of the swatch comes from the drape, we remove it from the tile, and clip generated colors to the contrast range of the swatch.
The tile is then laid over the panels at its physical scale.

\textbf{\texttt{RepeatPattern}.}
For a regular lattice such as plaid or stripes, we measure the tile directly from the observation.
The repeat comes from the autocorrelation above, and the tile spans the full period of the pattern.
Panels whose observed appearance disagrees, such as an argyle front over a solid back, form separate groups with separate tiles.
Each panel then aligns the lattice phase to its own observation, so the lattice stays continuous across seams, and observed texels are blended into it.

\textbf{\texttt{PasteLogo}.}
The base fabric is synthesized first by the tool of its own class, with the texels under the reported artwork marked unobserved, so the fabric continues through the mark.
The mark itself is then identified in the partial texture as a compact region off the base color with sharp printed edges, using the reported box as a soft prior, and pasted back at its exact position.

\textbf{\texttt{ContinuePrint}.}
A print that falls into none of the classes above has no compact rule, so it is continued directly from the observed part.
We use FLUX.1 Fill~\citep{flux} to continue the observed print into unobserved cloth, prompted with the description from the router and a swatch of the fabric.
Since the unobserved part is unknown, we keep the generation inside the observed color gamut.
With $c_1,\dots,c_m$ the palette centers of the observed colors, every generated color $c$ is pulled back toward its nearest center $c_j$:
\begin{equation}
  c\ \mapsto\ c_{j}+\min\!\left(1,\ \frac{\tau}{\lVert c-c_{j}\rVert}\right)(c-c_{j}),
\end{equation}
where $\tau$ is a small tolerance that keeps texture detail and removes invented colors.
Observed texels are kept, and the completed panels are assembled into a texture atlas for rendering.

\section{Additional Experiments}
\label{appendix:additional_experiments}

\subsection{Baseline Implementation}
\label{appendix:baseline_protocol}

Table~\ref{tab:comparison} compares methods of very different design, and here we describe how each baseline is implemented and the protocol shared by all methods.
Each baseline builds on its official implementation and released weights where they are available, and is otherwise reproduced from its paper.

\textbf{SewFormer}~\citep{sewformer} predicts panels, edges, and stitches from one image with a two-level transformer.
We build on the official implementation and checkpoint, and run its full pipeline from the pattern prediction to its own draping of the result.
For texture generation, its release binds one fabric image to each garment through the panel layout as UV, with the fabric chosen from a library.
We keep this texturing and only replace the library lookup by a swatch cut from the garment region of the photo, so the texture follows the input.

\textbf{DressCode}~\citep{dresscode} generates a sewing pattern autoregressively from text and adds a tiled PBR texture.
We build on the official checkpoints of both its pattern and texture models.
Since it takes text rather than an image, we follow its interactive mode: its own prompt turns the photo into the shape and texture descriptions it expects, using the same language model as our agents, and the rest of its pipeline runs unchanged.
Its garments are draped with its own pipeline and body.
For texture generation, we use its own texture model with the fabric description from the same prompt.

\textbf{ChatGarment}~\citep{chatgarment} fine-tunes a vision-language model to write a GarmentCode program from the image.
We build on the official checkpoint and its two-step image mode, which first describes the outfit in text and then writes the program from the text and the image.
The program is then compiled and draped with its own pipeline on the mean body it targets without texture, since it does not support texture generation.

\textbf{GarmentGPT}~\citep{garmentgpt} tokenizes panel boundaries into a discrete codebook and decodes them with a vision-language model.
We build on the official checkpoints and its single-image inference script, and drape the decoded pattern with its own pipeline.
It does not generate texture.

\textbf{ISP}~\citep{isp} represents a garment as an implicit sewing pattern with a neural draping model, and fits it to an image by optimizing the mask overlap, one garment at a time from the outermost layer inward.
We build on the official implementation and checkpoints, with its own body estimation and segmentation, and keep its optimization schedule.
It does not generate texture.

\textbf{GarmentRecovery}~\citep{garmentrecovery} places a learned deformation prior on top of the ISP shape prior, and fits both to the image by optimization.
We build on the official implementation, including its preprocessing for normals, body, and masks, and run its per-category fitting scripts with their default iteration counts for each garment in the outfit.
It does not generate texture.

\textbf{Dress 1-to-3}~\citep{dress123} initializes a pattern from SewFormer, generates multi-view images with a diffusion model, and refines the pattern against them through differentiable simulation.
Its code is not public, so we reproduce the full pipeline from the paper as faithfully as we can, with the public models it builds on for the initial pattern, multi-view generation, and body estimation, followed by the differentiable refinement with the settings in the paper.
For texture generation, we follow the paper and tile a swatch from the photo with FabricDiffusion~\citep{fabricdiffusion}.

We note that every method, including ours, receives the same single input: the photograph in the wild, or the front rendering of the scan on CloSe and 4D-Dress.
From there, each method estimates the sewing pattern and drapes the garments, with its own pipeline and simulator.
We measure all times in Table~\ref{tab:comparison} on the same workstation, with one NVIDIA RTX PRO 6000 GPU per method and one case at a time.
Time is the wall clock from the input image to the final garment mesh.

\subsection{Quantitative Comparison on Texture Generation}
\label{appendix:texture_comparison}

Here, we provide additional quantitative results for texture generation, using the in-the-wild photos from our user study (Sec.~\ref{sec:results}).
We compare with the baselines that output textured garments: SewFormer~\citep{sewformer}, DressCode~\citep{dresscode}, and Dress 1-to-3~\citep{dress123}.
We measure the following metrics.

\begin{itemize}[leftmargin=*]
  \item \textbf{$\Delta E$} measures how far the garment colors in a render are from those in the photo.
  We cluster the garment pixels of each image into a small color palette, and compute the Earth Mover's Distance~\citep{emd} between the two palettes, with CIEDE2000~\citep{ciede2000} as the perceived difference between two colors.
  \item \textbf{CLIP-I} and \textbf{DINO} are the cosine similarities between the features of the two garment crops, extracted by CLIP~\citep{clip} and DINOv2~\citep{dinov2}.
  CLIP captures the overall semantics, and DINOv2 is more sensitive to prints and fine texture.
  \item \textbf{PSNR}, \textbf{SSIM}~\citep{ssim}, and \textbf{LPIPS}~\citep{lpips} measure the pixel error, the structural similarity, and the learned perceptual distance between two pixel-aligned images.
\end{itemize}

For a comprehensive comparison, we evaluate in two settings.
In the canonical A-pose, the render shares no pixel grid with the photo, so we report only $\Delta E$, CLIP-I, and DINO.
In the target pose, we repose each garment to the body in the photo and render it through the camera recovered from the photo, so the render lands on the person.
This lets us add the pixel-aligned metrics.
All metrics are computed only inside the garment region, so the body and the background never affect them.
The results are shown in Table~\ref{tab:texture}.

\begin{table}[!ht]
  \centering
  \caption{\textbf{Quantitative comparison on texture generation.} We evaluate the texture fidelity of each method against the input photo on different metrics. Dress 1-to-3 reconstructs the garment directly in the target pose, so we do not report its results in the canonical A-pose.}
  \label{tab:texture}
  \resizebox{\textwidth}{!}{
  \begin{tabular}{@{}lccccccccc@{}}
  \toprule
   & \multicolumn{3}{c}{\textsc{Canonical A-pose}} & \multicolumn{6}{c}{\textsc{Target Pose}} \\ \cmidrule(lr){2-4} \cmidrule(lr){5-10}
  \multirow{-2}{*}{\textsc{Method}} & $\Delta E$~$\downarrow$ & \textsc{CLIP-I}~$\uparrow$ & \textsc{DINO}~$\uparrow$ & \textsc{PSNR}~$\uparrow$ & \textsc{SSIM}~$\uparrow$ & \textsc{LPIPS}~$\downarrow$ & $\Delta E$~$\downarrow$ & \textsc{CLIP-I}~$\uparrow$ & \textsc{DINO}~$\uparrow$ \\ \midrule
  SewFormer & 20.61 & 0.661 & 0.252 & 8.39 & 0.543 & 0.433 & 14.05 & 0.681 & 0.324 \\
  DressCode & 9.67 & 0.633 & 0.305 & 8.68 & 0.565 & 0.407 & 12.93 & 0.652 & 0.275 \\
  Dress 1-to-3 & - & - & - & 8.26 & 0.531 & 0.410 & 21.26 & 0.650 & 0.229 \\ \midrule
  \rowcolor{blue!10}
  Ours & \textbf{9.57} & \textbf{0.821} & \textbf{0.658} & \textbf{9.94} & \textbf{0.615} & \textbf{0.352} & \textbf{8.43} & \textbf{0.800} & \textbf{0.597} \\ \bottomrule
  \end{tabular}
  }
\end{table}

Our method achieves the best results on all metrics in both settings.
We observe that the gap is largest on DINO, which responds most to prints and logos.
The baselines cover each garment with a tiled fabric swatch, which keeps a plausible average color but loses the print.

\subsection{Quantitative Comparison on Layer Structure}
\label{appendix:layer_comparison}

Multi-layer reconstruction is one of the central claims of our work, so here we measure the layer structure directly.
We compare with the methods that model more than one layer: DressCode~\citep{dresscode}, ISP~\citep{isp}, and GarmentRecovery~\citep{garmentrecovery}.
For a comprehensive comparison, we evaluate on both synthetic data and real photos.
We randomly sample 200 outfits from the GarmentCodeData-XL test set, with 50 outfits for each of 1 to 4 layers, where the ground-truth layer number and structure come with the data.
DeepFashion2~\citep{DeepFashion2} has no such labels, so we select 60 photos by hand, with 20 for each of 1 to 3 layers (since people rarely wear more than three layers), and compare the results against the photo by human judgment.
We measure two aspects of the layer structure:
\emph{Layer Count Acc.} is the fraction of outfits whose number of layers matches the ground truth.
\emph{Layer Structure Acc.} is conditional on a correct count: among the outfits whose count is correct, it is the fraction whose inner-to-outer order and relations between garments (\textit{e.g.}, tucked into, worn over) also match the ground truth.
We define it this way because an outfit with a wrong number of layers cannot have the right structure, so the two metrics separate the two kinds of errors.
The results are shown in Table~\ref{tab:layer}.

\begin{table}[!ht]
  \centering
  \caption{\textbf{Quantitative comparison on layer structure.} We report the accuracy of the layer count, and the accuracy of the layer structure conditional on a correct count, on GarmentCodeData-XL and DeepFashion2. Only the methods that can model more than one layer are included.}
  \label{tab:layer}
  \resizebox{0.95\textwidth}{!}{
  \begin{tabular}{@{}lcccc@{}}
  \toprule
   & \multicolumn{2}{c}{\begin{tabular}[c]{@{}c@{}}\textsc{GarmentCodeData-XL}\\\textsc{(max 4 layers)}\end{tabular}} & \multicolumn{2}{c}{\begin{tabular}[c]{@{}c@{}}\textsc{DeepFashion2}\\\textsc{(max 3 layers)}\end{tabular}} \\ \cmidrule(lr){2-3} \cmidrule(lr){4-5}
  \multirow{-2}{*}{\textsc{Method}} & \textsc{Count Acc.}~$\uparrow$ & \textsc{Struct. Acc.}~$\mid$~\textsc{Count}~$\uparrow$ & \textsc{Count Acc.}~$\uparrow$ & \textsc{Struct. Acc.}~$\mid$~\textsc{Count}~$\uparrow$ \\ \midrule
  DressCode & 34.0\% & 76.5\% & 31.7\% & 52.6\% \\
  ISP & 50.0\% & 87.0\% & 65.0\% & 84.6\% \\
  GarmentRecovery & 49.5\% & 85.9\% & 61.7\% & 83.8\% \\ \midrule
  \rowcolor{blue!10}
  Ours & \textbf{95.5\%} & \textbf{100.0\%} & \textbf{96.7\%} & \textbf{98.3\%} \\ \bottomrule
  \end{tabular}
  }
\end{table}

Our method recovers the most correct number of layers and the structure, on both synthetic data and in-the-wild photos.
This robustness comes from the Layer Planner, whose skills turn the perception ability of the base model into an explicit layer structure before any pattern is estimated.
GarmentRecovery cannot recognize more than two layers, and its count accuracy is zero on the three-layer and four-layer outfits. 
ISP also struggles beyond two layers, though it models the order and relations slightly better than GarmentRecovery.
DressCode behaves very differently on synthetic data and real photos.
It often misjudges the number of layers, and after reposing its garments often interpenetrate in simulation, which obscures the layer order (see Fig.~\ref{fig:comparison}).

We note that our agents and skills support arbitrary number of layers in a zero-shot way.
The Layer Planner infers the number of layers from the photo, and each layer is predicted and refined from innermost to outermost, so the task is well defined for varying number of layers.
In everyday clothing, people rarely wear more than three layers, as we observe in DeepFashion2, so we evaluate outfits with up to four layers throughout the paper.
This range covers what people wear and what garment production serves, and we consider it sufficient to support the conclusion on generalization.

\subsection{Faithfulness of Synthesized Views}
\label{appendix:view_faithfulness}

To provide sufficient supervision for every layer and to support zero-shot reconstruction, Stage 1 synthesizes the canonical front $I^{\mathrm{f}}$, the back $I^{\mathrm{b}}$, and the peeled views of inner layers with generation models.
In this way, our method builds on the strong prior of modern image generation models.
We verify their faithfulness here, and ablate their effect on the final recovery in Appendix~\ref{appendix:ablation_views}.

\textbf{Faithfulness to the photograph.}
To measure how faithful a synthesized view is, we compare the canonical front with the photograph it comes from, on the same CloSe and 4D-Dress datasets as Table~\ref{tab:comparison}.
4D-Dress records each outfit as a multi-frame sequence, so for reference, a real second photograph of the same outfit, taken from another frame of the same capture, gives the similarity of a faithful view.
On both datasets, the photographs of all other outfits give the similarity of an unrelated garment.
We crop the garment region of every image with a SegFormer clothes parser~\citep{segformer}, and compare the crops with CLIP-I, DINO, and $\Delta E$ from Appendix~\ref{appendix:texture_comparison}.
Since the images differ in pose, we leave out pixel-aligned metrics.

\begin{table}[!ht]
  \centering
  \caption{\textbf{Faithfulness of the canonical view.} We compare garment crops with the input photograph. \emph{Second photo} is another frame of the same 4D-Dress capture, a real view of the same outfit. \emph{Other outfits} compares the canonical view with the photographs of all other outfits in the dataset.}
  \label{tab:view_faithful}
  \resizebox{0.75\textwidth}{!}{
  \begin{tabular}{@{}lcccccc@{}}
  \toprule
   & \multicolumn{3}{c}{\textsc{CloSe}} & \multicolumn{3}{c}{\textsc{4D-Dress}} \\ \cmidrule(lr){2-4} \cmidrule(lr){5-7}
  \multirow{-2}{*}{\textsc{Comparison}} & \textsc{CLIP-I}~$\uparrow$ & \textsc{DINO}~$\uparrow$ & $\Delta E$~$\downarrow$ & \textsc{CLIP-I}~$\uparrow$ & \textsc{DINO}~$\uparrow$ & $\Delta E$~$\downarrow$ \\ \midrule
  Other outfits vs.\ photo & 0.869 & 0.678 & 7.55 & 0.705 & 0.391 & 23.29 \\
  Second photo vs.\ photo & - & - & - & 0.917 & 0.776 & 3.46 \\ \midrule
  \rowcolor{blue!10}
  Canonical view vs.\ photo & 0.907 & 0.845 & 5.49 & 0.806 & 0.729 & 4.31 \\ \bottomrule
  \end{tabular}
  }
\end{table}

As shown in Table~\ref{tab:view_faithful}, the canonical view preserves both the colors and the identity of the garment.
In color, its distance to the photograph is close to that of a real second photograph, and far below that of other outfits.
In features, it is also clearly closer to its own photograph than to the photographs of other outfits.
The margin is smallest on CloSe, where most outfits are a single layer with a top and pants.
In this way, we show that the canonical view stays faithful to the garment, and serves as clear supervision for our pipeline, free of the occlusion, unusual poses, and motion-driven deformation in the photograph, with the cloth at rest under gravity.

\subsection{Stress Test: Refinement from a Plain T-shirt}
\label{appendix:stress_tshirt}

To fully demonstrate the capability of Stage 2, we conduct an interesting stress test: from a plain T-shirt to various styles.
Here we replace the coarse program with the same plain T-shirt on every case and ask the agentic trio to optimize the outfit in the photo from there.

\begin{figure}[!ht]
  \centering
  \includegraphics[width=1.0\textwidth]{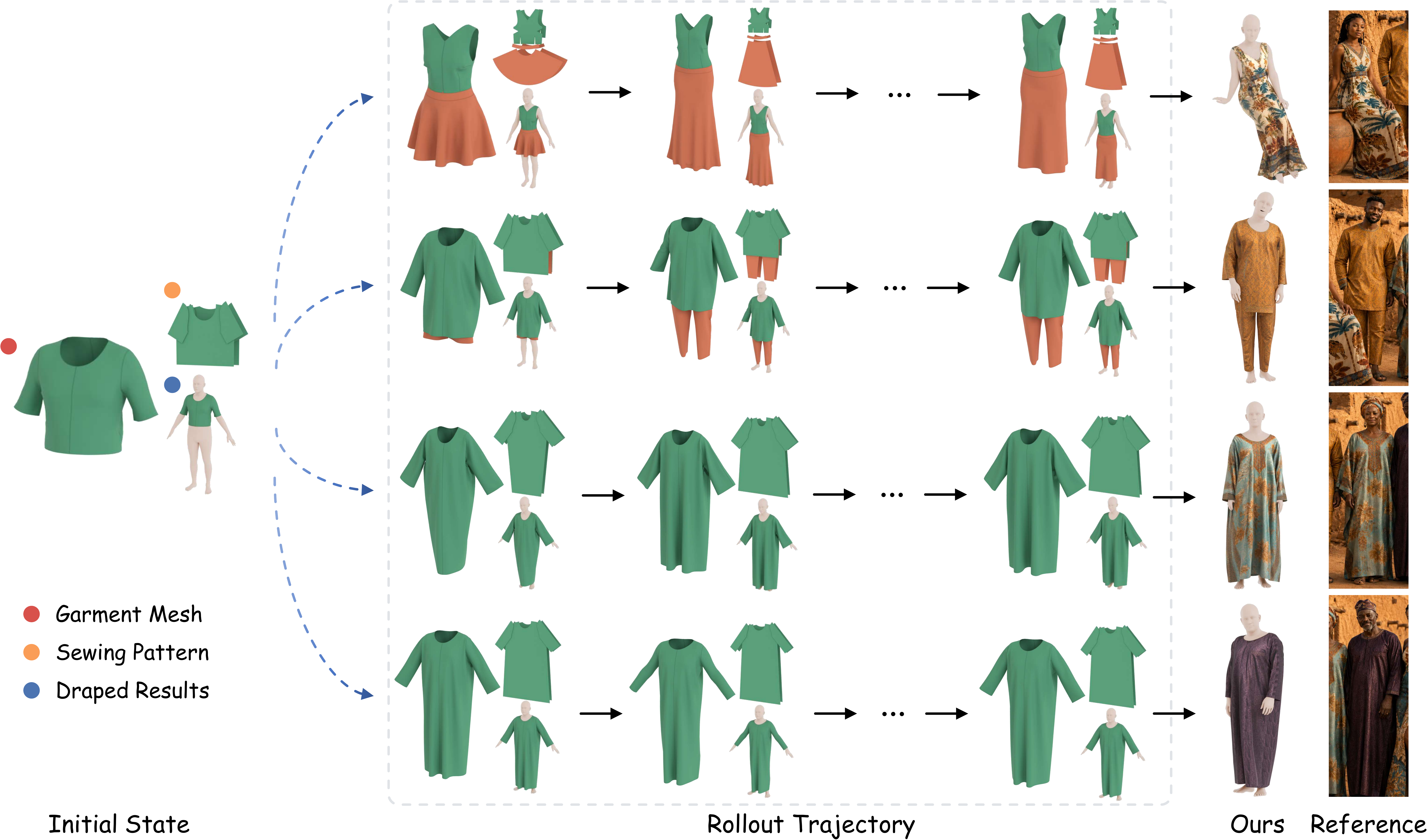}
  \caption{\textbf{From one T-shirt to any garment.} We show four examples of refinement runs that start from the same plain T-shirt (left). The trio first intelligently modifies the construction, then refines the sizes to match the photo.}
  \label{fig:appendix_tshirt_to_all}
\end{figure}

For every case, we discard the program estimated by Stage 1 and replace it with a plain T-shirt.
Everything else stays as in the main experiments, including the canonical views, the recovered body, the layer plan, the backbone, the budget of six simulations, and the agent trio.
The trio thus starts from the same T-shirt on every outfit and has to rebuild the garments from this program.
Fig.~\ref{fig:appendix_tshirt_to_all} shows four such runs on in-the-wild photos.
In every case the first edit changes the construction, and refine the sizes in the later steps.

\begin{figure}[t]
  \centering
  \includegraphics[width=0.5\textwidth]{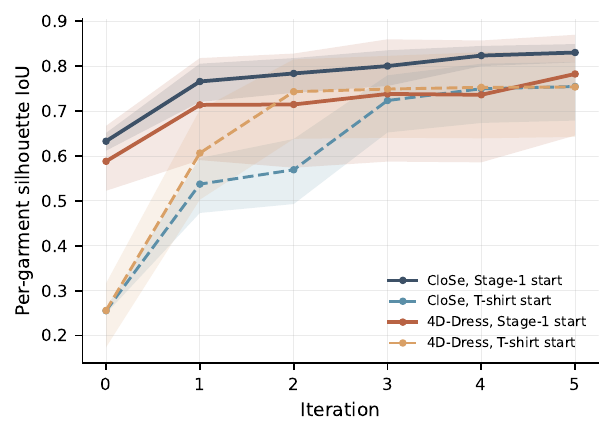}
  \caption{\textbf{Refinement from a plain T-shirt.} Silhouette IoU on CloSe and 4D-Dress over the refinement iterations, starting from the Stage-1 program (solid) or from a plain T-shirt (dashed). Bands show the 95\% bootstrap confidence interval.}
  \label{fig:appendix_tshirt_curve}
\end{figure}

We also run this setting on the real scans of CloSe and 4D-Dress, with the same cases and metrics as Table~\ref{tab:comparison}.
Fig.~\ref{fig:appendix_tshirt_curve} shows the per-garment silhouette IoU of the final garment after each simulation, and Table~\ref{tab:tshirt_stress} reports the geometry of the reposed garment against the scan.

\begin{table}[t]
  \caption{\textbf{Refinement from a plain T-shirt.} Comparison of the geometry of the reposed garment against the scan, with the protocol of Table~\ref{tab:comparison}.}
  \label{tab:tshirt_stress}
  \centering
  \footnotesize
  \resizebox{0.75\textwidth}{!}{
  \begin{tabular}{@{}lcccc@{}}
  \toprule
   & \multicolumn{2}{c}{\textsc{Geometry (CloSe)}} & \multicolumn{2}{c}{\textsc{Geometry (4D-Dress)}} \\ \cmidrule(lr){2-3} \cmidrule(lr){4-5}
  \multirow{-2}{*}{\textsc{Starting point}} & \textsc{CD}~$\downarrow$ & \textsc{IoU}~$\uparrow$ & \textsc{CD}~$\downarrow$ & \textsc{IoU}~$\uparrow$ \\ \midrule
  From Plain T-shirt & 1.75 & 0.699 & 1.31 & 0.871 \\ \midrule
  \rowcolor{blue!10}
  From Stage-1 Program (Ours) & \textbf{1.20} & \textbf{0.764} & \textbf{1.21} & \textbf{0.880} \\ \bottomrule
  \end{tabular}
  }
\end{table}

The results show that starting from the T-shirt, the agentic trio rebuilds the outfit close to the full pipeline.
After six simulations, the silhouette IoU rises to $0.754$ on 4D-Dress and $0.755$ on CloSe, against $0.783$ and $0.831$ from the Stage-1 program.
The trio first rewrites the construction and adds the missing garments, then adjusts the sizes, while the Critic and the checkpoints keep the loop from drifting.
The agentic trio thus recovers both the style and the sizes through closed-loop refinement, which shows its strong capability.

The results also confirm the importance of Stage 1.
The coarse program provides a reasonable and good initial solution, which fixes the construction from the photo in a single forward pass.
From the T-shirt, the loop has to find the construction by trial, so the first two or three simulations go to it, and the curve lags the full pipeline by about two iterations and ends $0.03$ to $0.08$ lower at the same budget.
With the coarse program, the loop starts close to the target and spends almost all its budget on the fit, so it converges to a better result in fewer iterations.
The closed loop is thus the engine of accuracy, and the coarse estimate is the warm start that saves the budget.

\subsection{Additional Visual Results}
\label{appendix:more_results}

\begin{figure}[t]
  \centering
  \includegraphics[width=1.0\textwidth]{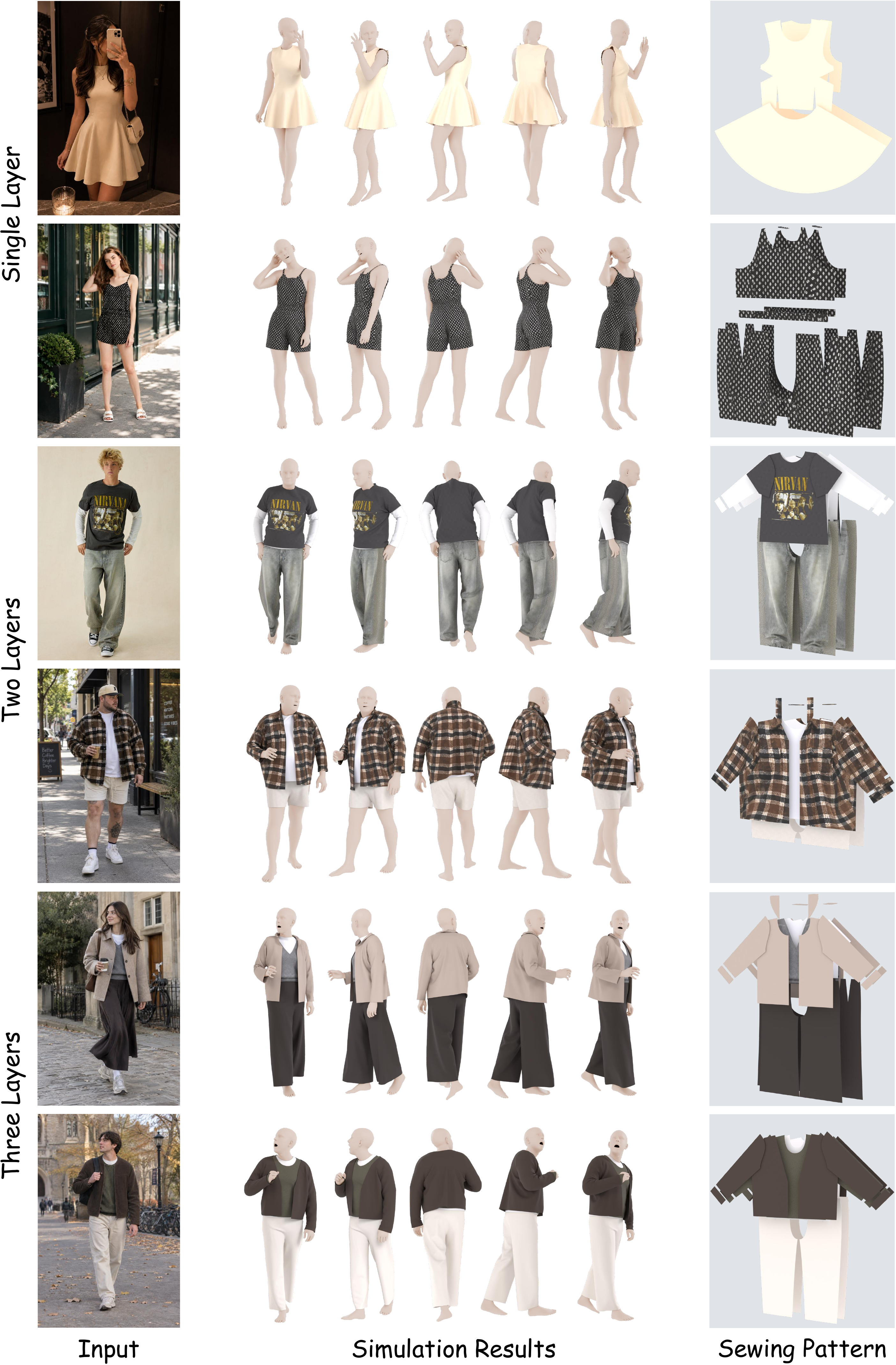}
  \caption{Additional results on in-the-wild photos with different layering, poses, and body shapes.}
  \label{fig:appendix_more_results}
\end{figure}

Here, we present more visual results of our method on in-the-wild photos, covering different layering, poses, and body shapes (Fig.~\ref{fig:appendix_more_results}).

\section{Ablation Study}
\label{appendix:ablation_study}

\subsection{Ablation on Synthesized Views}
\label{appendix:ablation_views}

To validate the source of supervision, we replace our synthesized views with weaker supervision and measure the final recovery against real scans, on the same CloSe and 4D-Dress datasets as Table~\ref{tab:comparison}.
We compare four sources of supervision:
\emph{Input only} supervises the agents with the photograph alone, with no canonical view, back view, or peeling.
\emph{Front only} keeps the canonical front and the peeling, and removes every back view.
\emph{No peeling} keeps the layer plan, and reads and refines every layer against the original canonical A-pose views.
\emph{Full} is our method.
All settings share the recovered body and the budget of six simulations.

\begin{table}[t]
  \centering
  \caption{\textbf{Ablation on synthesized views.} We report the geometry of the reposed garment against the scan, graded as in Table~\ref{tab:comparison}. Since CloSe has few multi-layer outfits, while 4D-Dress captures outfits with an outer garment and labels it as its own class, we test no peeling only on 4D-Dress.}
  \label{tab:view_ablation}
  \setlength{\tabcolsep}{18pt}
  \resizebox{0.6\textwidth}{!}{
  \begin{tabular}{@{}lcccc@{}}
  \toprule
   & \multicolumn{2}{c}{\textsc{CloSe}} & \multicolumn{2}{c}{\textsc{4D-Dress}} \\ \cmidrule(lr){2-3} \cmidrule(lr){4-5}
  \multirow{-2}{*}{\textsc{Setting}} & \textsc{CD}~$\downarrow$ & \textsc{IoU}~$\uparrow$ & \textsc{CD}~$\downarrow$ & \textsc{IoU}~$\uparrow$ \\ \midrule
  Input only & 1.36 & 0.743 & 2.28 & 0.782 \\
  Front only & 1.29 & 0.750 & 1.73 & 0.833 \\
  No peeling & - & - & 1.64 & 0.846 \\ \midrule
  \rowcolor{blue!10}
  Full & \textbf{1.20} & \textbf{0.764} & \textbf{1.21} & \textbf{0.880} \\ \bottomrule
  \end{tabular}
  }
\end{table}

As shown in Table~\ref{tab:view_ablation}, each synthesized view improves the recovery.
The canonical view brings the largest gain, and we find that the gain grows with the difficulty of the pose.
It is modest on CloSe, whose inputs are standing scans already close to the A-pose, and large on the dance poses of 4D-Dress.
The back view lowers the error further, which shows that the generated back carries real information about the garment.
Peeling further improves the recovery by providing supervision for every layer, which is essential for multi-layer outfits.
In this way, we show that our synthesized views are an effective source of supervision, and each of them brings the recovery closer to the real garment.
Together with Appendix~\ref{appendix:view_faithfulness}, we show that the synthesized views are faithful to the photograph and serve as reliable supervision for our pipeline.

\subsection{Ablation on Agentic Trio Design}
\label{appendix:ablation_trio}

\textbf{w/o Rollback Tool.}
\texttt{Rollback} is a key control tool of our agentic trio.
It returns the sewing pattern to an earlier checkpoint and keeps the failed attempt in memory as a reference for later edits (Sec.~\ref{sec:stage2}).
When an edit overshoots the target or compromises physical plausibility, the Orchestrator rolls it back at once.
This keeps the refinement from optimizing further in a wrong region of the design space.
To ablate it, we remove \texttt{Rollback}, so every edit builds on the latest step and the last step is delivered.
Both variants start from the same coarse pattern.
As shown in Table~\ref{tab:ablation_rollback}, without \texttt{Rollback} a wrong construction edit or a tangled drape is carried to the end.
Structure accuracy and IoU drop, and the delivered drapes carry four times as many self-intersecting edges.

\begin{table}[!ht]
  \caption{\textbf{Ablation on Rollback.} We ablate the \texttt{Rollback} tool of the agentic trio. Self-intersection is the share of cloth edges that intersect the garment itself on the delivered drape.}
  \label{tab:ablation_rollback}
  \centering
  \footnotesize
  \resizebox{0.8\textwidth}{!}{
  \begin{tabular}{@{}lccc@{}}
  \toprule
   & \textsc{GarmentCodeData-XL} & \textsc{DeepFashion2} & \\ \cmidrule(l){2-2} \cmidrule(l){3-3}
  \multirow{-2}{*}{\textsc{Comparison}} & \textsc{Struct. Acc.}~$\uparrow$ & \textsc{IoU}~$\uparrow$ & \multirow{-2}{*}{\textsc{Self-Intersection}~(\textperthousand)~$\downarrow$} \\ \midrule
  w/o Rollback & 58.70 & 0.736 & 0.16 \\ \midrule
  \rowcolor{blue!10}
  Full & \textbf{73.59} & \textbf{0.839} & \textbf{0.04} \\ \bottomrule
  \end{tabular}
  }
\end{table}

\subsection{Ablation on Single-Agent Refinement}
\label{appendix:ablation_single}

Our refinement is built on an agentic trio with complementary roles.
The Tailor edits the program, the Critic judges the drape, and the Orchestrator controls the session, and each agent keeps its own tools and its own memory (Appendix~\ref{appendix:s2_workflow}).
This decomposition separates editing, evaluation, and control, so an edit is judged by an independent Critic and a bad step is rolled back to a previous checkpoint.
To measure how much this design contributes, we ablate it and hand the whole refinement to a single agent that judges its own drape and rewrites the program by itself.

Since no prior work provides such a baseline for single-agent garment refinement, we build one on TextGrad~\citep{textgrad}, which gives the single-agent loop a clean and well-defined form.
TextGrad treats an LLM system as a computation graph and optimizes a variable with \emph{textual gradients}, which replace the numeric gradients of backpropagation with natural-language feedback.
In our setting, the variable is the program $P^{t}$ of the current layer, and the forward pass is the simulation that turns it into an observation $O^{t}$.
A VLM first evaluates this observation against the reference views and writes the loss in natural language:
\begin{equation}
  \mathcal{L}^{t} = \mathrm{VLM}\big(O^{t},\ I^{\mathrm{f}}_k,\ I^{\mathrm{b}}_k\big),
\end{equation}
where $\mathcal{L}^{t}$ is a textual evaluation that names what differs between the drape and the photograph.
The backward pass then turns this evaluation into feedback on the variable, the textual gradient:
\begin{equation}
  \frac{\partial \mathcal{L}^{t}}{\partial P^{t}} = \nabla_{\mathrm{VLM}}\big(P^{t},\ O^{t},\ \mathcal{L}^{t}\big),
\end{equation}
where $\nabla_{\mathrm{VLM}}$ is a VLM call that reads the program, its observation, and the loss, and writes which parameters to change and how.
Finally, a textual gradient descent step rewrites the variable according to the feedback:
\begin{equation}
  P^{t+1} = \mathrm{TGD.step}\Big(P^{t},\ \frac{\partial \mathcal{L}^{t}}{\partial P^{t}}\Big),
\end{equation}
which is a VLM call that outputs the updated parameter map of $P^{t}$, with the same tools available as to our Tailor.
All three calls are made by the same agent, which therefore judges its own drape and edits its own program.

For a fair comparison, both methods start from the same coarse pattern and use the same simulator, backbone, observations, tools, memory, and budget of ten simulations per layer.
The only difference is the role decomposition: the single agent evaluates its own drape, plans its next edit, and edits the program within one context, where the trio separates these into the Critic, the Orchestrator, and the Tailor.
We further implement two variants that differ in which step is delivered.
The plain variant delivers the last iterate, as TextGrad does.
The self-selection variant lets the agent score each drape itself and delivers its best-scored step, which is the most generous reading of a single agent.
We evaluate both on the real scans of CloSe and 4D-Dress to show robustness to in-the-wild photos, the same setting as Table~\ref{tab:comparison}.

\begin{figure}[!ht]
  \centering
  \includegraphics[width=0.95\textwidth]{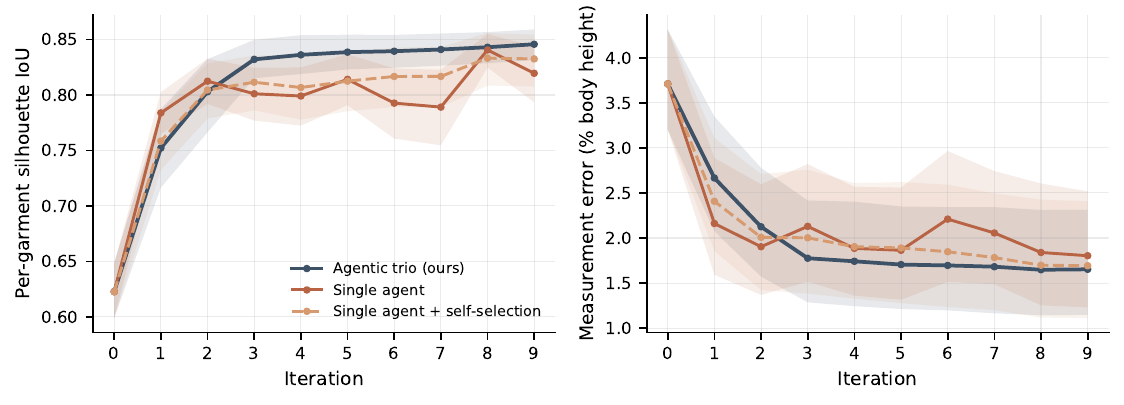}
  \caption{\textbf{Ablation on single-agent refinement.} Per-garment silhouette IoU and measurement error of the delivered garment after each simulation on CloSe and 4D-Dress. Bands show the 95\% bootstrap confidence interval. The single agent oscillates and plateaus early, while our agentic trio converges progressively.}
  \label{fig:ablation_single_curve}
\end{figure}

\begin{table}[!ht]
  \caption{\textbf{Ablation on single-agent refinement.} All methods use the same coarse pattern, backbone, tools, memory, and 10 simulations per layer. Measurement error is in percent of body height. Tokens and time are the output tokens and VLM time per case. We find that the single agent is worse in quality, biased in self-evaluation, and higher in cost.}
  \label{tab:ablation_single}
  \centering
  \footnotesize
  \resizebox{0.9\textwidth}{!}{
  \begin{tabular}{@{}lccccc@{}}
  \toprule
  \textsc{Method} & \textsc{IoU}~$\uparrow$ & \textsc{Meas. Err.}~$\downarrow$ & \textsc{Edits / Iter.} & \textsc{Tokens (k)}~$\downarrow$ & \textsc{Time (s)}~$\downarrow$ \\ \midrule
  Single agent & 0.820 & 1.80\% & 6.2 & 38.7 & 748 \\
  Single agent + self-selection & 0.833 & 1.69\% & 6.2 & 38.7 & 748 \\ \midrule
  \rowcolor{blue!10}
  Agentic trio (ours) & \textbf{0.846} & \textbf{1.65\%} & \textbf{2.8} & \textbf{15.0} & \textbf{412} \\ \bottomrule
  \end{tabular}
  }
\end{table}

Fig.~\ref{fig:ablation_single_curve} and Table~\ref{tab:ablation_single} show the results, which reveal three problems of the single agent.

\textbf{Worse quality.}
The single agent plateaus early and oscillates: its IoU reaches $0.812$ after three simulations and stays at $0.820$ after ten, while our trio keeps improving to $0.846$ with lower measurement error.
Between consecutive steps, its IoU moves by $0.063$ on average, twice the $0.035$ of the trio.
This oscillation comes from its edits: although free to change any subset of the parameters, the single agent touches more than twice as many per iteration as the trio and reverses its own edits more often, since the textual evaluation describes the current drape but not how far each parameter moved it.
Our agentic trio avoids this:
The Critic reports a few discrepancies with measured evidence, the Orchestrator picks which of them to address in this round, and the Tailor edits only those parameters, so every round makes a small, targeted change and the loop converges progressively with previous memory for reference.

\textbf{Self-bias.}
The single agent evaluates the drape, derives the feedback, and rewrites the program with the same model, so every edit is judged by its own author.
The self-selection variant isolates this judgement: it sees every step of the run with the same views and measurements the Critic sees, and differs from the trio only in who scores them.
Selecting by its own scores raises the IoU from $0.820$ to $0.833$, only half of the way to the $0.846$ of the trio, so the remaining gap comes from the judgement itself, not from the evidence.
This is consistent with the self-bias in Sec.~\ref{sec:stage2}~\citep{selfbias,selfpreference,selfcorrect}: a model tends to favor its own outputs, so the agent that made an edit is a poor judge of it, and an independent Critic provides exactly the judgement it lacks.

\textbf{Higher cost.}
Every iteration of the single agent is three long generations, an evaluation, a textual gradient, and a rewrite of the full parameter map, so it emits $2.6\times$ the output tokens and spends $1.8\times$ the VLM time of the trio, and it still needs more iterations for lower quality.
The calls of the trio are short, since each role answers one well-posed question through structured function calling~\citep{toolformer} and the tools carry the arithmetic.
The fixed system prompt and tool schema of each agent also form a stable prefix across rounds, which prompt caching~\citep{promptcache} reuses, and older views are compacted into text notes so the context stays bounded (Appendix~\ref{appendix:s2_workflow}).

\subsection{Ablation on Texture Generation}
\label{appendix:ablation_texture}

\begin{figure}[t]
  \centering
  \includegraphics[width=0.8\textwidth]{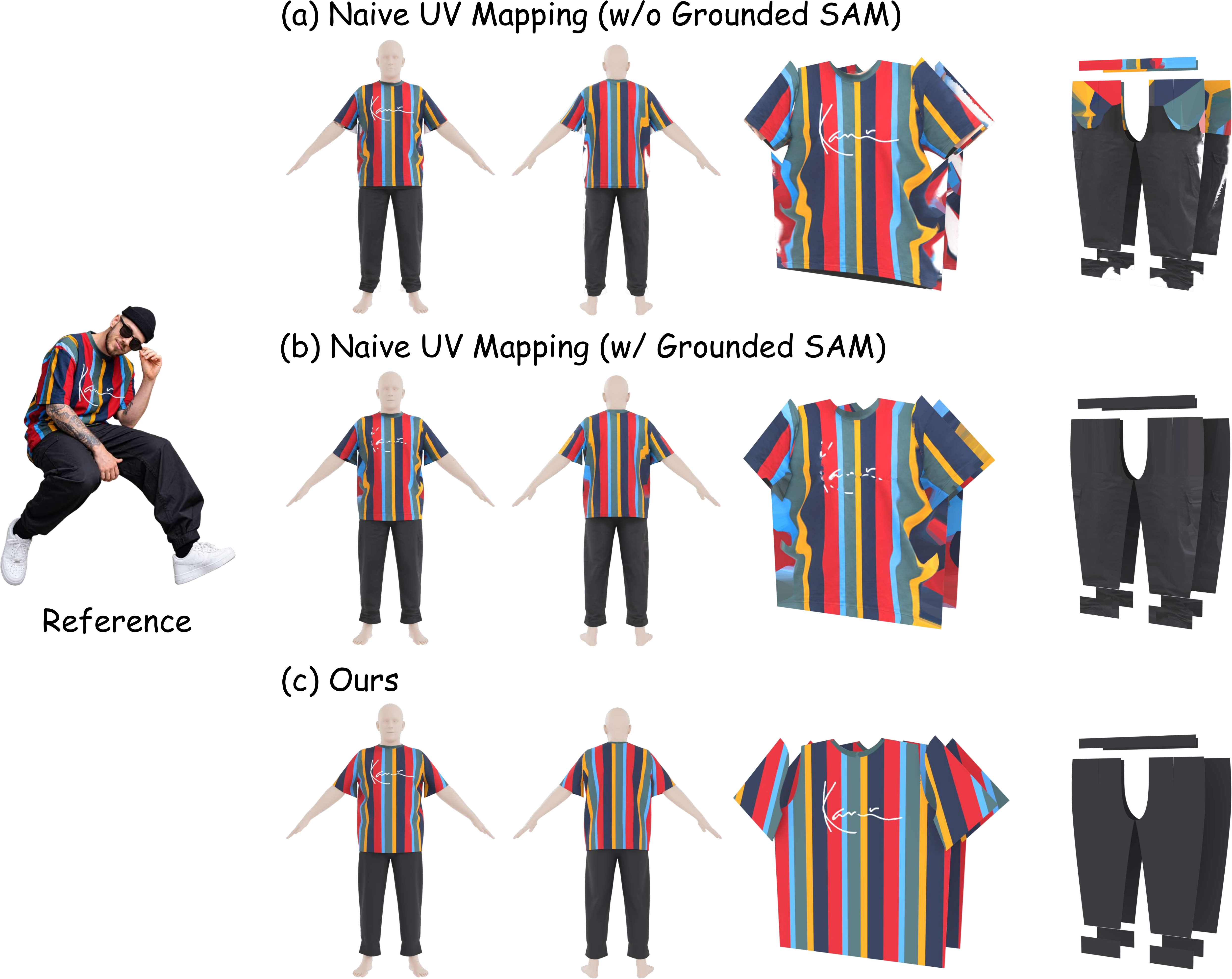}
  \caption{\textbf{Ablation on texture generation.} (a) Naive UV mapping without Grounded SAM mixes garments: the stripes of the tee bleed onto the pants. (b) With Grounded SAM, the garments separate, but unobserved cloth admits no extrapolation and the observed print lands warped. (c) Our agentic generation method continues the stripes at their measured repeat, restore the logo, and keep the pants a uniform black.}
  \label{fig:appendix_texture_ablation}
\end{figure}

We ablate our texture generation with two variants of naive UV mapping (Fig.~\ref{fig:appendix_texture_ablation}): direct back-projection without the instance masks of Grounded SAM, and with the masks restored.

Without the masks, every pixel behind the drape silhouette counts as texture.
Colors therefore bleed across garments: the stripes of the tee print onto the pants, and residual skin and background enter along every silhouette. 

With the masks, the garments separate cleanly, yet the projection is still a fixed mapping.
The cloth no camera saw admits no extrapolation at all, and even the observed print lands warped on the flat panels, since the folds and foreshortening of the drape are baked into every sample.

Our method instead divides garment textures into five structured classes, each handled by a carefully designed extrapolation tool.
Letting the agent judge the class itself keeps the stage robust across in-the-wild garments.

\section{User Study}
\label{appendix:user_study}

\begin{figure}[!ht]
  \centering
  \includegraphics[width=0.8\textwidth]{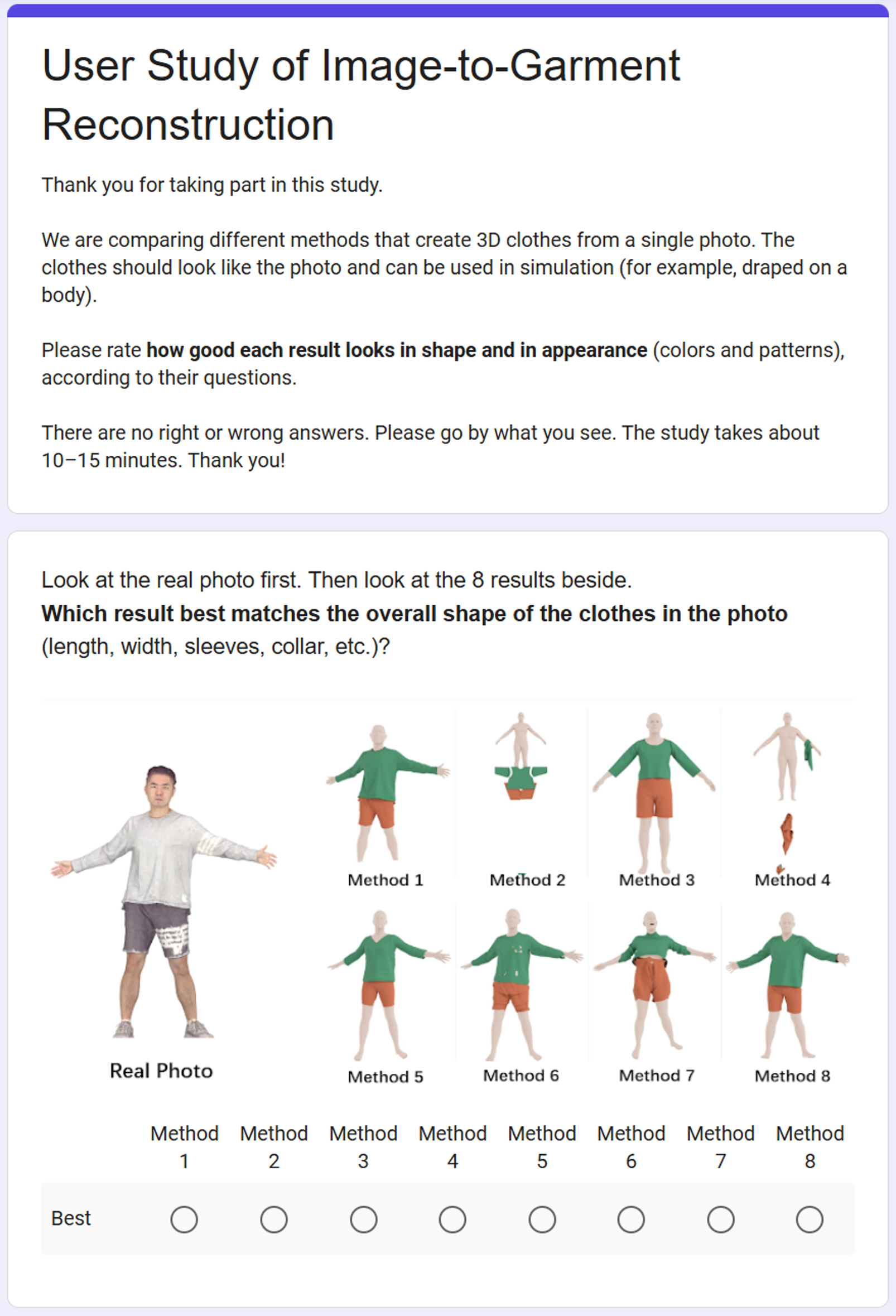}
  \caption{Screenshot of the user study question on geometry.}
  \label{fig:user_shape}
\end{figure}

\begin{figure}[!ht]
  \centering
  \includegraphics[width=0.9\textwidth]{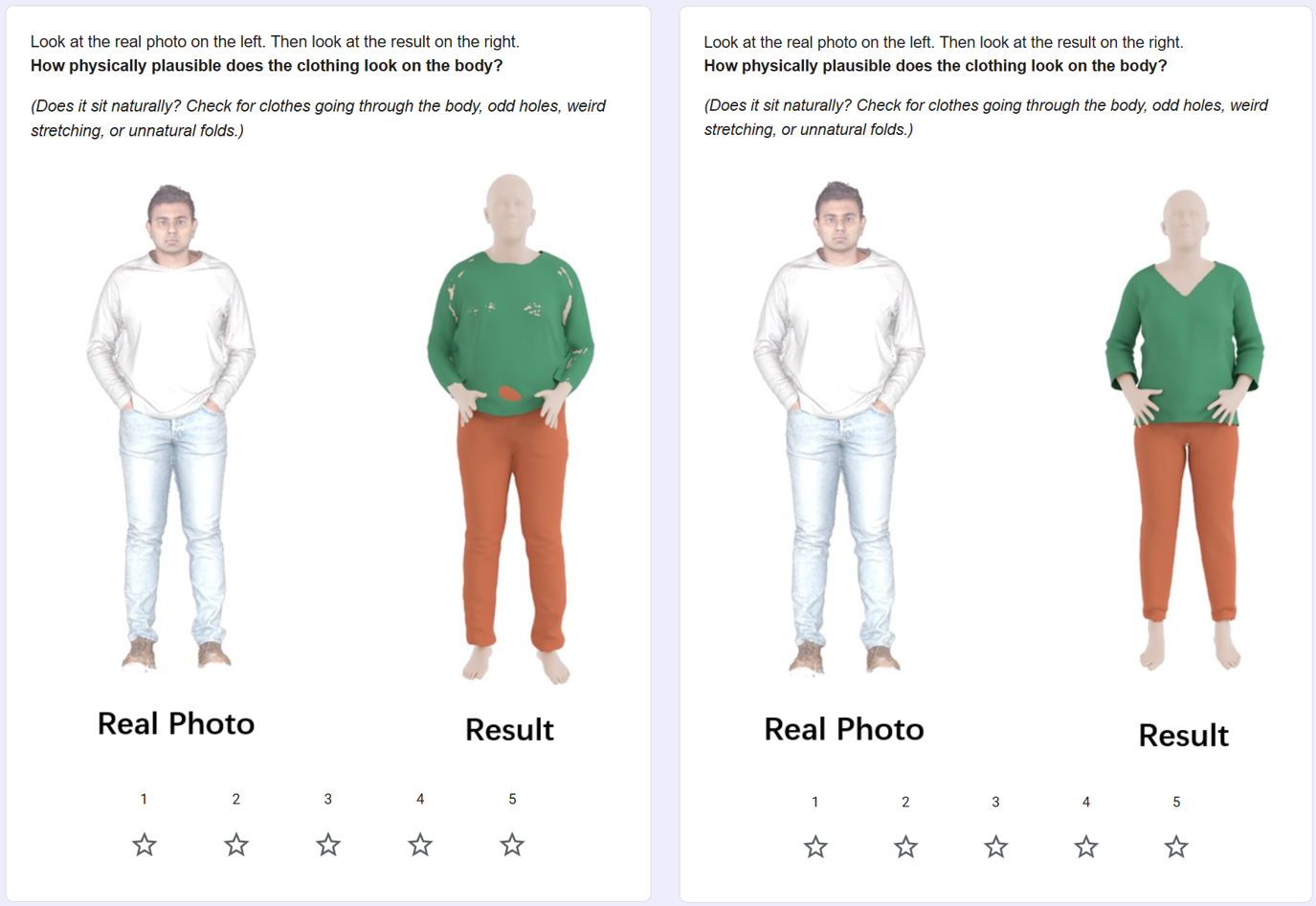}
  \caption{Screenshot of the user study question on plausibility.}
  \label{fig:user_physics}
\end{figure}

\begin{figure}[!ht]
  \centering
  \includegraphics[width=1.0\textwidth]{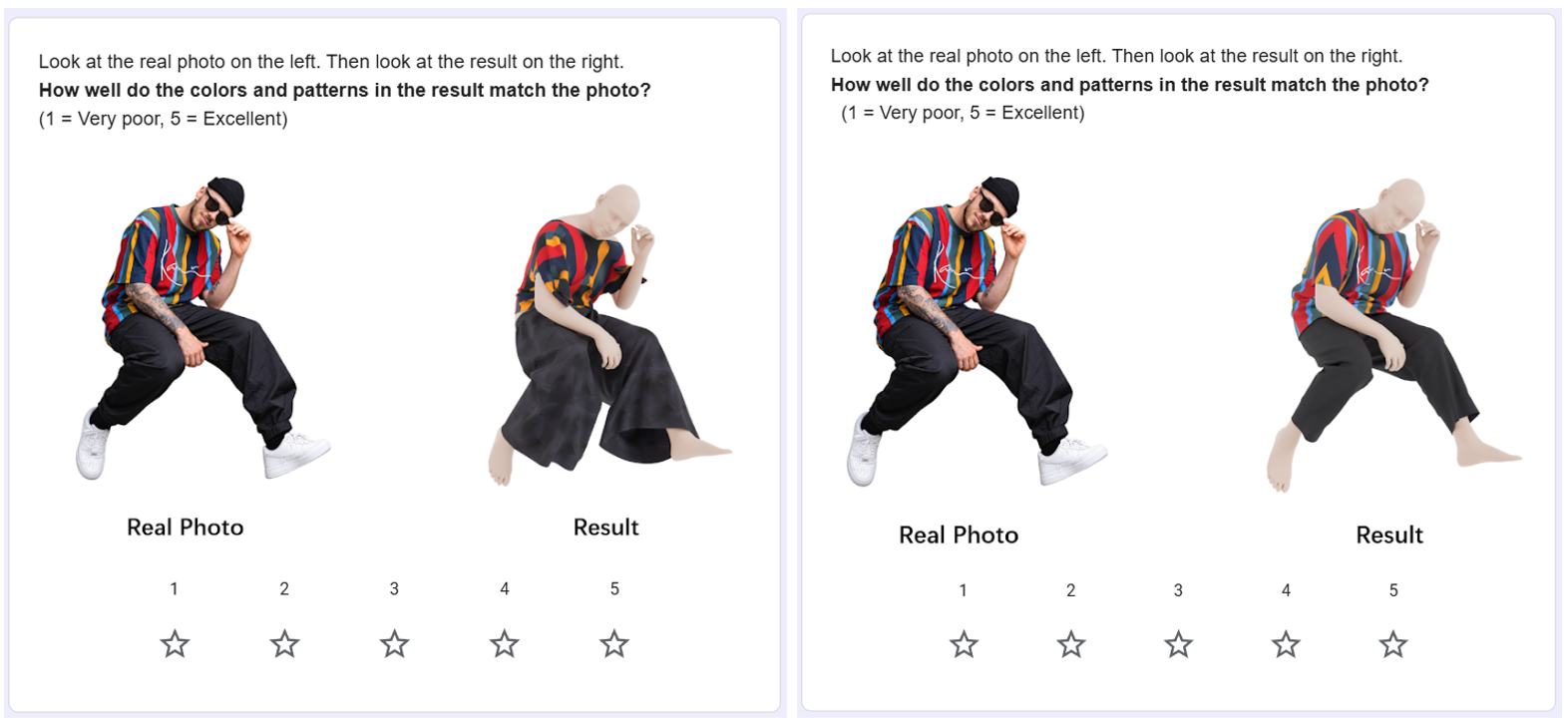}
  \caption{Screenshot of the user study question on texture.}
  \label{fig:user_texture}
\end{figure}

Here, we show examples of the Google Forms questions used in our user study.
Fig.~\ref{fig:user_shape} shows the instructions and the geometry question, where non-artistic participants pick the result that best matches the shape of the clothes in the photo among all methods.
Fig.~\ref{fig:user_physics} and Fig.~\ref{fig:user_texture} show the plausibility and texture questions, where participants rate each result from 1 to 5.
Methods are anonymized and shuffled for every case.

\section{Limitations and Future Work}
\label{appendix:limitations}

\begin{figure}[!ht]
  \centering
  \includegraphics[width=0.6\textwidth]{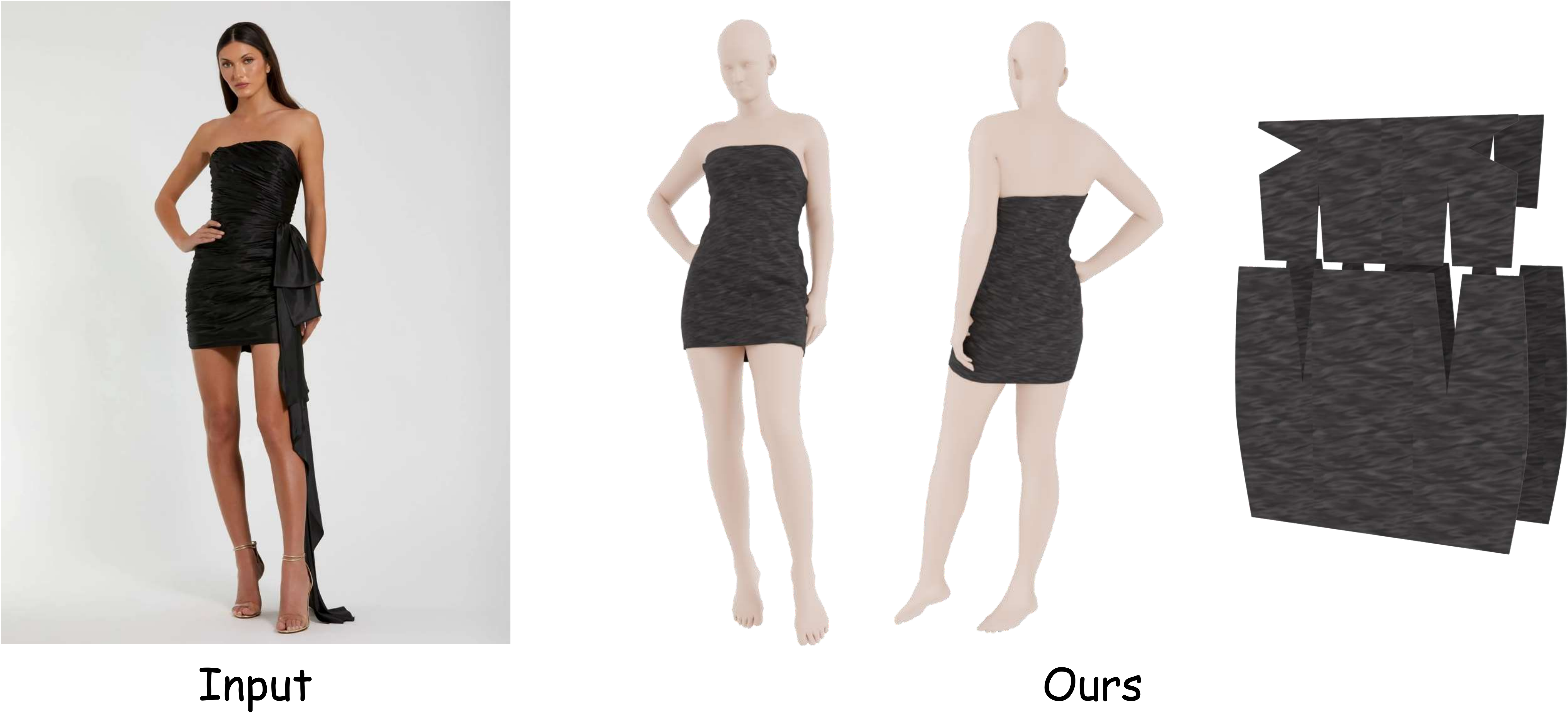}
  \caption{\textbf{Failure case.} The sash hanging free at the waist cannot be recovered by the current representation. Future work may let the agent write code to extend the representation on its own.}
  \label{fig:appendix_failure_case}
\end{figure}

Although \ours generates structured, simulation-ready sewing patterns for a wide range of in-the-wild outfits, three directions remain open for future work:

\textbf{Irregular decoration.}
GarmentCode-XL expresses the vast majority of everyday outfits, but there are free-form ornament that cannot be represented by the current representation, such as the loose sash in Fig.~\ref{fig:appendix_failure_case}.
Future work may let the agent write code to extend the representation on its own, with the input image as guidance.

\textbf{Cloth thickness.}
Current method does not support inverse garment design of garments with thickness, such as a padded down jacket.
As simulation advances and agents grow stronger, future work may let the agent author thickness and material properties along with the pattern.

\textbf{API cost.}
Currently, the full system prioritizes high-fidelity, simulation-ready generation over cost.
In our experiments with GPT-5.6-luna, the average API cost per case is \$0.52, \$0.75, \$0.91, and \$1.49 for outfits with one to four layers.
Cheaper or open-source models lower the cost but also the quality (Fig.~\ref{fig:ablation_models}), reflecting a quality-cost trade-off.

